\documentclass[review]{elsarticle}
\usepackage{lineno,hyperref}
\pdfoutput=1
\usepackage{graphicx}
\usepackage{subfigure}
\usepackage{amsmath}
\usepackage{url}
\usepackage{algorithm}
\usepackage[noend]{algpseudocode}
\usepackage{mathtools}

\DeclarePairedDelimiterX{\infdivx}[2]{(}{)}{%
  #1\;\delimsize\|\;#2%
}
\newcommand{\infdiv}{KL\infdivx}

\makeatletter
\def\BState{\State\hskip-\ALG@thistlm}
\makeatother

\usepackage{pbox}

\usepackage{color}

\newcommand{\NEW}[1]{\textcolor{black}{#1}}

\newcommand{\SEC}[1]{\textcolor{black}{#1}}

\usepackage{cancel}

\usepackage{array}
\usepackage{multirow}

\newcommand\MyBox[2]{
  \fbox{\lower0.75cm
    \vbox to 1.0cm{\vfil
      \hbox to 1.7cm{\hfil  \parbox{0.6cm}{\centering #1\\#2}\hfil}
      \vfil}%
  }%
}
\journal{Robotics and Autonomous Systems}

\begin{document}

\begin{frontmatter}

\author[kovan]{\.Ilker Bozcan \corref{mycorrespondingauthor}}
\cortext[mycorrespondingauthor]{Corresponding author}
\ead{ilker.bozcan@metu.edu.tr}
\author[kovan]{Sinan Kalkan}
\ead{skalkan@metu.edu.tr}

\address[kovan]{KOVAN Research Lab, Dept. of Computer Engineering, Middle East Technical University, Ankara, Turkey}

\title{COSMO: Contextualized Scene Modeling with Boltzmann Machines}


\begin{abstract}
Scene modeling is very crucial for robots that need to perceive, reason about and manipulate the objects in their environments. In this paper, we adapt and extend Boltzmann Machines (BMs) for contextualized scene modeling. Although there are many models on the subject, ours is the first to bring together objects, relations, and affordances in a highly-capable generative model. For this end, we introduce a hybrid version of BMs where relations and affordances are incorporated with shared, tri-way connections into the model. Moreover, we introduce a dataset for relation estimation and modeling studies. We evaluate our method in comparison with several baselines on object estimation, out-of-context object detection, relation estimation, and affordance estimation tasks. Moreover, to illustrate the generative capability of the model, we show several example scenes that the model is able to generate, and demonstrate the benefits of the model on a humanoid robot. The code and the dataset are publicly made available at:
\url{https://github.com/bozcani/COSMO}
\end{abstract}
\end{frontmatter}

\section{Introduction}

\begin{figure}[hbt!]
\centerline{
	\subfigure[]{
    	\includegraphics[width=0.49\textwidth]{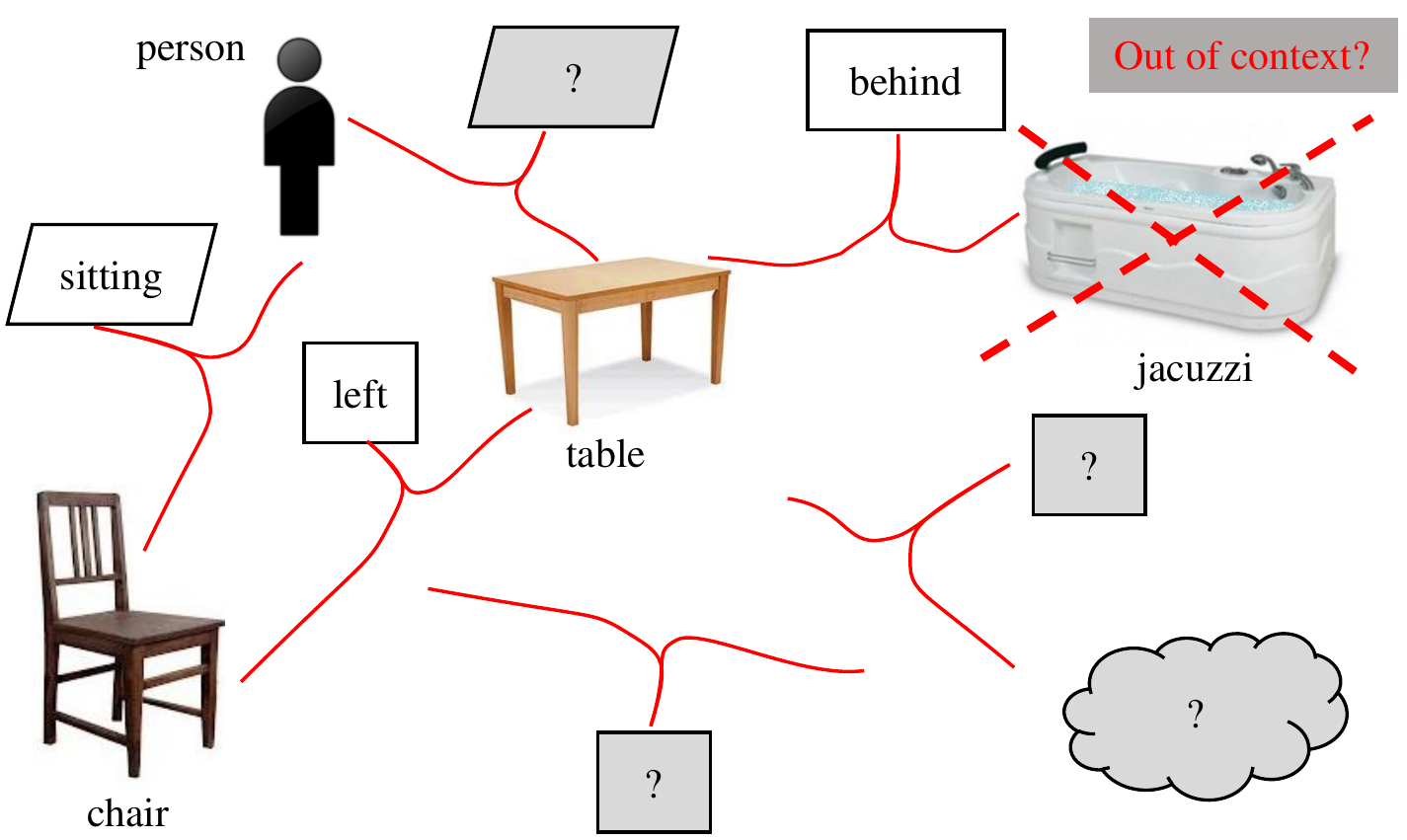}
    }
}
\centerline{
	\subfigure[]{
        \label{fig:triway_BM}
    	\includegraphics[width=0.49\textwidth]{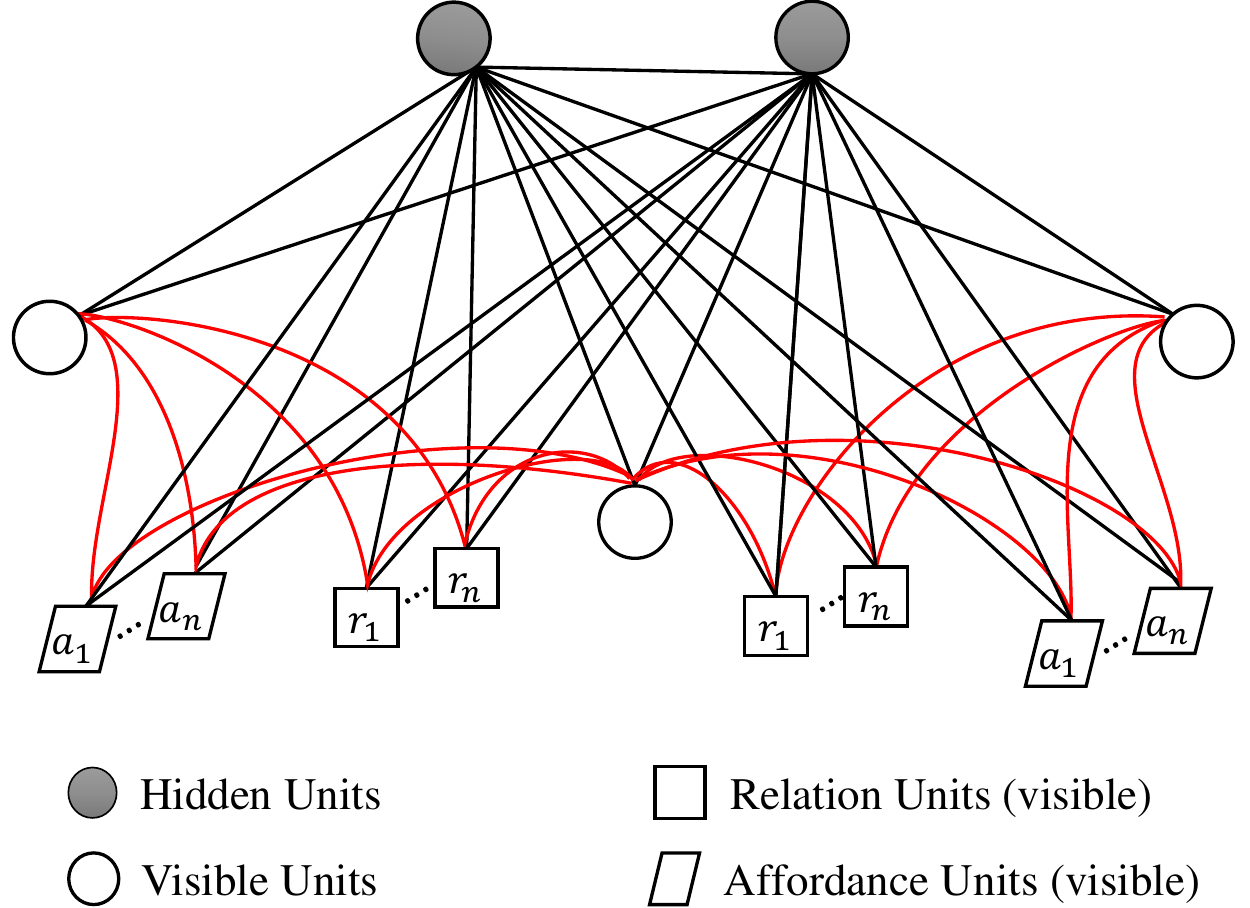}
    }
}
\caption{(a) Example problems for which scene models help robots (given some incomplete or wrong observations from the environment). With our model, we can answer questions marked in gray. (b) An overview of COSMO, our hybrid tri-way Boltzmann Machine, where the tri-way edges are shown in red, as a contextualized scene model. [Best viewed in color]  \label{fig:overview}}
\end{figure}

Having a model, i.e., a representation, of the environment (the current scene) is essential for artificial and biological cognitive agents. A scene model is a representation that allows a robot to reason about the scene and what it contains in an efficient manner. For example, as shown in Figure \ref{fig:overview}(a), using a scene model, a robot can determine (i) whether a certain object is present in the scene and if yes, where it is;  (ii) whether an object is in the right-place in the scene; or (iii) whether there is something not expected or redundant in the scene.

A contextualized scene model, on the other hand, integrates the context of the scene into representing the scene and making inferences about what it contains. This is critical since it has been noted that context plays critical role in perception, reasoning, communication and action \cite{yeh2006situated,barsalou2009simulation}. Context helps these processes in resolving ambiguities, rectifying mispredictions, filtering irrelevant details, and adapting planning. These processes and problems are closely linked to a scene model, and therefore, scene models should contextualize what they represent.

In this paper, we use higher-order Boltzmann Machines \cite{ackley1985learning,sejnowski1986higher} for the first time for contextualized scene modeling. In our model, as shown in Figure \ref{fig:overview}, objects,  spatial relations between objects and affordances are considered as visible units. The hidden (latent) units then represent high-order co-occurrence relations among the visible units, i.e., they capture contextual information about the scene and what it contains. See Section \ref{sect:contributions} for a more detailed analysis of our contributions.

Although there are many studies on scene modeling in robotics, ours is the first to use (and adapt) Boltzmann Machines \NEW{(BMs)} for scene modeling, which not only represent objects or relations between objects in the scene but also affordances of objects. \NEW{BMs have not been used for scene modeling before because (i) BMs need to be adapted to the requirements of the scene modeling problem, such as integrating spatial relations with higher-order edges, and weight-sharing, which are challenging, and (ii) learning in BMs is impractical. In this paper, we propose methods for addressing both challenges.}


 
\section{Related Work}

In this section, we review related work on scene modeling, relation estimation and affordance estimation.

\subsection{Scene Modeling}

\begin{table}[hbt!]
\caption{Comparison with existing studies on Scene Modeling. 
\label{tbl:comparison}}
\begin{center}
\footnotesize
\begin{tabular}{c|ccccc}\hline
Study & Main Method & Generative? & Relations? & Affordances? & Explicit Context?\\  \hline \hline

\cite{joho2013nonparametric} & DB processes & Y  & N & N & N \\
\cite{anand2013contextually} & MRF & Y  & Y & N & N \\
\cite{blumenthal2014towards} & scene graphs & N & N & Y & N \\
\cite{CelikkanatConceptWeb2014}  & MRF & Y & N & Y & N \\
\cite{mastrogiovanni2011robots} & PL & N & Y & Y & N \\
\cite{CelikkanatICAR2015}  & MRF & Y & Y & Y & N \\ 
\cite{pronobis2012large} & chain-graphs & N & Y & N & Y \\
\cite{hwang2006ontology} & PL & N & Y & N & Y \\
\cite{li2017context} & BN & N & Y & N & Y \\
\cite{wang2008spatial, Philbin08a} & LDA v. & Y & Y & N & Y \\
\cite{lin2013holistic} & MRF  & Y & Y & N & Y \\
\cite{CelikkanatContext2014}  & LDA & Y & N & Y & Y \\
\cite{tenorth2009knowrob} & ontology & N & Y & Y & Y \\
\cite{saxena2014robobrain} & ontology & N & Y & Y & Y \\
 \hline
COSMO 	      & BM &  Y & Y & Y & Y \\
\hline

\end{tabular}
\end{center}

\end{table}

Scene modeling is an important problem in Computer Vision and Robotics. During the last decade, especially probabilistic methods or probabilistic graphical models such as Markov Random Fields or Conditional Random Fields \cite{anand2013contextually,CelikkanatConceptWeb2014,lin2013holistic,CelikkanatContext2014}, Bayesian Networks (BN) \cite{li2017context,sheikh2005bayesian}, Latent Dirichlet Allocation variants (LDA v.) \cite{wang2008spatial, Philbin08a}, Dirichlet and Beta (DB) processes \cite{joho2013nonparametric}, chain-graphs \cite{pronobis2012large}, predicate logic (PL) \cite{mastrogiovanni2011robots,hwang2006ontology}, Scene Graphs \cite{blumenthal2014towards}, and ontologies \cite{hwang2006ontology,tenorth2009knowrob,saxena2014robobrain} have been proposed for solving the problem.

Among these studies, similar to ours, there are also models that explicitly integrate context into a scene model \cite{wang2008spatial,Philbin08a,CelikkanatContext2014}. For example, Wang et al. \cite{wang2008spatial} extend LDA to incorporate relative positions between pixels in a local neighborhood in order to segment an image into semantically meaningful regions. Philbin et al. \cite{Philbin08a}, on the other hand, include spatial arrangement between visual patches (i.e., words in LDA) to group similar images into a topic.

Among these, the work of \c{C}elikkanat et al. \cite{CelikkanatContext2014} is the closest to ours. \c{C}elikkanat et al. use object detections as visible variables and context as the latent variable in an LDA model. However, in their work, the main focus was on incremental learning of context nodes, and issues like spatial relations and generative abilities of the scene model were not considered.

\subsection{Relation Estimation and Reasoning}

Without loss of generality, we can broadly analyze relation estimation and reasoning studies in three main categories: The first category of methods use hand-crafted rules to determine whether a predetermined set of spatial relations are present between objects in 2D or 3D, e.g., \cite{stopp1994utilizing}, \NEW{\cite{gatsoulis2016qsrlib, thippur2015comparison, kunze2014combining}}.

In the second category of methods, which use probabilistic graphical models such as Markov Random Fields \cite{anand2013contextually, CelikkanatICAR2015}, Conditional Random Fields \cite{lin2013holistic}, Implicit Shape Models \cite{meissner2013recognizing}, and latent generative models \cite{joho2013nonparametric}, a probability distribution is modeled for relations between objects or entities. In these studies, Anand et al. \cite{anand2013contextually} considered relations like ``on-top'' and ``in-front'' (and their symmetries); Celikkanat et al. \cite{CelikkanatICAR2015} used ``left'', ``on'', and ``in-front'' (and their symmetries); Lin et al. \cite{lin2013holistic} worked with ``on-top'', ``close-to'' relations; Meissner et al. \cite{meissner2013recognizing} took into consideration 6-DoF relations (rotation and translation) between objects. In Joho et al. \cite{joho2013nonparametric}, an implicit model over local arrangements of objects was learned.

In the third category of methods, relation estimation is formulated as a classification problem and solved using discriminative models, such as logistic regression \cite{guadarrama2013grounding}, and deep learning \cite{johnson2016clevr}.	The study by Guadarrama et al. \cite{guadarrama2013grounding} studied relations like ``above'', ``behind'', ``close to'',  ``inside of'', ``on'', and ``left'' (and their symmetries), whereas only two relations (``left'', ``behind'' - and their symmetries) are considered in \cite{johnson2016clevr}.

We see that existing efforts on modeling or estimating relations generally address the problem either for relations or relations and objects, and not consider related concepts such as affordances. Moreover, Boltzmann Machines have not been used for the problem in a scene modeling context.


\subsection{Affordance Prediction}

The concept of affordance, owing to Gibson \cite{Gib86}, pertains to the actions that are provided by entities in the environment to the agents. With suitable formalisms for robotics studies \cite{Sahin2007}, affordance-based models have been used for many important problems, such as manipulation \cite{moldovan2012learning}, navigation \cite{ugur2007learning}, imitation learning \cite{lopes2007affordance}, planning \cite{uyanik2013learning,kalkan2014verb}, and conceptualization \cite{kalkan2014verb,atil2010affordances}  -- see \cite{zech2017computational,jamone2016affordances} for a review.

An important challenge in affordance models is to be able to estimate the affordances of objects from visual input. For this end, support vector machines \cite{uyanik2013learning, koppula2013learning}, bayesian networks \cite{montesano2007modeling,montesano2008learning}, markov random fields \cite{CelikkanatConceptWeb2014, boularias2011learning}, and deep networks \cite{nguyen2016detecting,do2017affordancenet,kokic2017affordance} have been widely used in the literature. However, affordance prediction is generally addressed independently from scene modeling tasks, and to the best of our knowledge, Boltzmann Machines have not been used for modeling affordances.

\subsection{Contributions of the Current Study}
\label{sect:contributions}

Looking also at the summary of the existing studies in Table \ref{tbl:comparison}, we see the following as the main contributions of the current paper:

\begin{itemize}

\item To the best of our knowledge, ours is the first to use Deep Boltzmann Machines (DBM) \cite{salakhutdinov2009deep} for scene modeling. With DBM, we introduce a generative scene model which incorporates objects, spatial relations and affordances. 

\NEW{We prefer BMs for scene modeling for several reasons: (i) Being generative, BMs are able to complete any missing information in the scene and make predictions given any information that may be available. (ii) BMs have explicit representation of nodes. (iii) Input layer can be structured according to the problem domain.}

\item In order to be able to model concepts like relations and affordances that require tri-way connections, we adapt and extend DBM by (i) combining together General BM \cite{ackley1985learning} with higher-order BM \cite{sejnowski1986higher}, and (ii) introducing weight-sharing in order to have the same concepts of relations and affordances between different sets of variables. 
\end{itemize}

\NEW{Note that our model assumes a bag of objects model, where only the presence of objects are considered and their locations are not used. However, we show that, even in this case, our model is very capable at solving many practical robotic tasks.}

We apply our model on relevant robot problems: Determining (i) what is missing in a scene, (ii) relations between objects, (iii) what should not be in a scene, (iv) the affordances of objects, and (v) generating novel scenes given some objects or relations from the to-be-generated scene. We compare our model (COSMO) against DBM \cite{salakhutdinov2009deep} with 2-way relations (GBM), and Restricted Boltzmann Machines (RBM) \cite{salakhutdinov2007restricted}.

The current paper extends our previous work presented at a conference \cite{bozcan2018missing}. To be specific, the current paper extends the model by also including affordances, and performing a more rigorous investigation of the proposed model with extensive experiments on a detailed investigation of the architecture and with experiments with another dataset (namely, visual genome \cite{krishna2017visual}). Moreover, the current paper includes experiments with a humanoid robot.

\section{Background: Boltzmann Machines}

\begin{figure}[hbt!]
\centerline{
\includegraphics[width=0.60\textwidth]{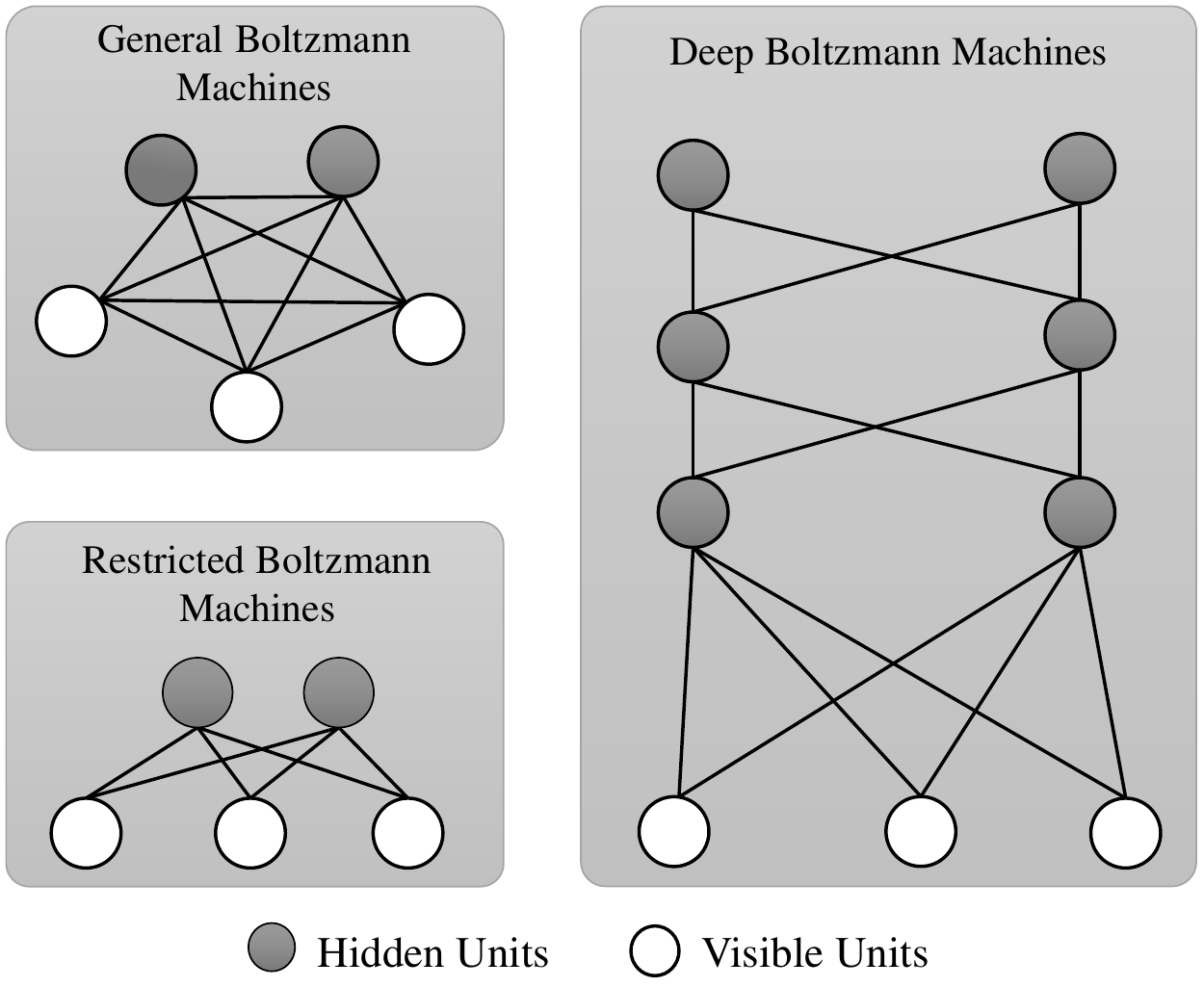}
}
\caption{An illustration of different types of Boltzmann Machines (BM): General BM, Restricted BM and Deep BM. BM is stochastic network that is able to model probability distributions of high-dimensional data, and therefore, generate novel samples. \label{fig:all_bm}}
\end{figure}

A Boltzmann Machine (BM) \cite{ackley1985learning} is a stochastic, generative network\footnote{This section is necessary for explaining our method, although what it covers is textbook material.}. A BM can model the probability distribution of data, denoted by $\mathbf{v}$, with the help of hidden variables, $\mathbf{h}$:
\begin{equation}
p(\mathbf{v}) = \sum_{\mathbf{h}} p(\mathbf{v}, \mathbf{h}).
\end{equation}
In BMs, $\mathbf{v}=\{v_i\}_{i=1}^{V} \subset \{0,1\}^V$  is called the set of visible nodes, and $\mathbf{h}=\{h_i\}_{i=1}^{H} \subset \{0,1\}^H$ the hidden nodes. The visible nodes and the hidden nodes are connected to each other and how they are connected have led to different models -- see Figure \ref{fig:all_bm}. In BMs, the connections are bi-directional; i.e., information can flow in both directions.

In a BM, one can talk about the compatibility, i.e., harmony, between two nodes connected by an edge. If, e.g., $n_i w_{ij} n_j$ is high for two nodes connected by an edge with weight $w_{ij}$, then nodes $n_i$ and $n_j$ are said to be in harmony. However, generally in BMs, the negative harmony, i.e., the energy of the network is used:
\begin{equation}
E(\textbf{v}, \textbf{h})=-\sum_{i<j}v_i w^{vv}_{ij} v_j
              -\sum_{i<j}h_i w^{hh}_{ij} h_j
              -\sum_{i<j}h_i w^{hv}_{ij} v_j, \label{eqn:energy_BM}
\end{equation}
{\noindent}where $w^{vv}, w^{hh}$ and $w^{hv}$ are  the weights of the edges connecting visible-visible nodes, hidden-hidden nodes, and hidden-visible nodes respectively.

Being inspired from statistical mechanics, where systems with lower energies are favored more, BM associates the probability of being in a state (i.e., a configuration of $(\mathbf{v}, \mathbf{h})$) with the energy of the system as follows:
\begin{equation}
p(\mathbf{v}, \mathbf{h})= \frac{1}{Z} \exp(-E(\textbf{v}, \textbf{h})),
\end{equation}
{\noindent}where the normalizing term, also called the partition function, is defined as: $Z=\sum_{\textbf{v'}, \textbf{h'}} E(\textbf{v'}, \textbf{h'})$. Notice that $Z$ requires an integration over all possible states of the system, which is impractical to calculate in practice. Therefore, $p(\mathbf{v}, \mathbf{h})$ is iteratively learned by stochastically activating nodes in the network with probability based on the change in the energy of the system for an update:
\begin{equation}
p(n=1) = \frac{1}{1+e^{\Delta E_n/T}}, \label{eqn:T}
\end{equation}
where $n$ is a visible or a hidden node; $\Delta E_n$ is the change in energy of the system if node $n$ is turned on; and $T$ is the temperature of the system, which is gradually decreased (annealed) to a low value over time. When $T$ is high, the system can make radical updates that can even increase its energy; and when $T$ is lowered, Equation \ref{eqn:T} forces the network to make more deterministic updates, which lower the energy of the system.

\subsection{Training a BM}

Training a BM means that its weights are iteratively updated to model $p(\mathbf{v})$ as accurately as possible. Let us use $p^+(\mathbf{v})$ to denote the true probability of the data, and $p^-(\mathbf{v})$, the probability estimated by the model. Then, a BM is trained in order to minimize the dissimilarity, e.g., the Kullback-Leibler divergence, between $p^+(\mathbf{v})$ and $p^-(\mathbf{v})$. Taking the gradient of the divergence with respect to a weight, $w_{ij}$, gives us the rate at which we should update it in order to minimize the divergence:
\begin{equation}
w_{ij} \leftarrow w_{ij} - \alpha (p^+_{ij} - p^-_{ij}), \label{eqn:grad}
\end{equation}
{\noindent}where $p^+_{ij}$ is the expected joint activation of nodes $s_i$ and $s_j$ when samples from the data are clamped to the visible units and the states of all nodes are iteratively updated until equilibrium (this is called the positive phase); $p^-_{ij}$ is the expected joint activation of nodes $s_i$ and $s_j$ when the network is randomly initialized and the states of the neurons is iteratively updated until equilibrium (called the negative phase); and $\alpha$ is a learning rate.

For training BMs, maximum Likelihood based methods can be used \cite{ackley1985learning, salakhutdinov2009deep,neal1992connectionist}. However, since the partition function, $Z$, is intractable, directly computing $p^+_{ij}$ and $p^-_{ij}$ is not possible for general BMs. Therefore, Monte Carlo Markov Chain methods such as Gibbs sampling or Variational Inference methods such as mean field approaches are used to approximate $p^+_{ij}$ and $p^-_{ij}$. Despite these methods, learning is still impractical owing to the connections within hidden and visible nodes, and potentially high number of hidden nodes.

\subsection{BM Variants}

Since training a BM is rather slow and an obstacle, its restricted version (Restricted Boltzmann Machines) with only connections between hidden and visible nodes have been proposed \cite{salakhutdinov2007restricted}. In a Deep Boltzmann Machine \cite{salakhutdinov2009deep}, on the other hand, there are layers of hidden nodes. See Figure \ref{fig:all_bm} for a schematic comparison of the alternative models.

Some problems require the edges to connect more than two nodes at once, which have led to the Higher-order Boltzmann Machine (HBMs) \cite{sejnowski1986higher}. With a HBM, one can introduce edges of any order to link multiple nodes together.

\section{COSMO: A Contextualized Scene Model with Triway BM}

We extend and adapt DBM for the contextualized scene modeling problem. As shown in Figure \ref{fig:overview}, our model consists of visible (input) layer, where information about the scene is provided, and hidden layer(s), which capture a contextual representation of the scene and its contents. 

We define a scene ($\mathbf{s}\in\mathcal{S}$) to be the tuple of an object vector ($\mathbf{o}$ -- describing objects currently visible to the robot), the vector of the spatial relations between the objects ($\mathbf{r}$), and the vector of affordances ($\mathbf{a}$). A visible node  corresponds to an object, a relation or an affordance, and is set to be active (with value 1) if the corresponding object, affordance or relation is present in the scene (in this sense, $\mathbf{v} = (\mathbf{o}, \mathbf{r}, \mathbf{a})$). The hidden nodes ($\mathbf{h}$) then represent latent joint configurations of the visible nodes; i.e., they correspond to contextual information eminent in the scene.

\begin{figure}[hbt!]
\centerline{
\includegraphics[width=0.25\textwidth]{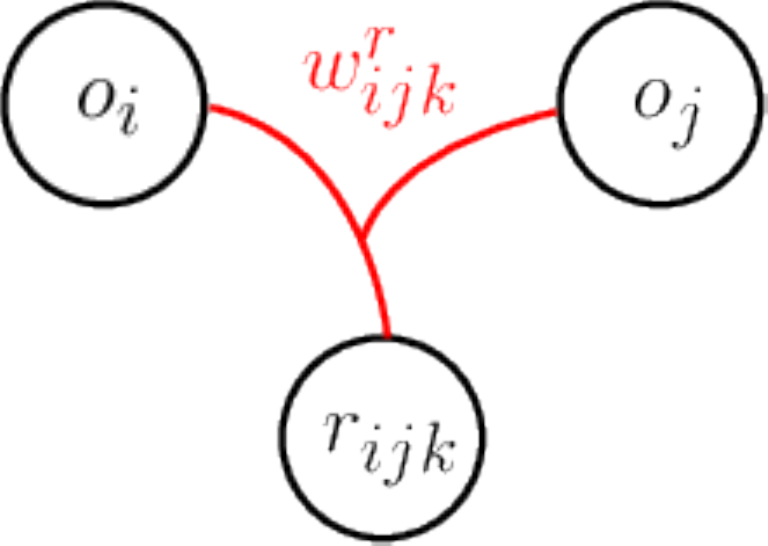}
}
\caption{\NEW{An illustration of a tri-way edge: In this figure, two object nodes ($o_i$ and $o_j$) and one relation ($r_{ijk}$) node are connected with a single edge (drawn in red). The weight of this tri-way edge is denoted by $w^r_{ijk}$. \label{fig:triway_connection}}}
\end{figure}

Relation and affordance nodes link two object nodes with a single \textit{tri-way} edge \NEW{(see Figure \ref{fig:triway_connection})}, and visible nodes are fully connected to hidden nodes ($\mathbf{h}$). To incorporate these changes, the overall energy of the hybrid BM is updated as follows: 
\begin{eqnarray}
E(\textbf{o}, \textbf{h}, \textcolor{red}{\textbf{r}}, \textcolor{red}{\textbf{a}}) & = &   - \sum_{i<j}h_i w^{hv}_{ij} o_j  \\
         & &  \textcolor{red}{-\sum_{i,j,k} w^{r}_{ijk} r_{ijk} o_j o_k}
              \textcolor{red}{-\sum_{i,j,k,l} w^{rh}_{ij} r_{ijk}  h_l} \nonumber \\ 
         & &  \textcolor{red}{-\sum_{i,j,k} w^{a}_{ijk} a_{ijk} o_j o_k}
              \textcolor{red}{-\sum_{i,j,k,l} w^{ah}_{ij} a_{ijk} h_l} , \nonumber \label{eqn:energy_COSMO}
\end{eqnarray}
%
{\noindent}where the new terms compared to the definition in Equation \ref{eqn:energy_BM} are highlighted in red; $r_{ijk}$ denotes a spatial relation node with relation type $i$ between object nodes $o_j$ and $o_k$ (i.e., $r_{ijk}=1$ if relation $i$ exists between objects $o_j$ and $o_k$); $a_{ijk}$ is an affordance relation with affordance type $i$ between objects nodes $o_j$ and $o_k$ (i.e., $a_{ijk}=1$ if object $o_j$ affords ``affordance"' $i$ with object $o_k$); $w^{r}_{ijk}$ is the weight of the tri-way edge connecting object nodes $o_j$, $o_k$ and spatial relation node (visible) $r_i$; and, similarly, $w^{a}_{ijk}$ is the weight of the tri-way edge connecting object nodes $o_j$, $o_k$ and affordance node (visible) $a_i$.

\subsection{Training and Inference}

In order to make training faster, we dropped the connections between the hidden neurons and took the gradient of the divergence ($\infdiv{p^+(\textbf{o}, \textbf{r}, \textbf{a})}{p^-(\textbf{o}, \textbf{r}, \textbf{a})}$) with respect to each type of weight as in Equation \ref{eqn:grad}.

According to the new energy definition (Equation \ref{eqn:energy_COSMO}) and connections, the probabilities of being active for visible and hidden units are given by:
\begin{align}
     p(o_{i}=1\ |\ \mathbf{o}, \mathbf{h}, \mathbf{r}, \mathbf{a})  & =   \sigma \Big(   \sum_{j} h_j w^{ho}_{ij}  +  \sum_{j,k} w^{r}_{ijk} r_{ijk} o_{j}  
      + \sum_{j,k} w^{a}_{ijk} a_{ijk} o_{j}\Big)  ,   \label{eqn:po} \\
p(h_{l}=1\ |\ \mathbf{o}, \mathbf{r}, \mathbf{a}) & =  \sigma \Big( \sum_{i} o_i w^{ho}_{il}  +  \sum_{i,j,k} r_{ijk} w^{rh}_{ij} +  \sum_{i,j,k} a_{ijk} w^{ah}_{ij} \Big)  ,   \label{eqn:ph} \\
p(r_{ijk}=1\ |\ \mathbf{o}, \mathbf{h}) & = \sigma \Big( w^{r}_{ijk} o_j o_k   +  \sum_{l} w^{rh}_{ij} h_l \Big)  , \label{eqn:pr} \\
p(a_{ijk}=1\ |\ \mathbf{o}, \mathbf{h}) & =  \sigma \Big( w^{a}_{ijk} o_j o_k   +  \sum_{l} w^{ah}_{ij} h_l \Big) , \label{eqn:pa}
\end{align}
\NEW{{\noindent}where $\sigma$ is the sigmoid function:
\begin{equation}
\sigma(x) = \frac{1}{1+e^{-x}}.
\end{equation}}
For training COSMO, in the positive phase, as usual, we clamp the visible units with the objects, the relations and the affordances between the objects and calculate $p^+$ for any edge in the network. 

In the negative phase, object units are first sampled with a two-step Gibbs sampling by using the activations of the hidden units only. In this way, initially, the model sees the environment as a bag of objects by not considering relations and affordances. Then, the relation and affordance nodes are sampled by using hidden nodes (context) and recently sampled object nodes. We calculate $p^-$ for any edge in the network with these two steps.

The overall method is summarized in Algorithm \ref{alg:COSMOtraining}.

At the end of the negative phase, the input scene ($\mathbf{s}$) is re-sampled, and $\mathbf{s^{\prime}}$ denotes new scene including recently sampled objects, relations and affordances during negative phase.

\begin{algorithm}[hbt!]
\caption{Training COSMO.}\label{alg:COSMOtraining}
\begin{algorithmic}[1]
\State \textbf{Input:} Training data, $\mathcal{S}=\{\mathbf{s}^i\}_i$; learning rate, $\alpha$; number of epochs, $m$.
\State \textbf{Output:} Learned weights, $\textbf{w}$.
\State
\For {$m$ epochs}
\For {$\mathbf{s} \in \mathcal{S}$}
\State \textit{/* Positive Phase */}
\State $\mathbf{o}^{(0)} \gets \mathbf{s^{\alpha}_o}$, $\mathbf{r}^{(0)} \gets \mathbf{s^{\alpha}_r}$, $\mathbf{a}^{(0)} \gets \mathbf{s^{\alpha}_a}$,
\State $\mathbf{h}^{(0)} \gets p( \mathbf{h}\ |\ \mathbf{o^{(0)}}, \mathbf{r^{(0)}}, \mathbf{a^{(0)}})$
\State Calculate $p^+$ for each edge.
\State
\State \textit{/* Negative Phase */}
\State Sample $\mathbf{\widehat{h}}^{(0)} $ using Eqn. \ref{eqn:ph}. 
\State $\mathbf{o^{(1)}} \gets 0$, $\mathbf{r^{(1)}} \gets 0$, $\mathbf{a^{(1)}} \gets 0$
\State $\mathbf{o}^{(1)} \gets p( \mathbf{o}\ |\ \mathbf{o}^{(1)}, \mathbf{\widehat{h}}^{(0)}, \mathbf{r}^{(1)}, \mathbf{a}^{(1)})$
\State $\mathbf{r}^{(1)} \gets p( \mathbf{r}\ |\ \mathbf{o}^{(1)}, \mathbf{\widehat{h}}^{(0)}, \mathbf{a}^{(1)})$
\State $\mathbf{a}^{(1)} \gets p( \mathbf{a}\ |\ \mathbf{o}^{(1)}, \mathbf{\widehat{h}}^{(0)}, \mathbf{r}^{(1)})$
\State Sample $\mathbf{\widehat{o}}^{(1)}, \mathbf{\widehat{r}}^{(1)}, \mathbf{\widehat{a}}^{(1)}$  using Eqn. \ref{eqn:po}, \ref{eqn:pr}, \ref{eqn:pa}.
\State $\mathbf{h}^{(1)} \gets p( \mathbf{h}\ |\ \mathbf{\widehat{o}}^{(1)}, \mathbf{\widehat{r}}^{(1)}, \mathbf{\widehat{a}}^{(1)})$
\State  Calculate $p^-$ for each edge.
\State
\State Update weights using Eqn. \ref{eqn:grad}.
\EndFor
\EndFor
\end{algorithmic}
\end{algorithm}

Since our dataset has small number of samples and input vectors are too sparse, precise inferences are crucial. Therefore, we prefer Gibbs sampling \cite{geman1984stochastic}, which is a Monte Carlo Markov Chain (MCMC) method to approximate true data distribution, instead of variational inference since MCMC methods can provide precise inference but variational inference methods cannot guarantee that \cite{blei2017variational}.

\NEW{During testing, COSMO is clamped with observed data at its nodes, and the states of the neurons are updated iteratively using Gibbs sampling towards thermal equilibrium. This iterative update process is called ``relaxing''. After the model is relaxed, the activations of the neurons can be used to reason about the scene.}

\section{Experiments and Results}

In this section, we evaluate COSMO on several scene modeling and robotics problems and compare the model against several baselines, and alternative methods whenever possible.

\subsection{The Dataset}
\label{sect:dataset}


For our experiments, we formed a dataset composed of $6,976$ scenes, half of which is sampled from the \textit{Visual Genome} (VG) dataset \cite{krishna2017visual} and the other half from the SUN-RGBD dataset \cite{song2015sun}. We used samples from both datasets since (i) the VG dataset has spatial relationships but these do not include relations useful for robots, such as left and right, and (ii) the VG dataset mostly includes outdoor datasets. We compensate these using the SUN-RGBD dataset, which is composed of indoor scenes only. Therefore, we included equal number of samples from both the VG and the SUN-RGBD datasets. However, the SUN-RGBD dataset did not have spatial relations labeled, therefore, we did manual labeling for the SUN-RGBD dataset.

Our dataset consists of 90 objects that commonly exist in scenes, including human-like (man, woman, boy etc.), physical objects (cup, bottle, jacket etc.), part of buildings (door, window etc.). 


Our dataset is composed of the following eight spatial relations: left, right, front, behind, on-top, under, above, below. These spatial relations are annotated in the VG dataset already. However, we extended the original SUN-RGBD dataset by manually annotating these eight spatial relations. Moreover, we included verb-relations in the VG dataset as affordances into the dataset. The set of affordances include eat-ability, push-ability, play-ability, wear-ability, sit-ability, hold-ability, carry-ability, ride-ability, push-ability, use-ability.


Let us use $\mathcal{S} =\{\textbf{s}_1,...,\textbf{s}_{6,976}\}$, where $\textbf{s}_i$ denotes $i^{th}$ sample, to denote the dataset. $\textbf{s}_i$ has a vector form that represents the presence of objects, relations and affordances among them in the scene. Active (observed) variables are set to value 1, or to value 0 otherwise. Opposite spatial relations (e.g., left and right) can be represented as single relations in BMs since if object $o_1$ is to the left of object $o_2$, then we can state that object $o_2$ is to the right of object $o_1$. As a result, each sample is represented by a binary vector that has length $113,490$ ($90+14\times 90\times 90$). 

\begin{figure}
\centering{
    	\includegraphics[width=0.6\textwidth]{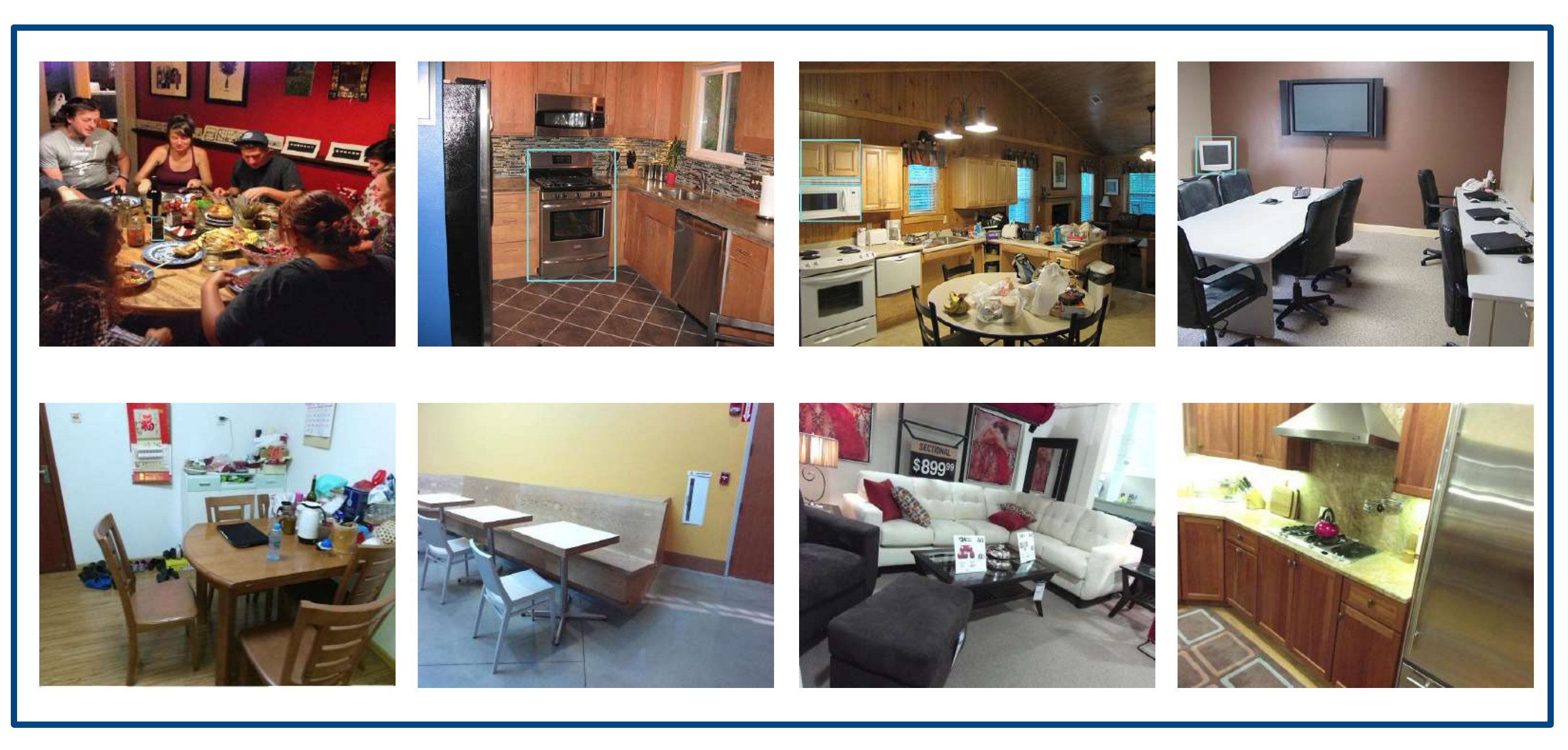}
}
\caption{Example scenes from the merged dataset used for the experiments. \label{fig:dataset}}
\end{figure}

\subsection{Compared Models}

We compare COSMO with General Boltzmann Machine (GBM), Restricted Boltzmann Machine (RBM) and Relational Network (RN) \cite{santoro2017simple} for several scene reasoning tasks that are crucial for various robotic scenarios. For GBM and RBM models, we used the same number of hidden nodes as in COSMO -- see Section \ref{sect:hyperparameters}.

\subsubsection{General Boltzmann Machine (GBM)}

GBMs are unrestricted in terms of connectivity or the hierarchy in the network (either among the hidden or the visible nodes). However, this may make learning impractical, especially when hidden nodes are connected to each other. We allow connections within visible nodes to incorporate interactions between objects as required for scene modeling. In this structure, visible nodes consist of object, relation and affordance nodes as in COSMO. Unlike COSMO, GBM uses two-way edges for relation and affordance nodes, instead of tri-way edges. Similar to COSMO, all visible nodes are fully connected to the hidden nodes but connections within hidden nodes are not allowed. Therefore, the only difference between COSMO and the GBM model is how relation and affordance nodes are connected to objects. To make it comparable with COSMO, we used the same number of layers and hidden neurons in the GBM model as in COSMO.

\subsubsection{Restricted Boltzmann Machine (RBM)}

Different from GBM, an RBM only allows connections between visible and hidden nodes. To make it comparable with COSMO, we used the same number of layers and hidden neurons in the RBM model as in COSMO.

\subsubsection{Relation Network (RN)}

\begin{figure}
\centering{
    	\includegraphics[width=0.7\textwidth]{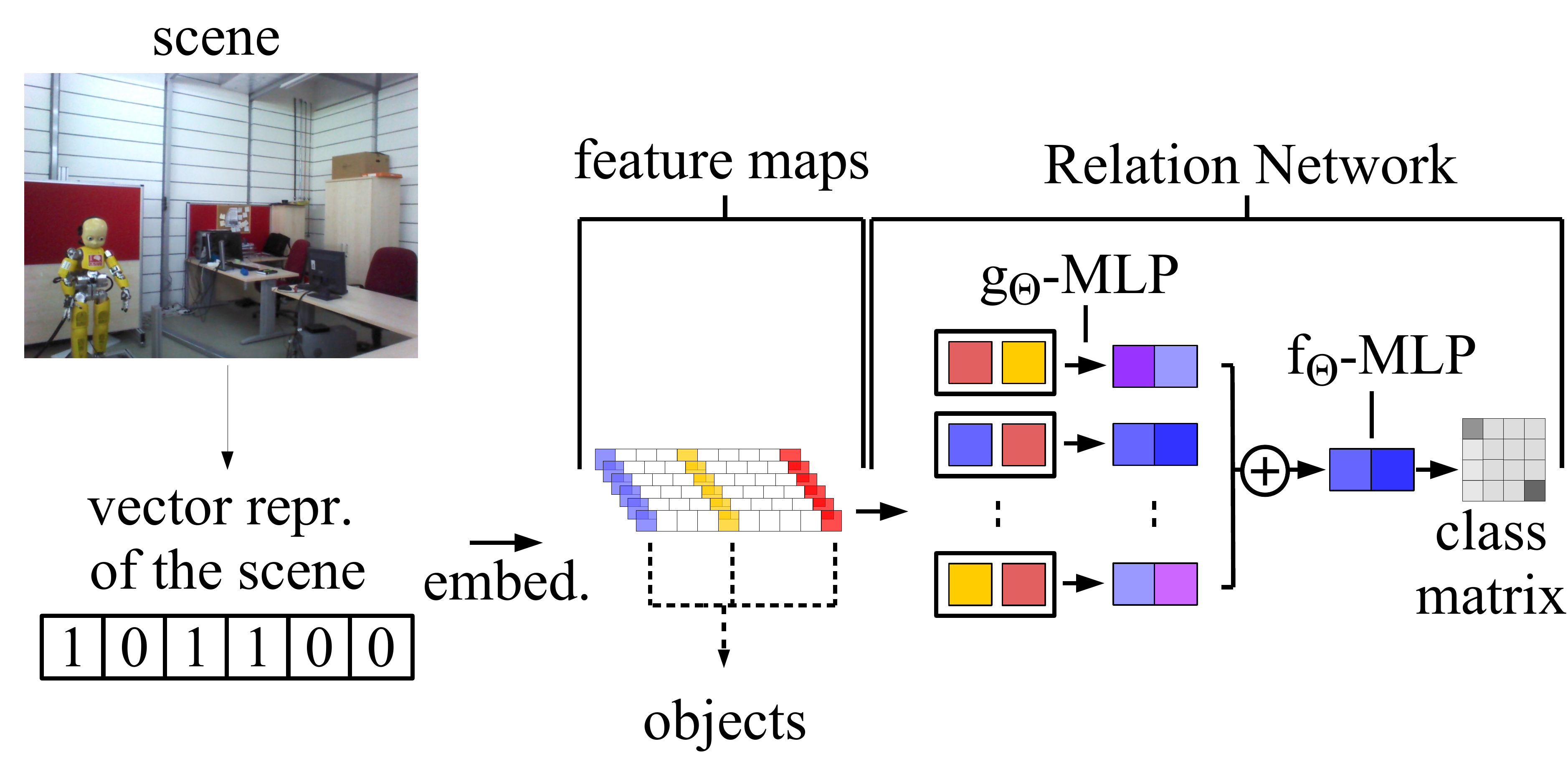}   
}
\caption{\textbf{Relational network (RN) architecture.} A scene is represented by binary vector that indicates presence of objects, spatial relations and affordances among them. The input vector is embedded using Multi-Layer-Perceptron (MLP). Activations of the MLP are used as feature maps to produce a set of objects for RN. Objects are illustrated as blue, yellow and red. Object pairs are fed into the $g$ network whose output is fed into the $f$ network to compute the relations. [Best viewed in color] \label{fig:RN}}
\end{figure}

RNs \cite{santoro2017simple} are simple neural networks to address problems related to relational reasoning. We modified RNs as shown in Figure \ref{fig:RN} to make them compatible for our experiments: The input vectors are embedded with a Multi-Layer-Perceptron (MLP) and the activations of MLP are used as object pairs for another MLP, called the $g$ network. In the original model, object pairs are concatenated with an embedding of a query text; however, we omit this since we assume that the model has one type of question for each scenario. For example, for the spatial relation estimation task, only object and affordance vectors are used as input and spatial relations are predicted. In this case, the model is trying to answer the question ``what are relations among all objects in the scene?". Training RN to answer a specific question ``what is the relation between object $a$ and object $b$" requires additional training samples, including question-answer pairs. These are crucial drawbacks of RN when compared to COSMO. Being generative, COSMO have more flexibility on what can be queried with the scene modeled.

Our implementation of the RN method closely follows the original study. However, we had to adjust the architecture to fit to our data sizes. The embedding MLP network is composed of 2 layers (with 128 and 128 neurons respectively) with the ReLU non-linearities. The $g$ network is a MLP with 2 layers (with 256 and 256 neurons respectively) with the sigmoid non-linearities. The prediction network, $f$, then is a MLP with 2 layers (with 64 and 64 neurons respectively) with sigmoid non-linearities. 

We trained the RN model using the Adam optimizer with default parameters (and learning rate of $0.001$), 32 sized-batches and early stopping.

\subsection{Network Training Performance}

The dataset (composed of $6,976$ scenes) is split into three randomly: 60\% for training, 30\% for testing and 10\% for validation. This split is used for training and testing all methods. For evaluating the training performance, we calculated an error on the difference between the clamped visible units and reconstructed visible states that are sampled in the negative phase: 
\begin{equation}
E_{train} = \frac{1}{|\mathcal{S}|}\sum_{\mathbf{s}\ \in\ \mathcal{S}}\sum_{i} \left(p(s_i^{+})-p(s_i^{-}) \right)^2 ,
\label{train_error_equation}
\end{equation}
{\noindent}where the cumulative sum is normalized with the total number of samples ($|\mathcal{S}|$).

\begin{figure}[hbt!]
\centerline{
	\subfigure[]{
    	\label{fig:error_vs_epoch_objects}
    	\includegraphics[width=0.6\textwidth]{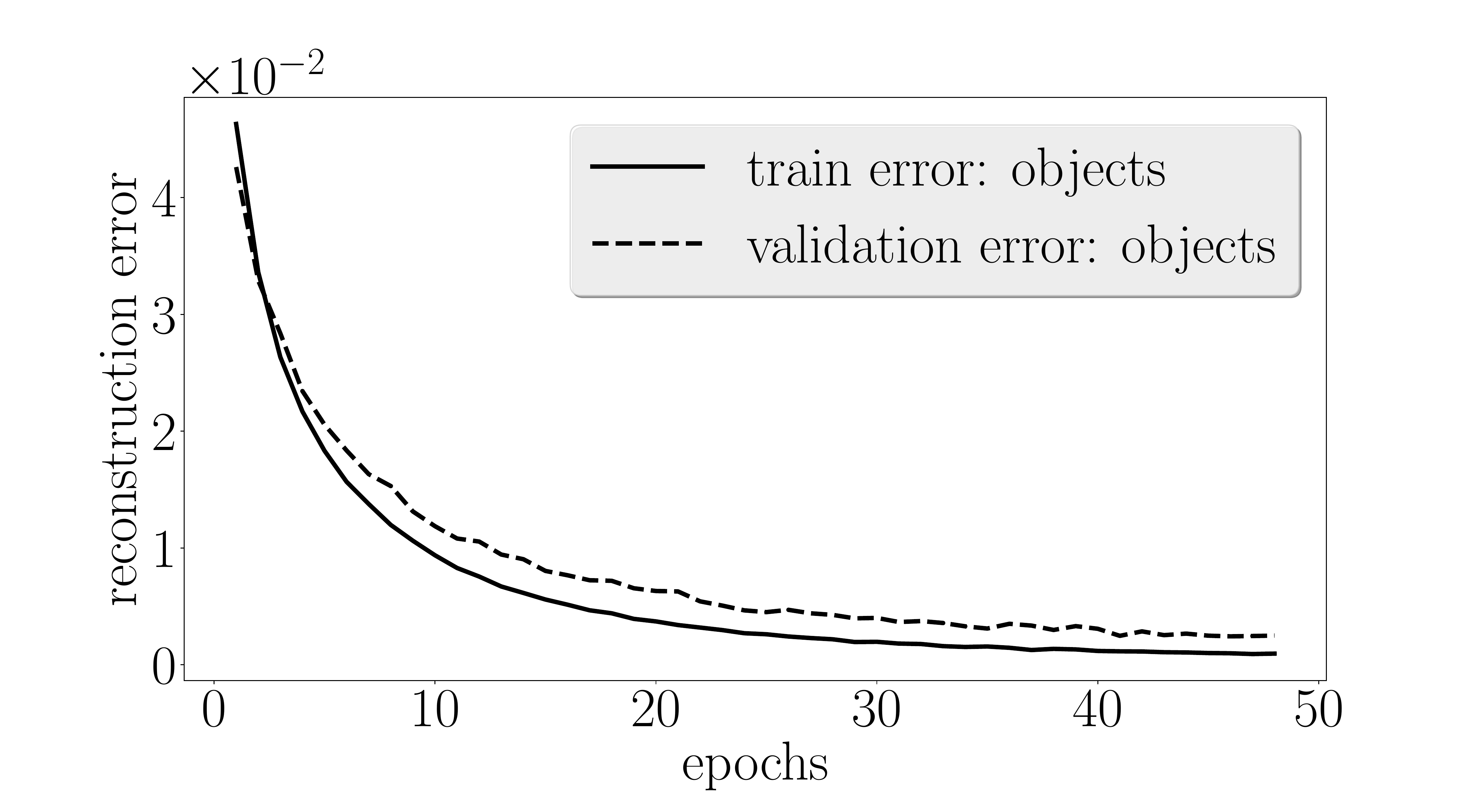}
        }
}
\centerline{ 
	\subfigure[]{
    	\label{fig:error_vs_epoch_relations}
    	\includegraphics[width=0.6\textwidth]{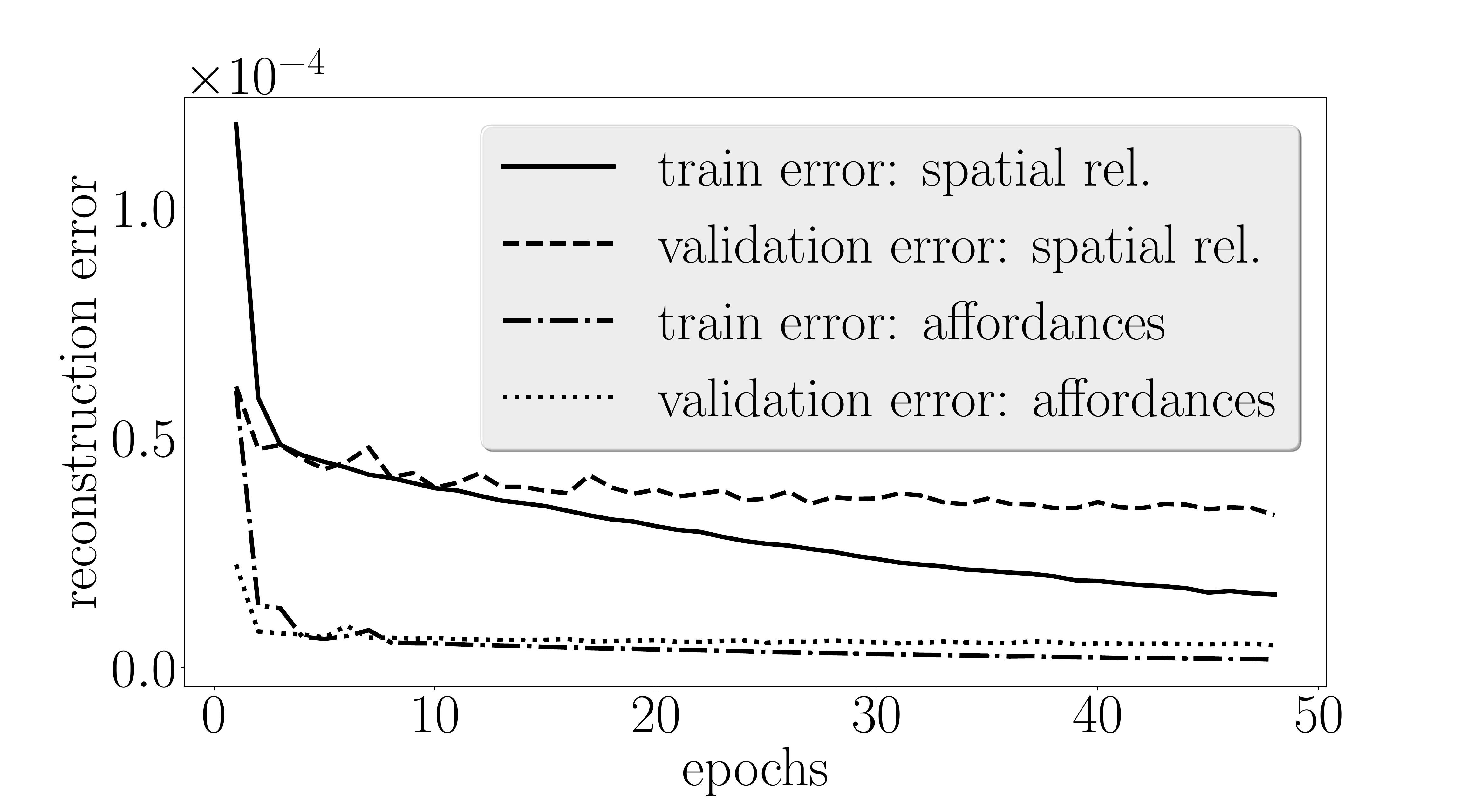}
    }
}
\caption{Reconstruction error vs. epochs plot during COSMO training for (a) objects and (b) spatial relations and affordances. \label{fig:error_vs_epochs}}
\end{figure}

Figure \ref{fig:error_vs_epochs} plots the error separately for the objects ($\mathbf{o}$), the spatial relations ($\mathbf{r}$) and the affordances ($\mathbf{a}$). From the figure, we observe that the error is consistently decreasing for all types of visible units for both the training data and the validation data, suggesting that the network is learning to represent the probability over objects, the spatial relations and the affordances very well.

However, we observe in Figure \ref{fig:error_vs_epochs}(b) that the network learns affordances  faster than objects and relations. This difference is owing to the fact that the set of possible affordances in a scene is much sparser than objects and relations, making the network quickly learn to estimate 0 (zero) for \NEW{most of the} affordances, leading to a sudden decrease in the loss.

\NEW{Now we compare models in terms of average running time for one epoch during training (Table \ref{tbl:time_and_numparams}). For comparability, the parameter counts were fixed as much as possible for this experiment. We observe that RN has the lowest running time and that the variants of Boltzmann Machines are slower compared to RN due to the Gibbs sampling step.}

\begin{table}[hbt]
\caption{Average running time in seconds (time) for one epoch and total number of parameters for different models.\label{tbl:time_and_numparams}}
\centering
\footnotesize
\begin{tabular}{|l|cccc|}\hline                                                                 
	&COSMO  & GBM  & RBM & RN  \\ \hline \hline
Time (seconds) & $181.06$  & $208.22$ & $179.06$ & \textbf{$102.30$}  \\ \hline 
\# of params. & $13,802,600$ & $13,871,200$ & $13,734,000$ & $12,463,104$ \\  \hline
\end{tabular}
\end{table}

\subsection{Analyzing the Hyper-parameters}
\label{sect:hyperparameters}

We evaluated the effects of various hyper-parameters on COSMO's training performance (Equation \ref{train_error_equation}). For all the analyses performed in this section, we looked at the error on the validation data (see Section \ref{sect:dataset}). \SEC{In the appendix, we provide an analysis of the hyper-parameters of GBM, RBM and RN.}


First, we analyzed the effect of the number of hidden layers. For this end, we tested models with  1, 2, 3, 4 and 5 hidden layers. \NEW{Layer-wise pretraining is used for good initialization of models' (with more than one hidden layers) weights as proposed in \cite{salakhutdinov2009deep}. For this end, DBM models are considered as a stack of RBMs and each RBM is trained separately by using hidden states of the previous RBM as the input. For calculating the activations of the internal hidden units in a RBM, the weights are doubled to compensate lack of feedback from the connected nodes that are not part of the current RBM, as suggested in \cite{salakhutdinov2009deep}.}

\begin{figure}[hbt!]
\centerline{
	\subfigure[objects]{
    	\includegraphics[width=0.6\textwidth]{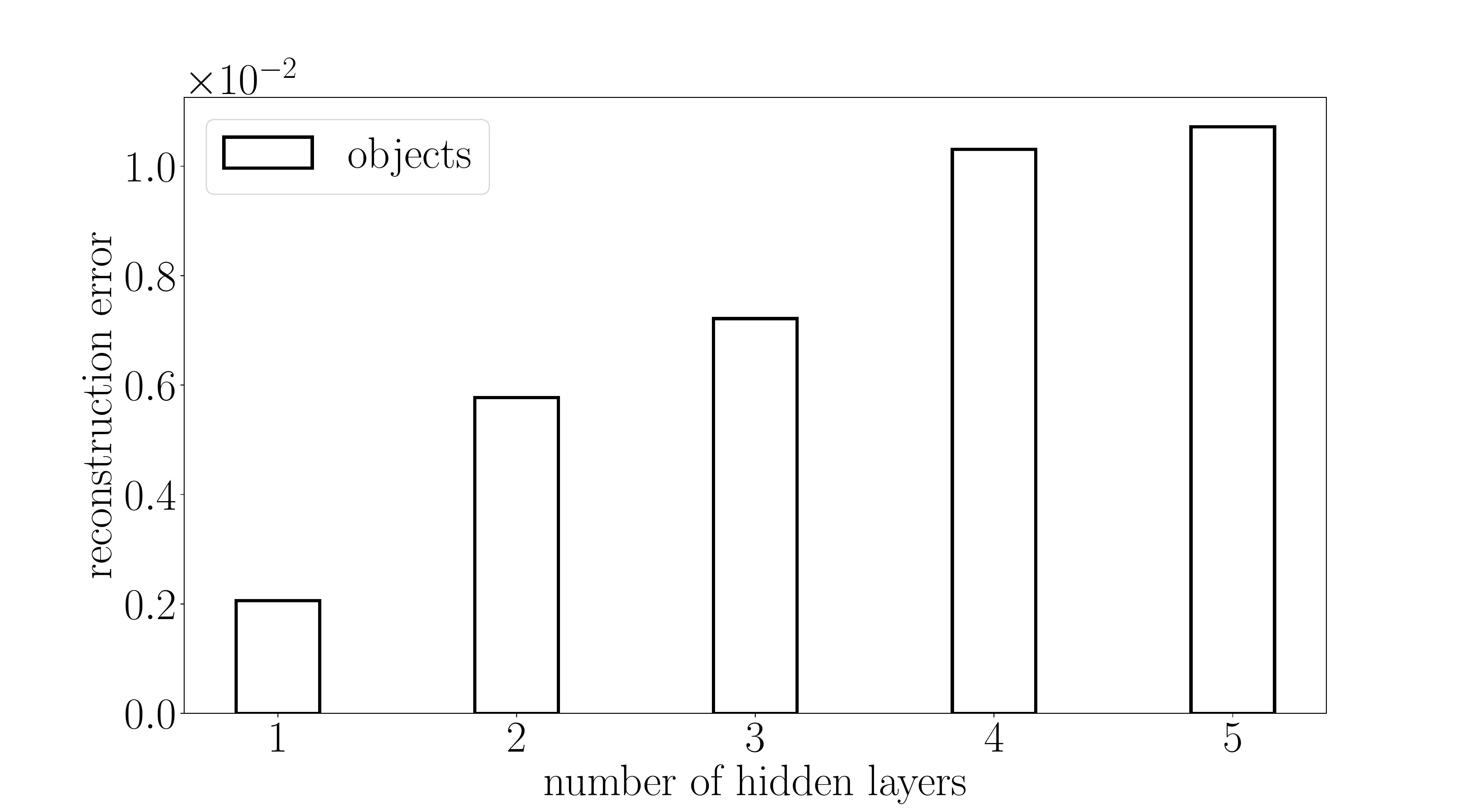}
    }
}
\centerline{
	\subfigure[relations and affordances]{
    	\includegraphics[width=0.6\textwidth]{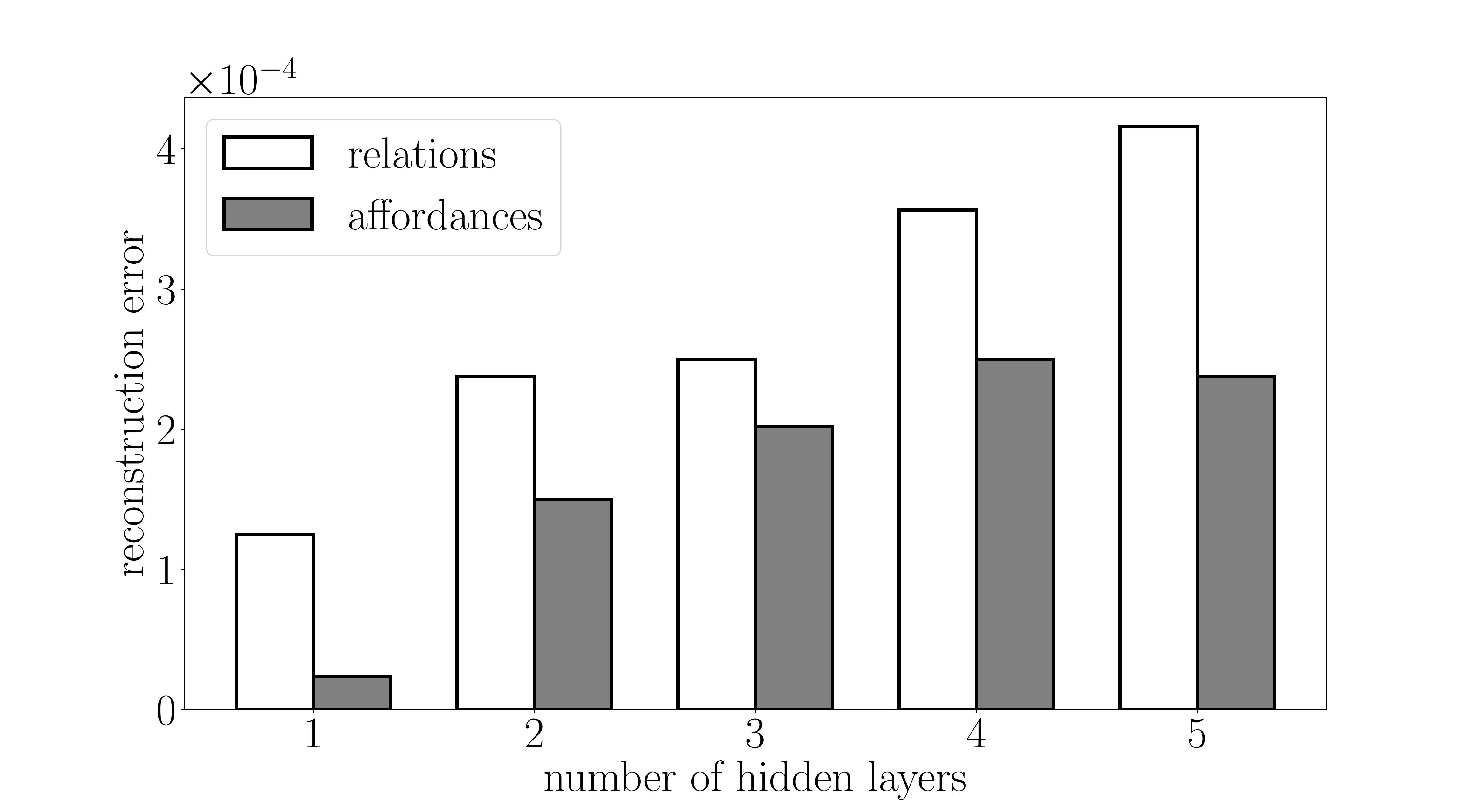}
    }
}
\caption{Reconstruction errors (after 30 epochs) for different numbers of hidden layers. (a) Reconstruction error for object nodes. (b) Reconstruction error for relation and affordance nodes.\label{fig:error_vs_layers}}
\end{figure}

As shown in Figure \ref{fig:error_vs_layers}, the model yields the lowest reconstruction error (for all types of visible nodes) with only one hidden layer, and the error increases when the number of hidden layers is incremented. \NEW{This is rather unexpected since, in deep learning, increasing the number of layers generally leads to better performance. This phenomenon with Boltzmann Machines has already been highlighted by Hinton and his colleagues \cite{salakhutdinov2010efficient,hinton2012better}. The issue is that training  deep BMs becomes problematic with increasing number of hidden layers since they require sampling at each hidden layer and dependencies between hidden units in internal layers can be very complex for ``deeper'' Boltzmann Machines. This is mainly because generating a sample from a deep BM is difficult, since it is necessary to use Monte Carlo Markov Chain (MCMC) methods across all hidden layers, making training much slower to converge with the increasing number of hidden layers.} 

\begin{figure}[hbt!]
\centerline{
	\subfigure[objects]{
    	\includegraphics[width=0.6\textwidth]{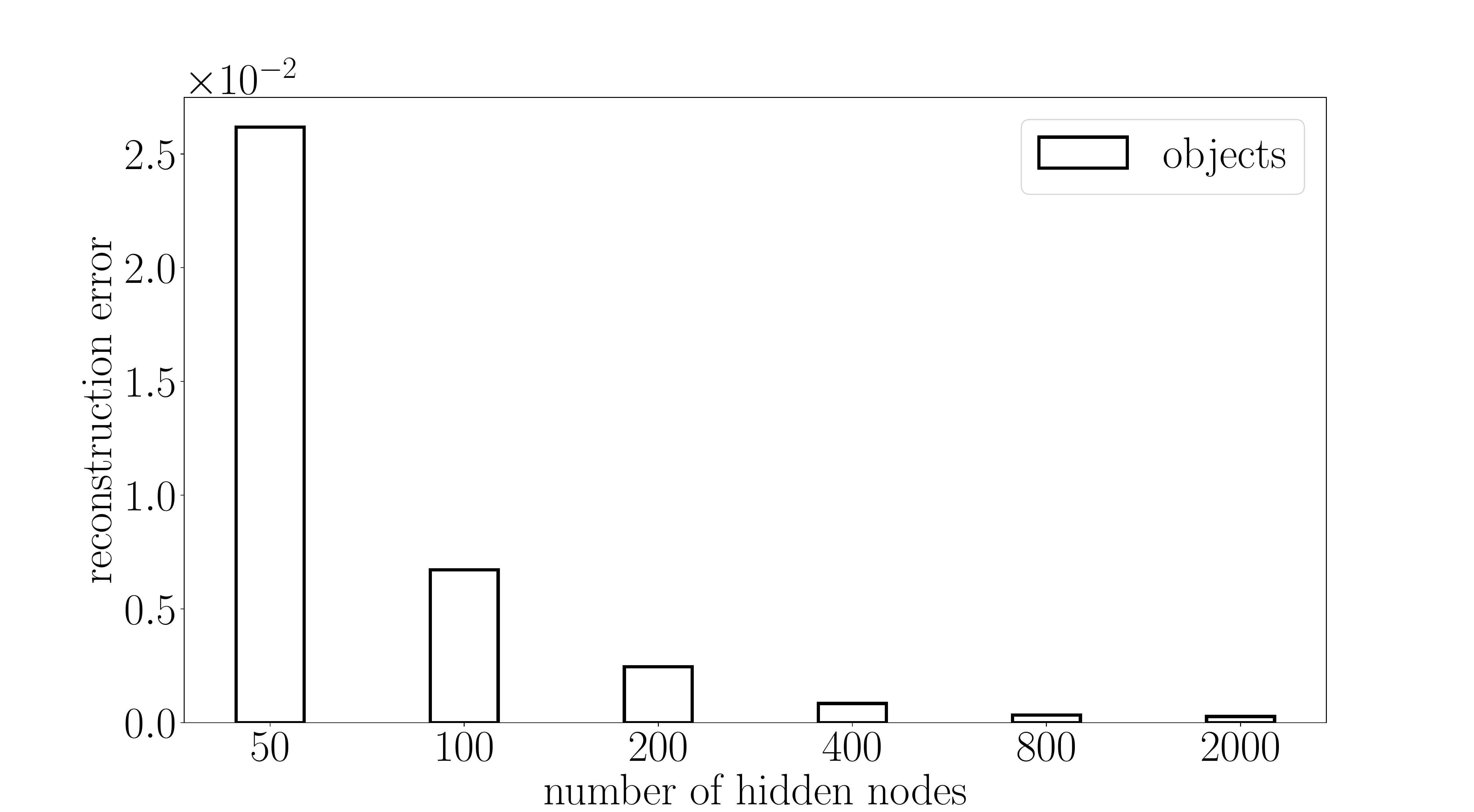}
    }
}
\centerline{
	\subfigure[relations and affordances]{
    	\includegraphics[width=0.6\textwidth]{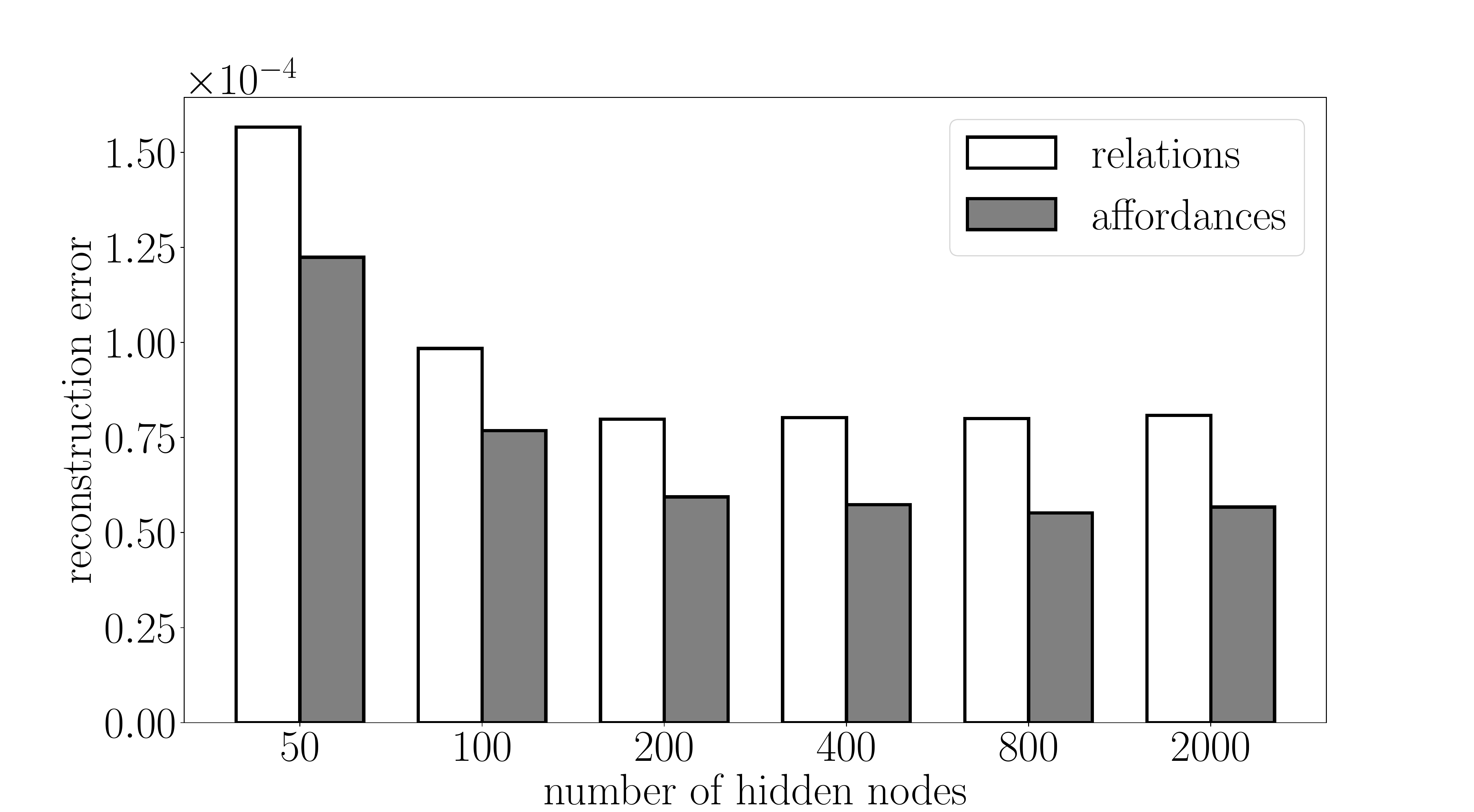}
    }
}
\caption{Reconstruction errors (after 30 epochs) for different number of hidden nodes. (a) Reconstruction error for object nodes. (b) Reconstruction error for relation nodes and affordance nodes. \label{fig:error_vs_hiddens}}
\end{figure}

Secondly, we analyzed the effect of the number of hidden neurons in a hidden layer. We tested 50, 100, 200, 400, 800 and 2000 hidden neurons. As shown in Figure \ref{fig:error_vs_hiddens}, the reconstruction error decreases when the number of hidden neurons increases, as expected.

\begin{figure}[hbt!]
\centerline{
	\subfigure[objects]{
    	\includegraphics[width=0.6\textwidth]{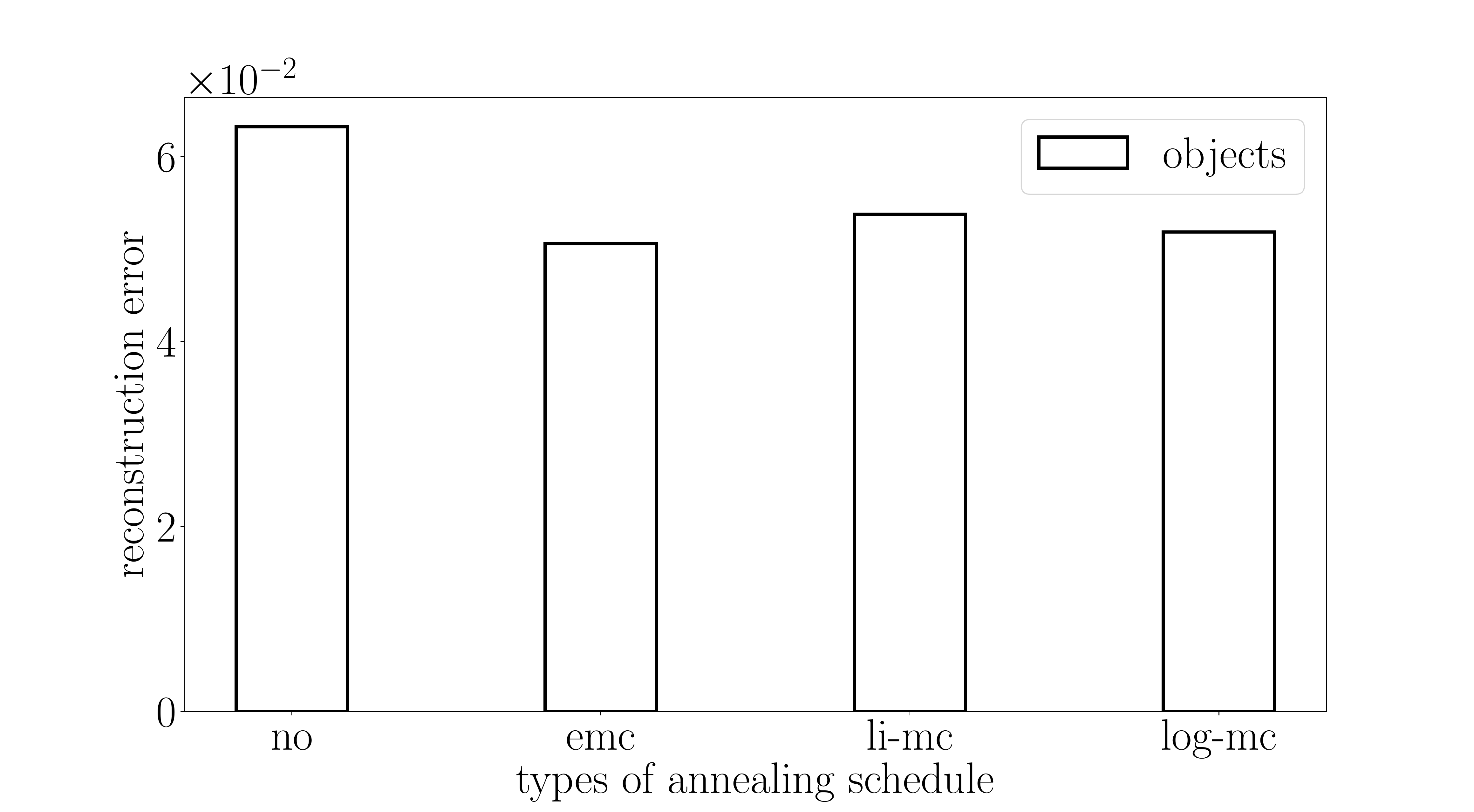}
    }
}
\centerline{
	\subfigure[relations and affordances]{
    	\includegraphics[width=0.6\textwidth]{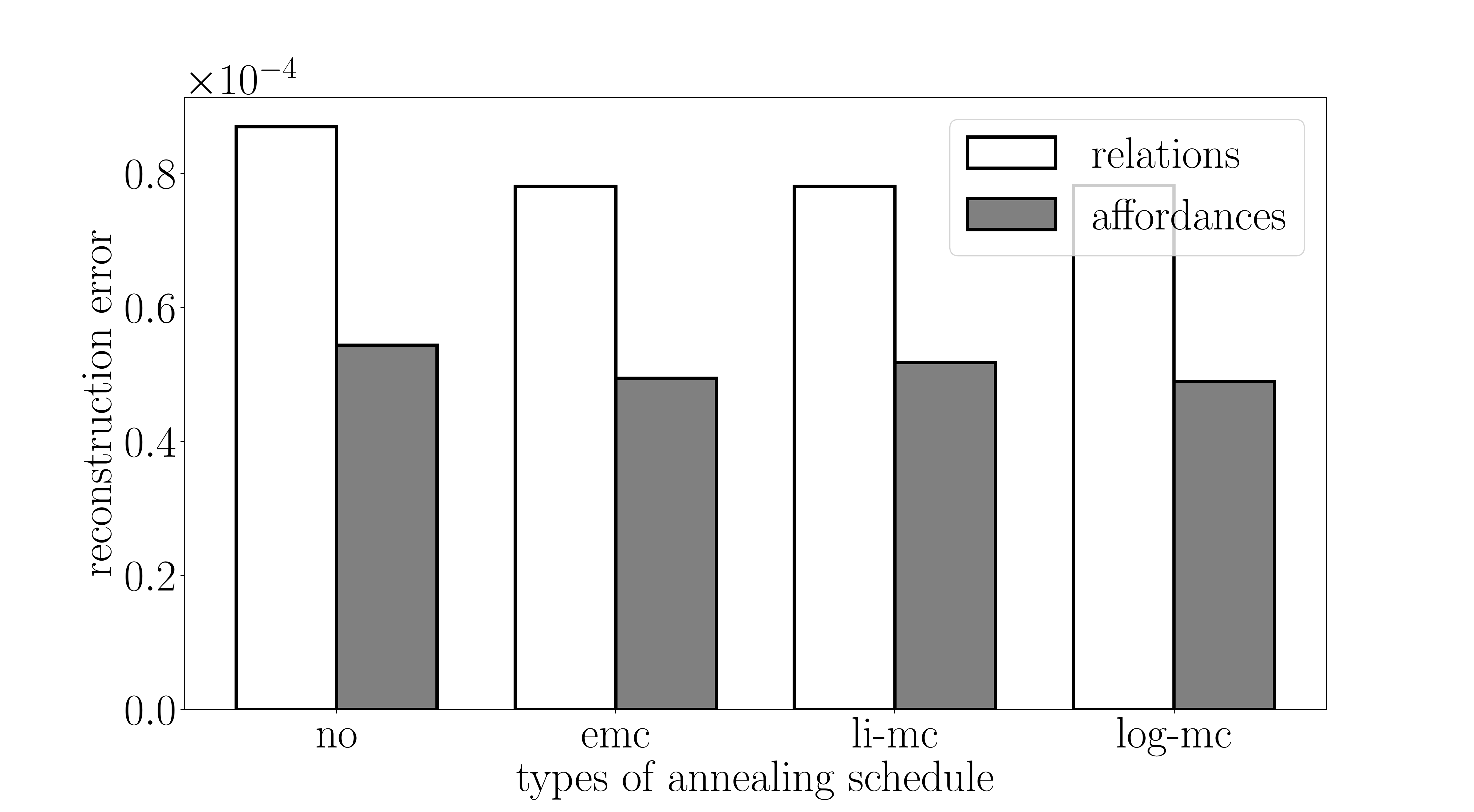}
    }
}
\caption{Reconstruction errors (after 30 epochs) for different annealing schedules with initial temperature 4.0. (a) Reconstruction error for object nodes. (b) Reconstruction error for relation and affordance nodes.\label{fig:error_vs_temperature}}
\end{figure}

Lastly, we analyzed the effect of different annealing schedules. We tried the following annealing schedules (selected from \cite{nourani1998comparison}), namely, exponential multiplicative cooling (emc, Equation \ref{eqn:emc}), linear multiplicative cooling (li-mc, Equation \ref{eqn:limc}) and logarithmic multiplicative cooling (log-mc, Equation \ref{eqn:logmc}), with initial temperature ($T_0$) set to $4.0$:
\begin{align}
T_i & = {T_0 \cdot a^i}, &  (0.8 \leq a \leq 0.9) 	\label{eqn:emc} \\ 
T_i & = \frac{T_0}{1+a\times i}, &  (a > 0) 	\label{eqn:limc}  \\ 
T_i & = \frac{T_0}{1+a\log(1+i)}, & (a > 1) \label{eqn:logmc} 
\end{align}
As shown in Figure \ref{fig:error_vs_temperature}, although annealing schedules affect reconstruction errors all types of nodes, differences between them are observed to be not significant.

In summary, our analysis suggests that COSMO with one hidden layer, with 400 hidden nodes (although, as shown in Figure \ref{fig:error_vs_hiddens}, 800 or more hidden nodes provide better performance, the performance gain is insignificant compared to the computational overload) and emc annealing performs best.  Therefore, in the rest of the paper, we adopted these settings for COSMO. For RBM and GBM, we used the same number of hidden nodes and layers as COSMO. 

\subsection{Comparison Measures}
\label{sect:measures}

For evaluating the performance of the methods, we use precision, recall and F1-score which are defined as follows:
\begin{eqnarray}
\textrm{Precision} & = & \frac{\textrm{TP}}{\textrm{TP}+\textrm{FP}},\\
\textrm{Recall} & = & \frac{\textrm{TP}}{\textrm{TP}+\textrm{FN}},\\
\textrm{F1-score} & = & 2\cdot \frac{\textrm{Precision}\cdot \textrm{Recall}}{\textrm{Precision} + \textrm{Recall}},
\end{eqnarray}
{\noindent}where TP, FP and FN stand for the number of true positives, false positives and false negatives, respectively. The definitions of TP, FP and FN are task-dependent, and therefore, they are defined for each task separately.

\subsection{Task 1: Spatial Relation Estimation}
Being generative, COSMO can estimate the relations in the scene given the  objects present in the scene. Contextual information that arises from active objects, regardless of spatial relation and affordance nodes, helps the model to determine which spatial relations should be active according to the context.

For testing, initially, the model sees the environment in a ``bag of objects'' sense by setting the \NEW{affordances and} objects to the visible nodes, and relation nodes are set to zero. Next, the hidden nodes (i.e., context) are sampled  using object nodes only. Then, the spatial relation nodes are sampled from objects, affordances and the context. This procedure is summarized in Algorithm \ref{alg:task1}.

\begin{algorithm}
\caption{Algorithm used for the relation estimation task (Task 1).}\label{alg:task1}
\begin{algorithmic}[1]
\State \textbf{Input:} A scene \textbf{s}; the number of Gibbs steps, $k$.
\State \textbf{Output:} Relation node activations, $\mathbf{r}$.
\State
\State $\mathbf{r} \gets 0$. \Comment{Set relation node activations to 0.}
\For{$k$ sampling steps}
\State $\mathbf{o} \gets \mathbf{s_o}$, $\mathbf{a} \gets \mathbf{s_a}$ \Comment{Clamp objects and affordances.}

\State Sample hidden nodes \textbf{h}  using Eq. \ref{eqn:ph}.
\State Sample relation nodes \textbf{r}  using Eq. \ref{eqn:pr}.
\EndFor
\end{algorithmic}
\end{algorithm}

For this task, we define True Positives (TP) as the number of spatial relation nodes which are active in both the input scene ($\mathbf{s}$) and the reconstructed scene ($\mathbf{s^{\prime}}$); True Negatives (TN) as the number of spatial relation nodes which are both in-active in $\mathbf{s}$ and $\mathbf{s^{\prime}}$; False Positives (FP) as the number of spatial relation nodes which are inactive in $\mathbf{s}$ but active in $\mathbf{s^{\prime}}$; False Negatives (FN) as the number of spatial relation nodes which are active in $\mathbf{s}$ and in-active in $\mathbf{s^{\prime}}$. These are defined formally as follows:
\begin{eqnarray}
TP & = & \left\vert \{x: x\in G^{+}_r \land x\in M^{+}_r  \} \right\vert , \\ 
TN & = & \left\vert \{x: x\in G^{-}_r \land x\in M^{-}_r  \} \right\vert , \\
FP & = & \left\vert \{x: x\in G^{-}_r \land x\in M^{+}_r  \} \right\vert , \\
FN & = & \left\vert \{x: x\in G^{+}_r \land x\in M^{-}_r  \} \right\vert ,
\end{eqnarray}
{\noindent}where $x$ is a relation node; $G^{+}_r$ and $G^{-}_r$ are respectively the sets of active and passive relation nodes in the sample; and $M^{+}_r$ and $M^{-}_r$ are the sets of active and passive relation nodes respectively at the end of the model's reconstruction. 

Table \ref{tbl:spatial_rel_estimation} lists the performance of COSMO for this task and compares it against RBM, GBM, and RN. In comparison to the other models, we see that COSMO provides the best performance. \SEC{To estimate the chance level, we (i) randomly activate TP+FN many relation nodes (TP+FN is the number of correct instances in the ground truth), (ii) calculate TP, FP, FN, TN, precision, recall and F1-score measures, and (iii) repeat this for 100 many instances to obtain an estimation. Note that, with this scheme, the FP and FN rates are equal, and therefore, we obtain the same values for precision, recall and F1-score.}

\begin{figure}
\centerline{  
	\subfigure[]{
    	\label{fig:task1a}
    	\includegraphics[width=0.45\textwidth]{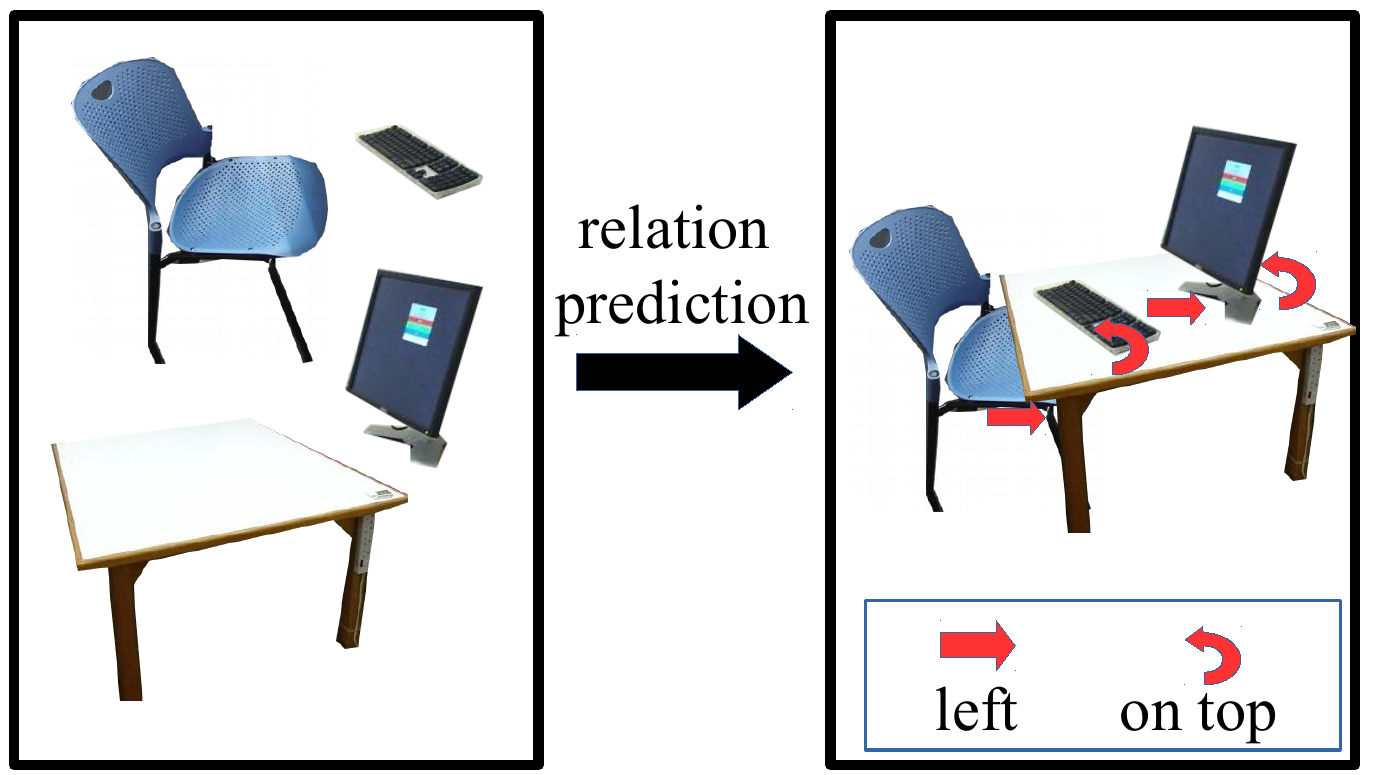}
        }
	\subfigure[]{
    	\label{fig:task1b}
    	\includegraphics[width=0.45\textwidth]{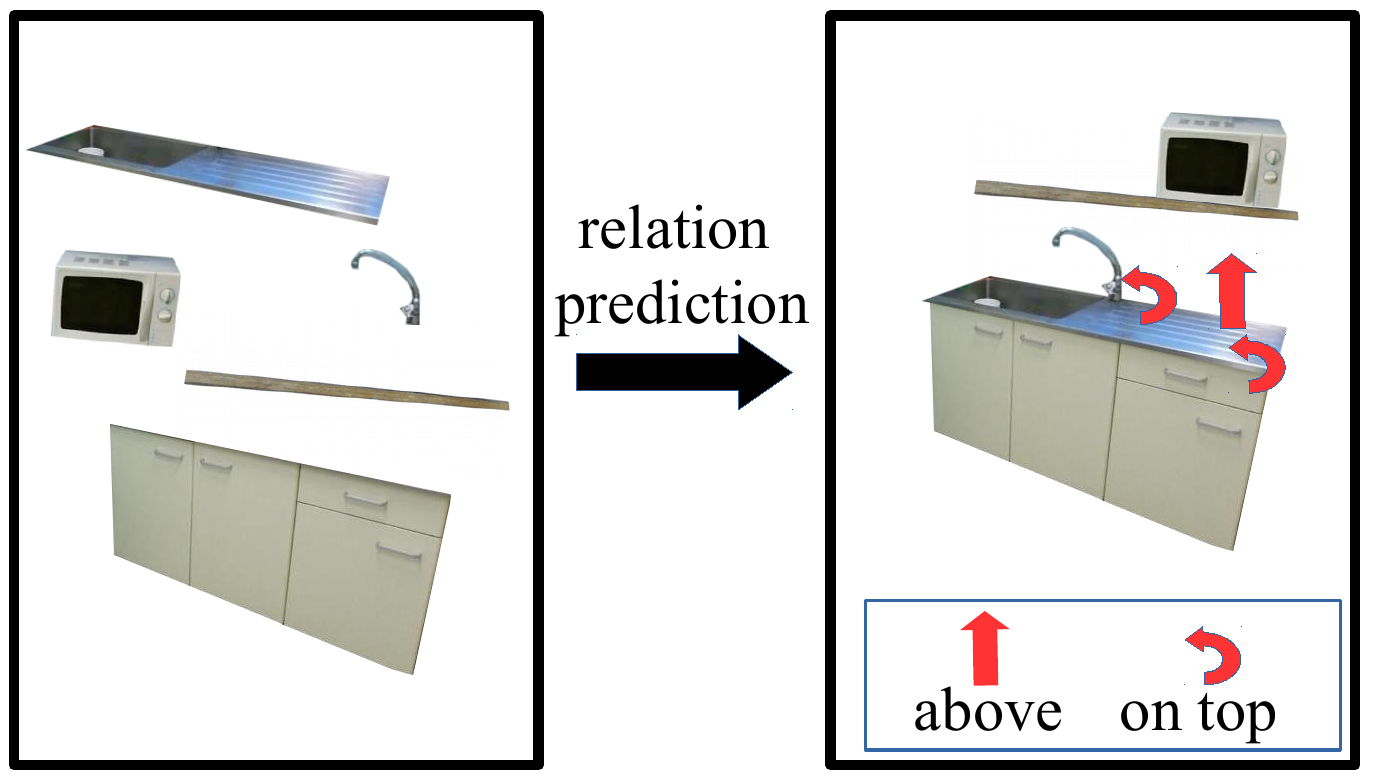}
        }
}
\caption{Some example relations estimated by COSMO for given sets of objects  (Task 1). Only a subset of the relations are shown for the sake of visibility. \label{fig:spatial_rel_estimation}}
\end{figure}

Moreover, we provide some visual examples in Figure \ref{fig:spatial_rel_estimation}, where we see that our model can discover spatial relations between objects. This means that, given a set of objects, COSMO can determine how to roughly place them in a scene.

\begin{table}[hbt]
\centering
\caption{Task 1 (Spatial Relation Estimation) performances. \label{tbl:spatial_rel_estimation}}

\footnotesize
\begin{tabular}{|l|ccc|} \hline                                                              
&\textbf{Precision}  &  \textbf{Recall}   & \textbf{F1-score}  \\ \hline \hline
COSMO & 0.1511  & \textbf{0.3112} & \textbf{0.2034}    \\ 
GBM & \textbf{0.1559} & 0.1125 & 0.1307  \\ 
RBM & 0.0043 & 0.0132 & 0.0066  \\ 
RN  & 0.0166 & 0.0132 & 0.0147   \\
\SEC{Chance level}  & \SEC{$9.9 \times 10^{-5}$} & \SEC{$9.9 \times 10^{-5}$}& \SEC{$ 9.9 \times 10^{-5}$}   \\\hline
\end{tabular}
\end{table}

In some cases, naturally, the ``bag of objects'' approach may not provide enough contextual information in order for the model to predict ground truth spatial relationships in the test set. For instance, consider a scene consisting of \textit{plate, table, cabinet} objects. In a ``kitchen with eating'' context, the \textit{plate} can be \textit{on} the \textit{table}, whereas, in a ``kitchen without eating'' context, the \textit{plate} is likely to be \textit{in} the \textit{cabinet}. These cases can reduce testing accuracy for the estimated relation between objects like a \textit{plate}. However, given such examples during training, COSMO is able to capture the probability of all these cases and therefore handle scene modeling tasks in such settings accordingly.

\subsection{Task 2: What is missing in the scene?}

In this task, COSMO predicts missing objects in the scene according to the current context. The model is provided ``partially observed scenes'' where some of the objects are removed randomly for testing. 

Firstly, observed objects, spatial relations and affordances are clamped to the visible units, then the model is relaxed to find the hidden node activations (i.e. the context of the scene). Finally, by using the visible (scene description) and hidden (context) node activations, the network tries to find the missing objects in the scene as outlined in Algorithm \ref{alg:task2}.

For this task, we define TP as the number of object nodes that are activated correctly according to the ground truth sample; FP as the number of object nodes that the model activates but should have been deactivated according to the ground truth; TN as the number of object nodes that are deactivated correctly according to ground truth, and FN as the number of objects that the model deactivates, yet  should have been activated according to the ground truth. We can formally define these as follows: 
\begin{eqnarray}
TP & = & \left\vert \{x: x\in G^{+}_o \land x\in M^{+}_o  \} \right\vert , \label{eqn:task2hypothesiseqns} \\
TN & = & \left\vert \{x: x\in G^{-}_o \land x\in M^{-}_o  \} \right\vert , \nonumber \\
FP & = & \left\vert \{x: x\in G^{-}_o \land x\in M^{+}_o  \} \right\vert , \nonumber \\
FN & = & \left\vert \{x: x\in G^{+}_o \land x\in M^{-}_o  \} \right\vert \nonumber ,
\end{eqnarray}
{\noindent}where $x$ is an object node; $G^{+}_o$ and $G^{-}_o$ are the sets of active and passive object nodes respectively in the ground truth sample; and $M^{+}_o$ and $M^{-}_o$ are the sets of active and passive object nodes respectively at the end of the model's reconstruction.

\begin{figure}[hbt!]
\centerline{
	\subfigure[]{
    	\label{fig:task2a}
    	\includegraphics[width=0.49\textwidth]{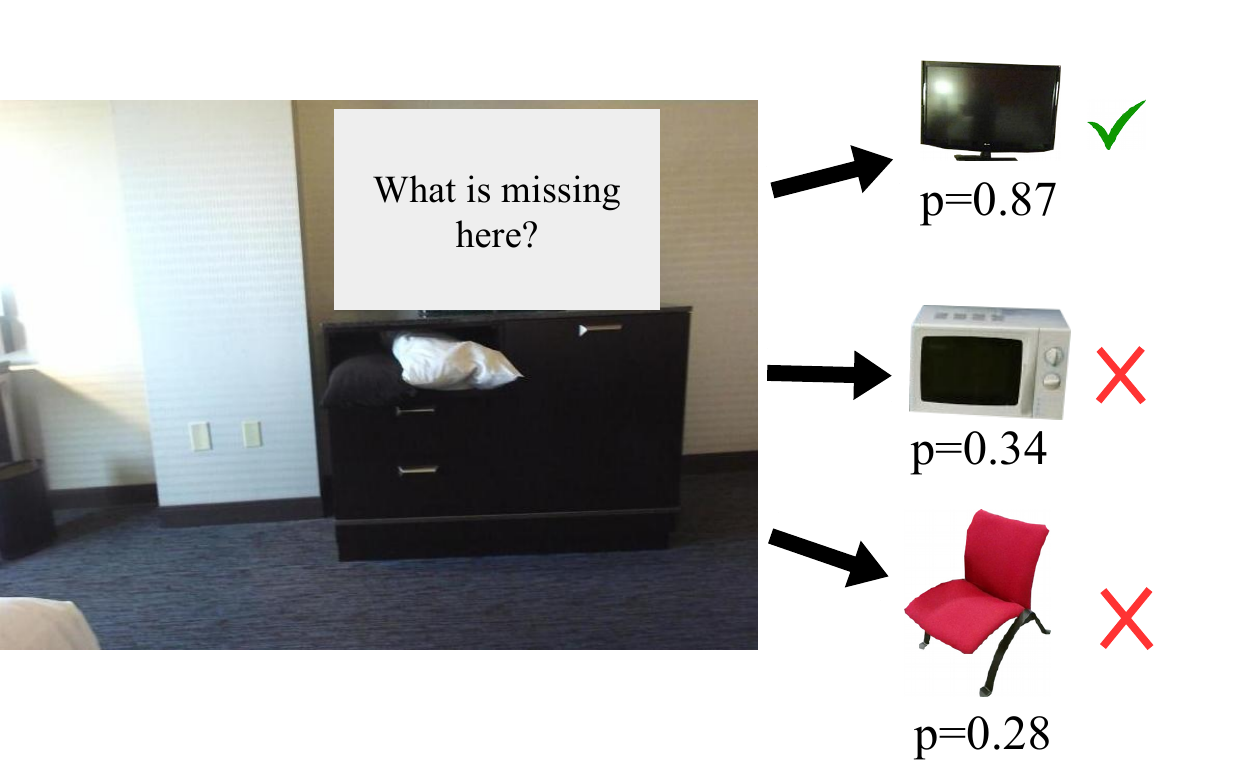}
        }
	\subfigure[]{
    	\label{fig:task2b}
    	\includegraphics[width=0.49\textwidth]{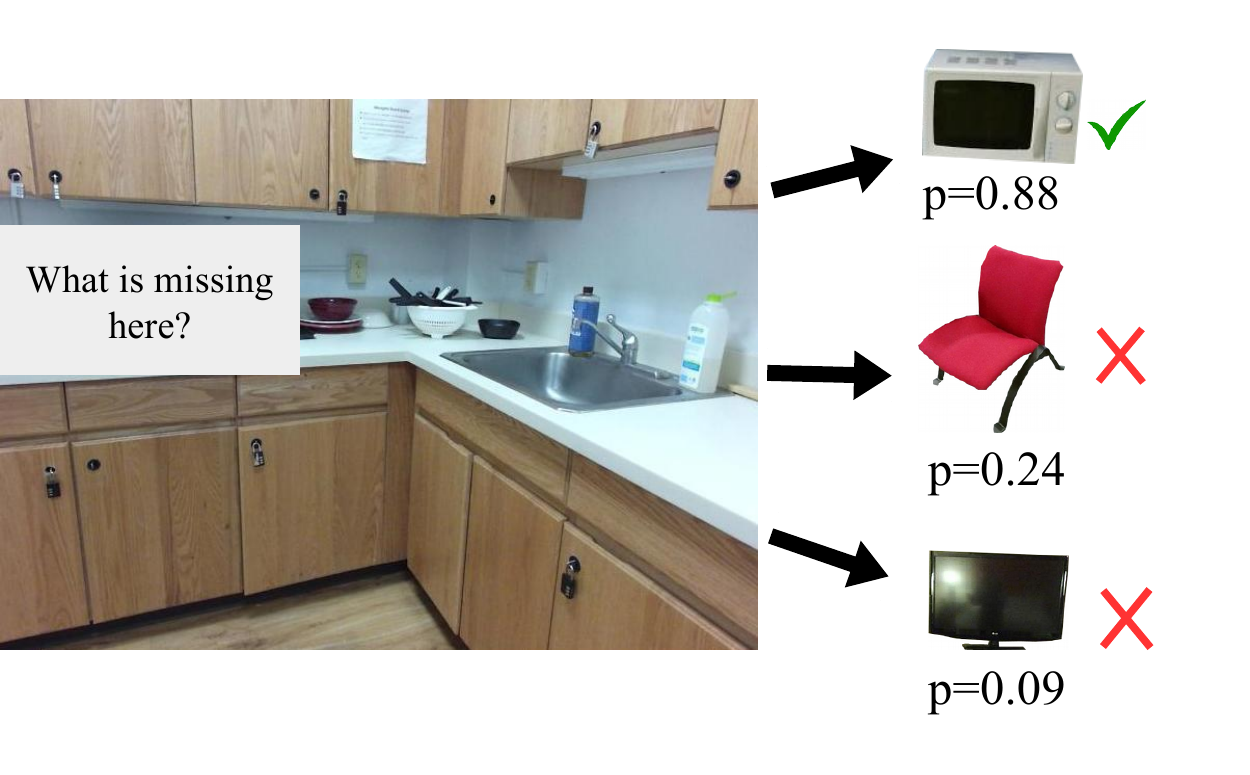}
    }
}
\caption{Some examples illustrating the performance of COSMO on finding a missing object in a scene  (Task 2). \label{fig:whatismissing}}
\end{figure}

As shown in Table \ref{tbl:whatismissing}, our model performs better than RBM, GBM and RN. See also Figure \ref{fig:whatismissing}, which shows some visual examples for most likely objects found for a target position in the scene. \SEC{We estimate the chance level similar to the one we did for relation estimation: We (i) randomly activate TP+FN many object nodes (TP+FN is the number of correct instances in the ground truth), (ii) calculate TP, FP, FN, TN, precision, recall and F1-score measures, and (iii) repeat this for 100 many instances to obtain an estimation. Note again that, with this scheme, the FP and FN rates are equal, and therefore, we obtain the same values for precision, recall and F1-score.}

%
%

\begin{algorithm}
\caption{The algorithm for finding missing objects (Task 2).}\label{alg:task2}
\begin{algorithmic}[1]
\State \textbf{Input:} A scene, $\mathbf{s}$; the number of Gibbs steps, $k$.
\State \textbf{Output:} Initially in-active object nodes in $\mathbf{s}$, $\mathbf{o^{{\prime}}}$.
\State
\For{ $k$ sampling steps }
\State $\mathbf{o} \gets \mathbf{s_o}$; $\mathbf{r} \gets \mathbf{s_r}$; $\mathbf{a} \gets \mathbf{s_a}$ \Comment{Clamp input scene.}
\State Sample hidden nodes $\mathbf{h}$  using Eq. \ref{eqn:ph}.
\State Sample \textit{in-active} object nodes $\mathbf{o^{{\prime}}}$  using Eq. \ref{eqn:po}.
\EndFor
\end{algorithmic}
\end{algorithm}

\begin{figure}[hbt!]
\centering{
	\subfigure[]{
    	\label{fig:supplementary1}
    	\includegraphics[width=0.3\textwidth]{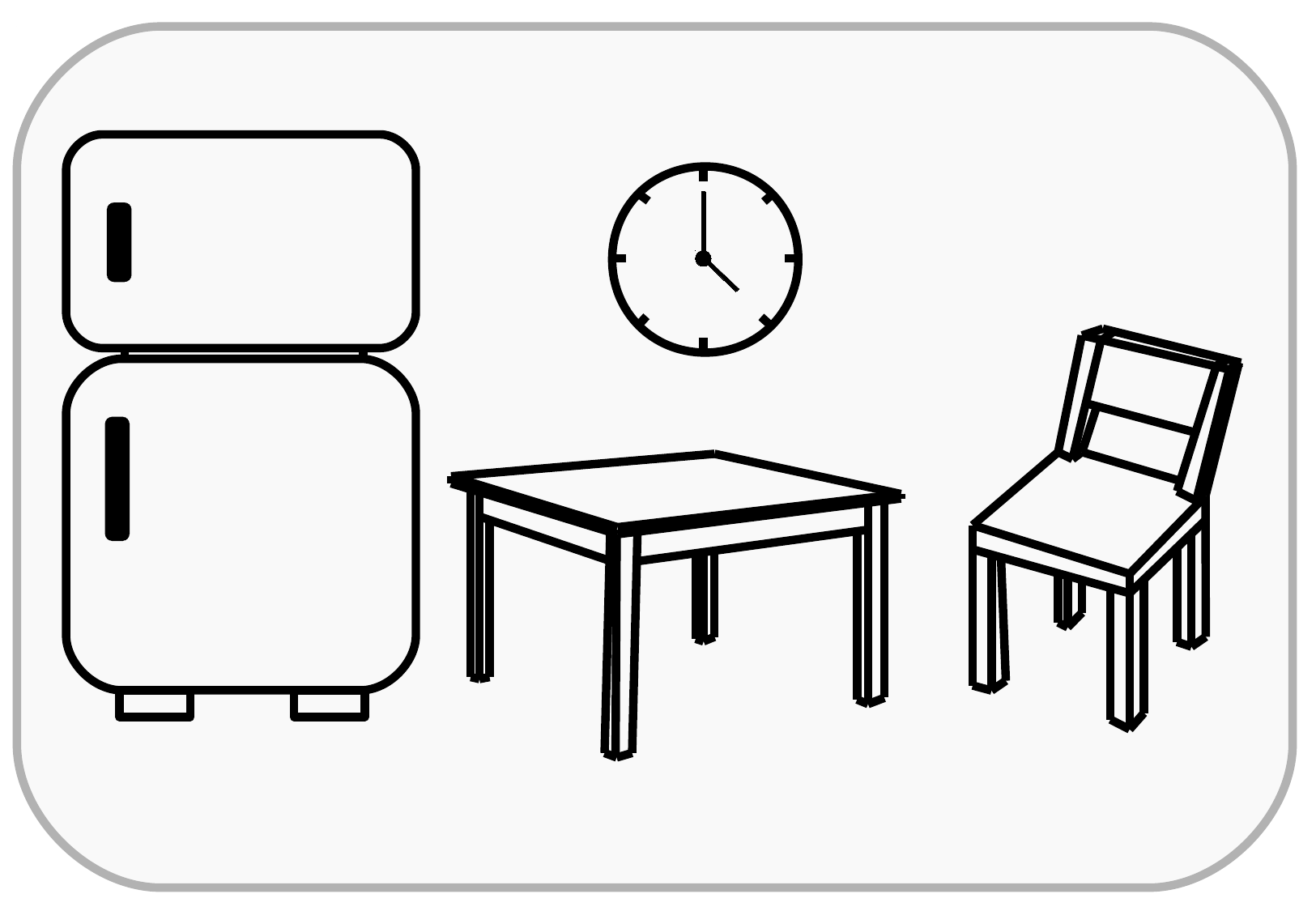}
        }
	\subfigure[]{
    	\label{fig:supplementary2}
    	\includegraphics[width=0.3\textwidth]{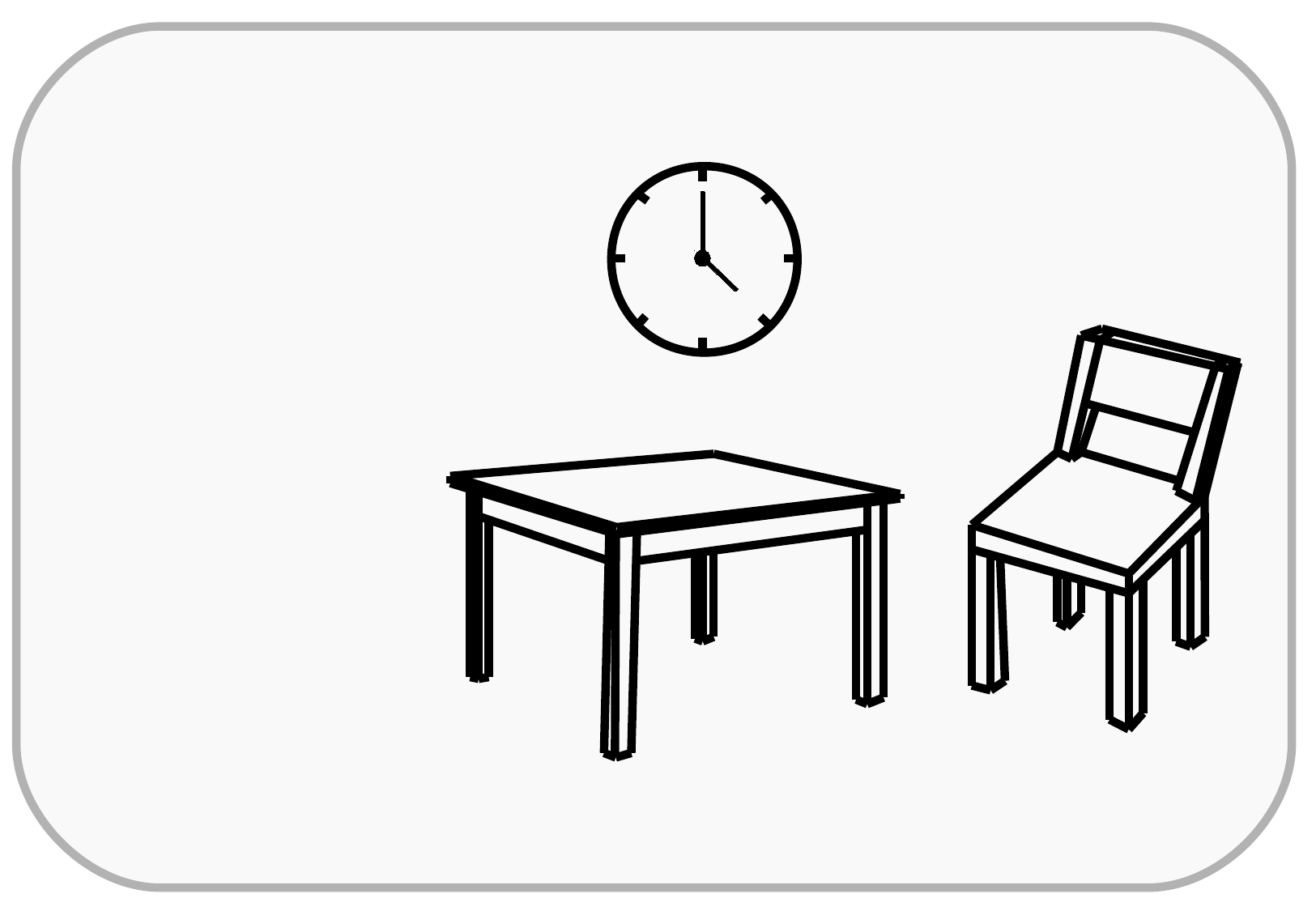}
    }
    \subfigure[]{
    	\label{fig:supplementary3}
    	\includegraphics[width=0.3\textwidth]{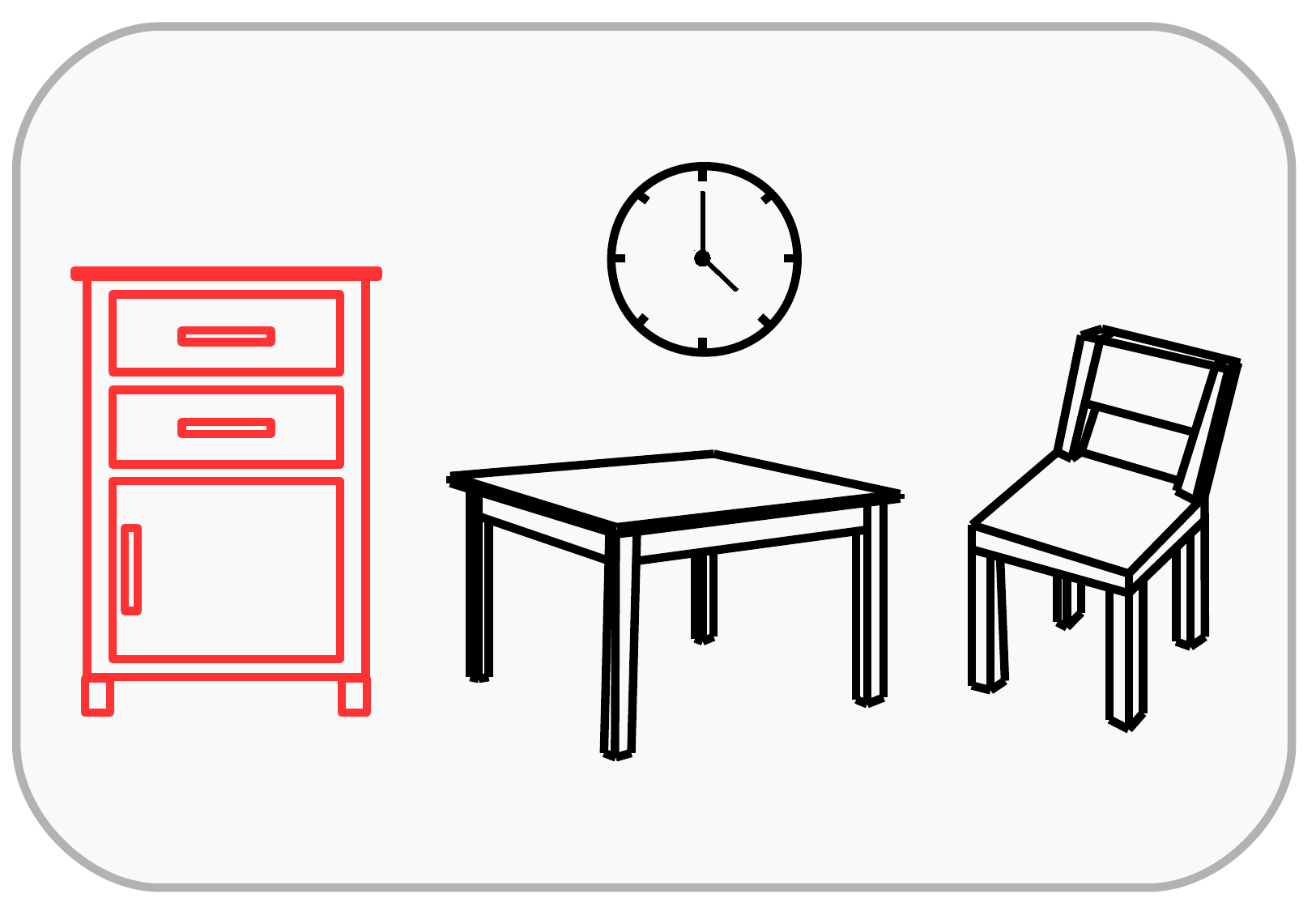}
    }
}
\caption{\NEW{COSMO and other models might not find removed objects due to contextual bias. (a) Ground truth test sample consists of a fridge, a table, a chair and a clock. (b) The fridge is removed from the scene and COSMO is run to find missing object in the scene. (c) COSMO activates the cabinet object instead, due to contextual bias. As far as context is concerned, the reconstructed scene is compatible with the dataset.}\label{fig:supplementary}}
\end{figure}

\begin{figure}[hbt!]
\centering{
	\subfigure[]{
    	\label{fig:supplementary4}
    	\includegraphics[width=0.3\textwidth]{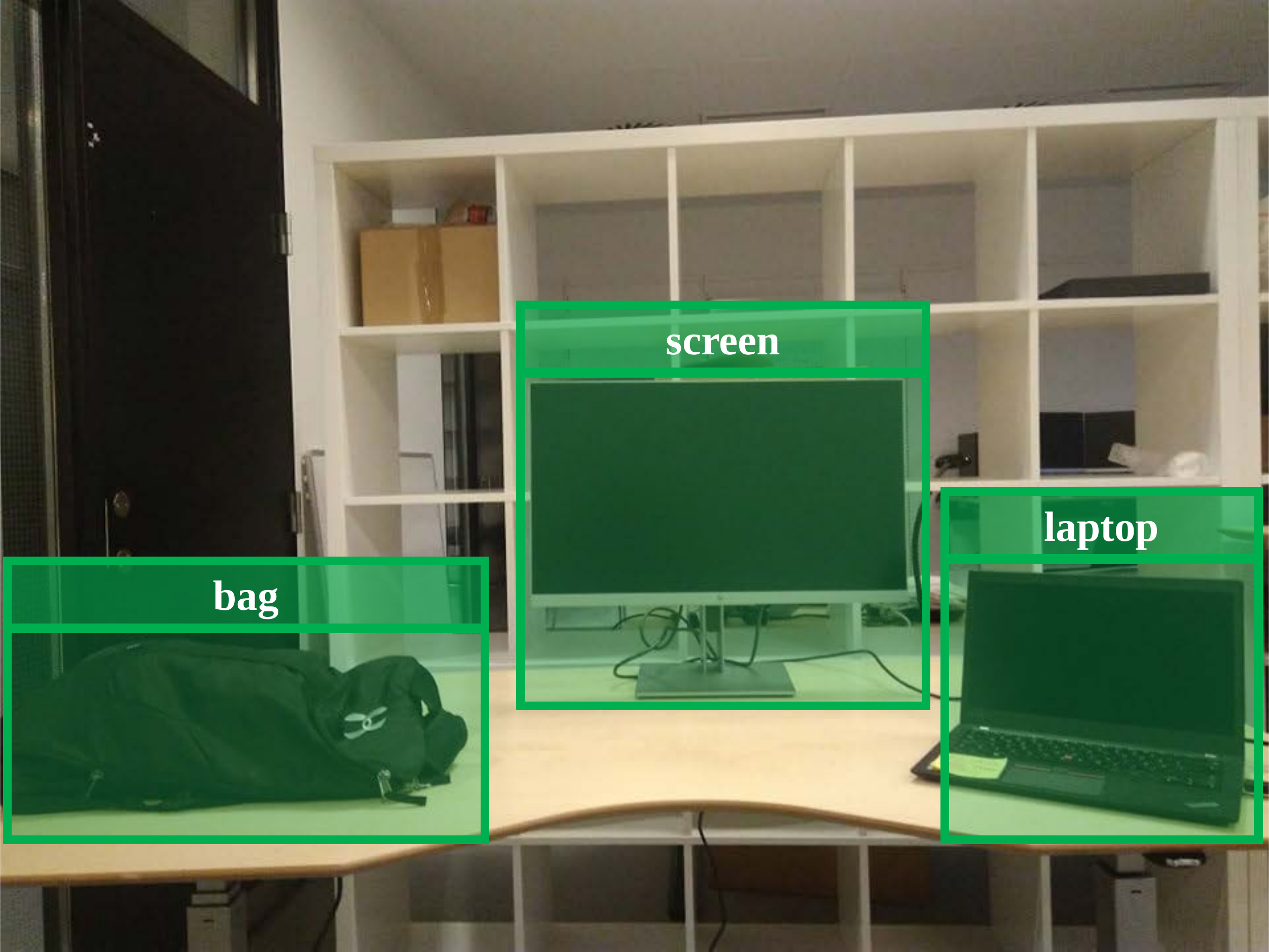}
        }
	\subfigure[]{
    	\label{fig:supplementary5}
    	\includegraphics[width=0.3\textwidth]{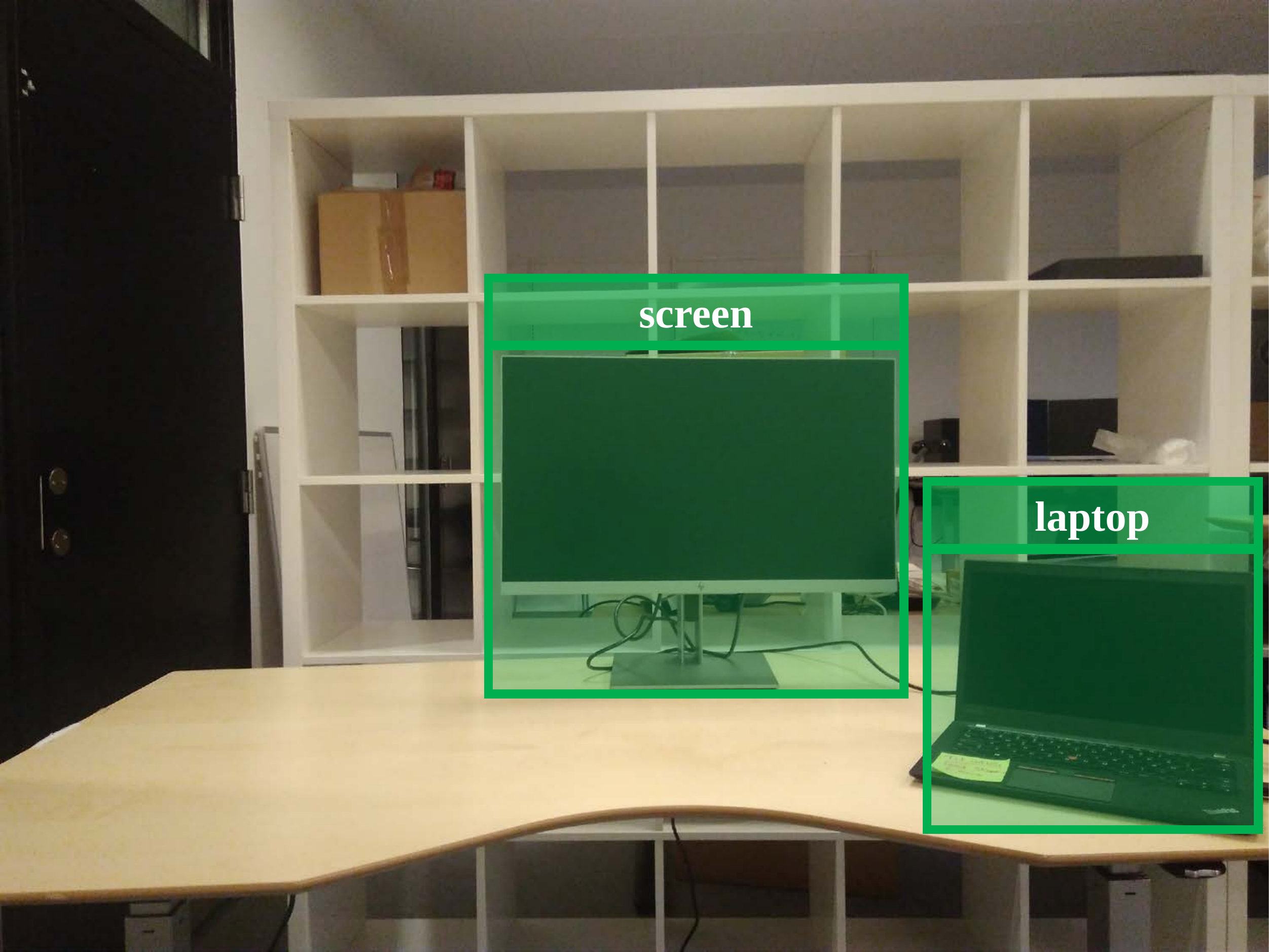}
    }
    \subfigure[]{
    	\label{fig:supplementary6}
    	\includegraphics[width=0.3\textwidth]{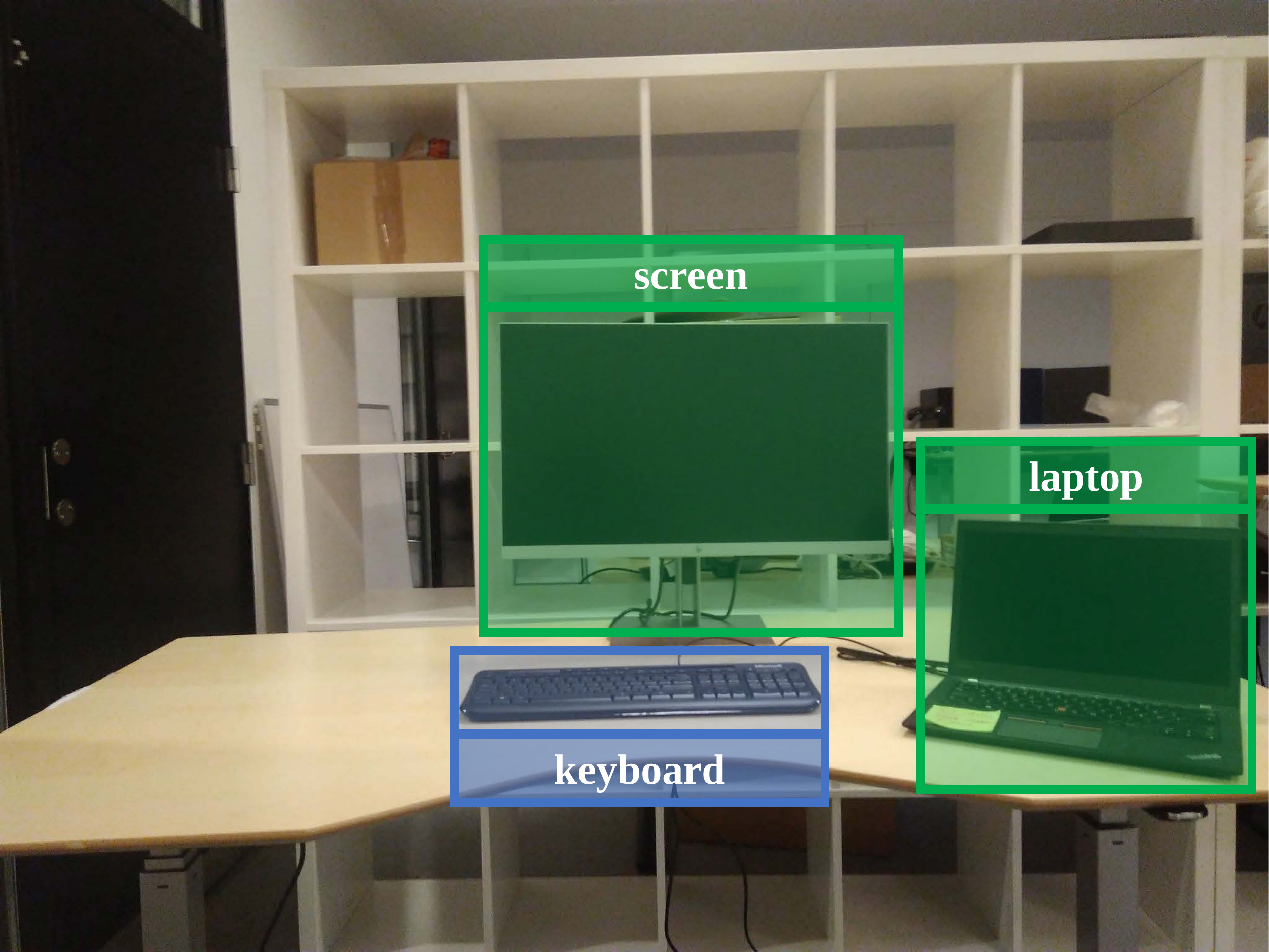}
    }
}
\caption{\SEC{An illustration of an actual ambiguous prediction due to contextual bias.  (a) Ground truth test sample consists of a bag, a monitor, and a laptop. (b) The bag is removed from the scene and COSMO is run to find the missing object in the scene. (c) COSMO activates the keyboard object instead, due to the contextual bias. }\label{fig:supplementary_2}}
\end{figure}

\NEW{The recall rates for this task are rather low because of a phenomenon that we may call contextual bias. Removing objects from a scene can change the current context, or removed objects may not be recalled if they have low contextual importance -- see Figure \ref{fig:supplementary} \SEC{and \ref{fig:supplementary_2}} for an illustration. In these cases, COSMO and the other models may not figure out the removed object, which counts towards a low recall rate.}

\begin{table}[hbt]
\caption{Task 2 (finding missing objects) performances. \label{tbl:whatismissing}}
\centering
\footnotesize
\begin{tabular}{|l|ccc|}\hline                                                                 
&\textbf{Precision}  & \textbf{Recall}  & \textbf{F1-score}  \\ \hline \hline
COSMO & \textbf{0.9387}	& \textbf{0.0527}   & \textbf{0.0998}   \\ 
GBM & 0.8260 & 0.0415  &  0.0790  \\ 
RBM & 0.7250 & 0.0301  &  0.0578   \\ 
RN  & 0.8000	 &  0.0212   &  0.0414 \\
\SEC{Chance level}  & \SEC{0.0056} & \SEC{0.0056}& \SEC{0.0056} \\ \hline
  
\end{tabular}
\end{table}


\subsection{Task 3: What is extra in the scene?}

In this task, COSMO predicts objects that are out of context in the scene. For this purpose, objects are randomly selected and added to the original scene for testing. 
 
Firstly, the observed objects, spatial relations and affordances are clamped to the visible units, then the model is relaxed to find hidden node activations (i.e. the context of the scene). Finally, by using the visible node and hidden (context) node activations, the network tries to remove objects that are out of context in the scene as outlined in Algorithm \ref{alg:task3}.

For this task, we use the TP, TN, FP and FN as defined in Equation \ref{eqn:task2hypothesiseqns}.

\begin{figure}[hbt!]
\centerline{
	\subfigure[]{
    	\label{fig:task3a}
    	\includegraphics[width=0.49\textwidth]{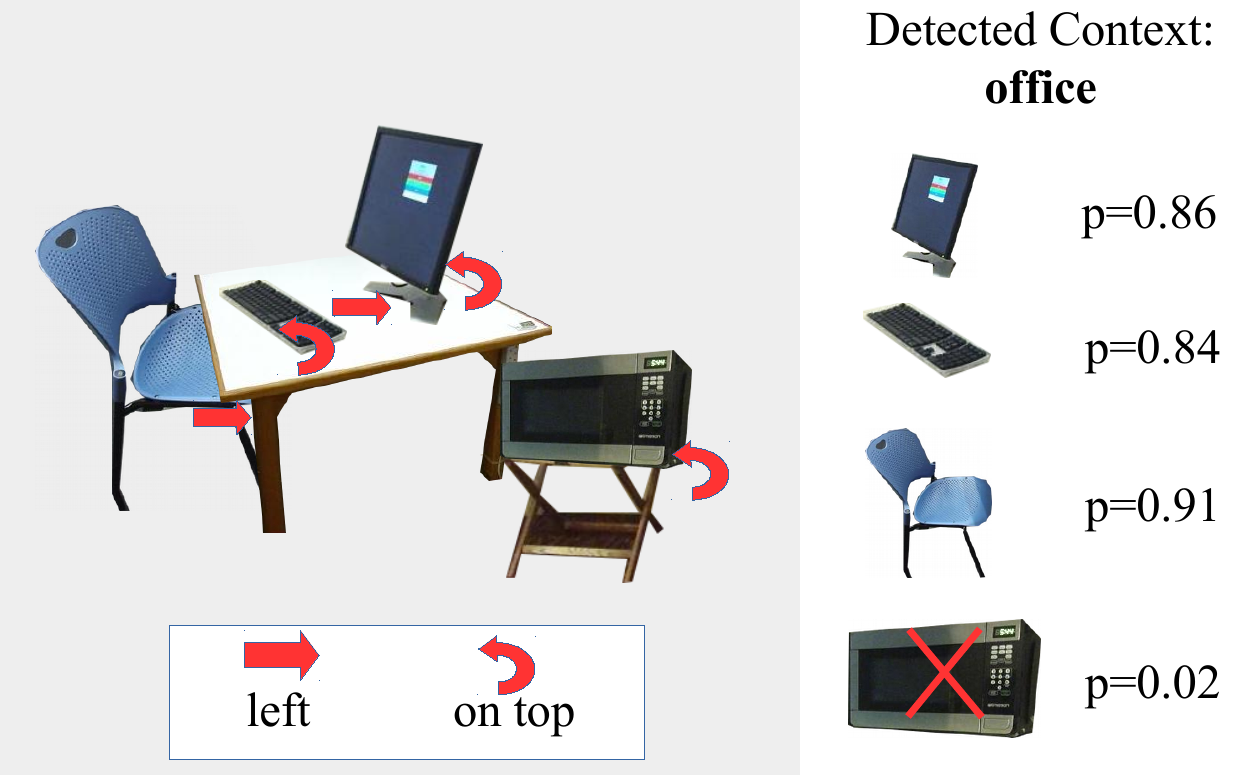}
        }
	\subfigure[]{
    	\label{fig:task3b}
    	\includegraphics[width=0.49\textwidth]{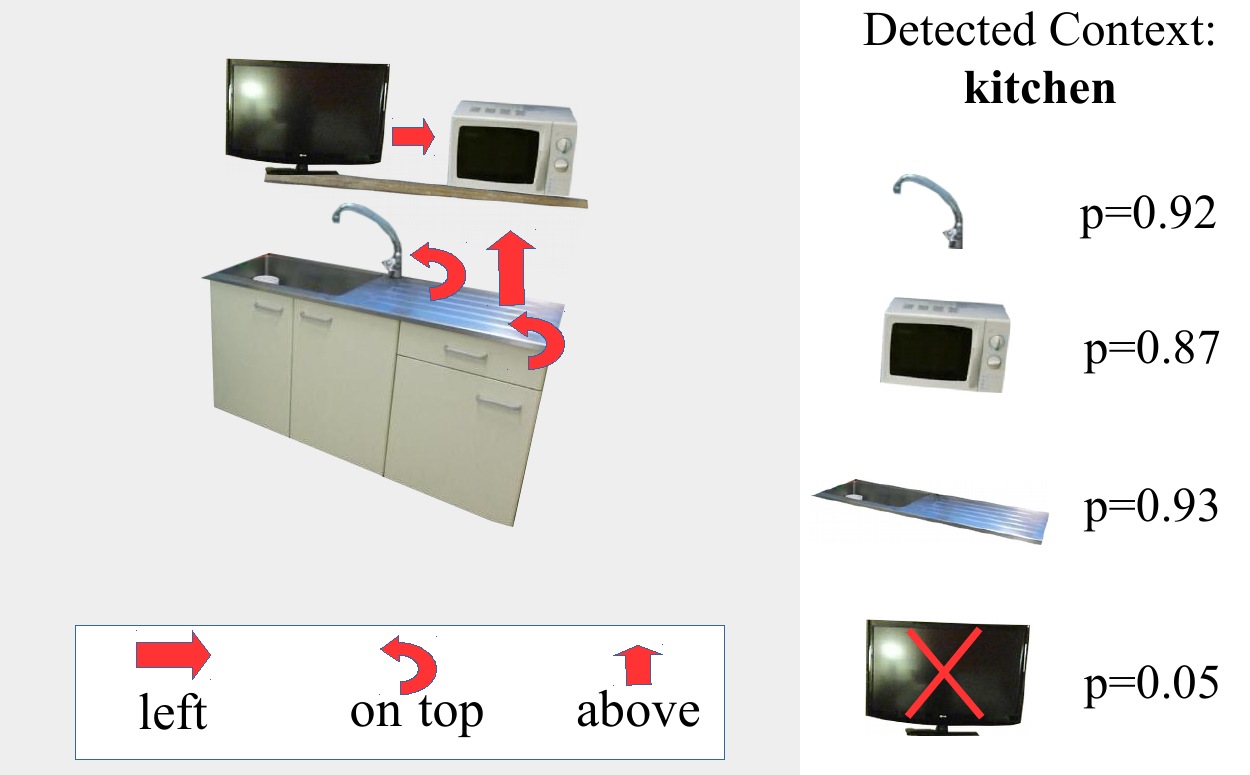}
    }
}
\caption{Some examples illustrating the performance of COSMO on finding the out-of-context object in a scene  (Task 3). \label{fig:whatisextra}}
\end{figure}

As shown in Table \ref{tbl:whatisextra}, our model performs better than RBM, GBM and RN for finding the extra objects in the scene.  See also Figure \ref{fig:whatisextra}, which shows some visual examples for finding the objects that are out of context in the scene. \SEC{To estimate the chance levels, similar to Task 2, we randomly activate object nodes and calculate average precision, recall and F1-score over 100 experiments.} 

\begin{table}[hbt]
\caption{Task 3 (finding extra objects) performances. \label{tbl:whatisextra}}
\centering
\footnotesize
\begin{tabular}{|l|ccc|}\hline                                                                 
&\textbf{Precision}  & \textbf{Recall}  & \textbf{F1-score}  \\ \hline \hline
COSMO & \textbf{0.9183}	& \textbf{0.0482}   & \textbf{0.0917}   \\ 
GBM & 0.8113 & 0.0479  &  0.0865  \\ 
RBM & 0.7826 & 0.0382  &  0.0729   \\ 
RN  & 0.7368	 &  0.0297   &  0.0572 \\ \SEC{Chance level}  & \SEC{0.0056} & \SEC{0.0056}& \SEC{0.0056} \\ \hline
  
\end{tabular}
\end{table}

\begin{algorithm}[hbt!]
\caption{The algorithm for finding the out-of-context object (Task 3).} \label{alg:task3}
\begin{algorithmic}[1]
\State \textbf{Input:} A scene, $\mathbf{s}$; the number of Gibbs steps, $k$.
\State \textbf{Output:} Initially active object nodes in $\mathbf{s}$, $\mathbf{o^{{\prime}}}$.
\State
\For{ $k$ sampling steps }
\State $\mathbf{o} \gets \mathbf{s_o}$; $\mathbf{r} \gets \mathbf{s_r}$; $\mathbf{a} \gets \mathbf{s_a}$ \Comment{Clamp input scene.}
\State Sample hidden nodes $\mathbf{h}$  using Eq. \ref{eqn:ph}.
\State Sample \textit{active} object nodes $\mathbf{o^{{\prime}}}$  using Eq. \ref{eqn:po}.
\EndFor
\end{algorithmic}
\end{algorithm}

\subsection{Task 4: Affordance Prediction} 

Affordances of objects may differ for different subjects in different contexts \cite{tucker1998relations}. Therefore, agents should be aware of the context that they are in, in order to reason about the affordances of the objects. We show that COSMO can allow agents to determine the affordances of objects using the current context.

For this task, firstly, the present objects and relations are clamped to the visible nodes, and the hidden nodes (i.e. context) are sampled. Then, the affordance nodes are sampled  from the hidden nodes (context), the object nodes and the relation nodes, as illustrated in Algorithm \ref{alg:task4}.

\begin{algorithm}[hbt!]
\caption{Algorithm for the affordance prediction task (Task 4).}\label{alg:task4}
\begin{algorithmic}[1]
\State \textbf{Input:} A scene, \textbf{s}; the number of Gibbs steps, $k$.
\State \textbf{Output:} Affordance node activations, $\mathbf{a}$.
\State
\State $\mathbf{a} \gets 0$. \Comment{Set affordance nodes to 0.}
\For{ $k$ sampling steps }
\State $\mathbf{o} \gets \mathbf{s_o}$, $\mathbf{r} \gets \mathbf{s_r}$ \Comment{Clamp objects and relations.}
\State Sample hidden nodes $\mathbf{h}$  using Eq. \ref{eqn:ph}.
\State Sample affordance nodes $\mathbf{a}$ using Eq. \ref{eqn:pa}.
\EndFor
\end{algorithmic}
\end{algorithm}

For this task, we define TP as the number of affordance nodes that are activated correctly according to the ground truth sample; FP as the number of affordance nodes that the model activates but should have been deactivated according to the ground truth; TN as the number of affordance nodes that are deactivated correctly according to ground truth; and FN as the number of affordance nodes that the model deactivates yet should have been activated according to the ground truth. We define them formally as follows:
\begin{eqnarray}
TP = \left\vert \{x: x\in G^{+}_a \land x\in M^{+}_a  \} \right\vert , \label{eqn:task4hypothesiseqns}\\
TN = \left\vert \{x: x\in G^{-}_a \land x\in M^{-}_a  \} \right\vert \nonumber , \\
FP = \left\vert \{x: x\in G^{-}_a \land x\in M^{+}_a  \} \right\vert \nonumber , \\
FN = \left\vert \{x: x\in G^{+}_a \land x\in M^{-}_a  \} \right\vert \nonumber ,
\end{eqnarray}
{\noindent}where $x$ is an affordance node; $G^{+}_a$ and $G^{-}_a$ are the sets of active and passive affordance nodes respectively in the ground truth sample; and $M^{+}_a$ and $M^{-}_a$ are the sets of active and passive affordance nodes respectively at the end of model's reconstruction.

As shown in Table \ref{tbl:affordance_estimation_f1_table}, our model performs better than RBM, GBM and RN. See also Figure \ref{fig:taskaffordance_2}, which shows some visual examples for affordance prediction for different objects. \SEC{For estimating the chance levels, similar to the previous tasks, we randomly activate affordance nodes and calculate average precision, recall and F1-score over 100 experiments.}

%
%

\begin{table}[hbt]
\caption{Task 4 (affordance prediction) performances. \label{tbl:affordance_estimation_f1_table}}
\centering
\footnotesize
\begin{tabular}{|l|ccc|}\hline                                                                 
&\textbf{Precision}  & \textbf{Recall}  & \textbf{F1-score}  \\ \hline \hline
COSMO & \textbf{0.2039}  & \textbf{0.3129} & \textbf{0.2469}    \\ 
GBM & 0.1372 & 0.1068 & 0.1201  \\ 
RBM & 0.0769 & 0.0076 & 0.0138   \\
RN & 0.0125 & 0.0091 & 0.0105   \\ 
\SEC{Chance level}  & \SEC{$1.0 \times 10^{-5}$} & \SEC{$1.0 \times 10^{-5}$}& \SEC{$1.0 \times 10^{-5}$} \\ \hline
\end{tabular}
\end{table}

\subsection{Task 5: Objects affording an action}

Being generative, COSMO allows reasoning about object affordances in various ways. In this task, we evaluate the methods on finding objects that afford a certain action. For this end, some of the object nodes, which are the object part of an affordance-triplet, and the corresponding affordance nodes are deactivated. Then, the model samples the hidden nodes using the partially observed scene. In the reconstruction phase, the model samples back the deactivated objects and affordance nodes that include these objects using the context and the observed scene. This is formalized in Algorithm \ref{alg:task5}.

For this task, we use the same definitions of TP, FP, TN and FN in Equation \ref{eqn:task4hypothesiseqns}. However, in this task, $G^{+}_a$, $G^{-}_a$, $M^{+}_a$ and $M^{-}_a$ include affordance nodes that correspond to a specific \textit{action} and \textit{subject} instead of all affordance nodes.

\begin{figure}[hbt!]
\centerline{
	\subfigure[]{
    	\label{fig:taskaffordance_2}
    	\includegraphics[width=0.4\textwidth]{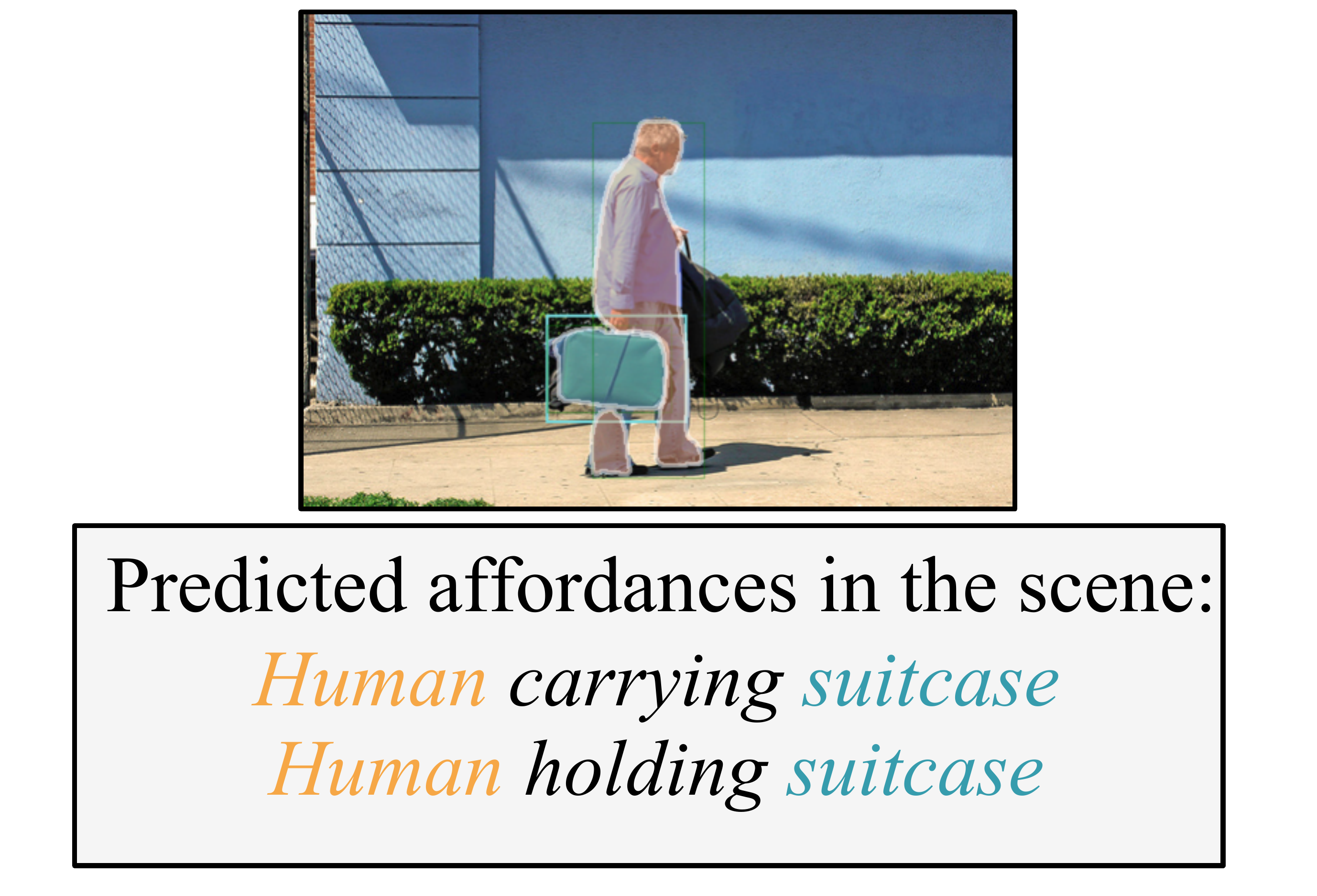}
        }
	\subfigure[]{
    	\label{fig:taskaffordance}
    	\includegraphics[width=0.4\textwidth]{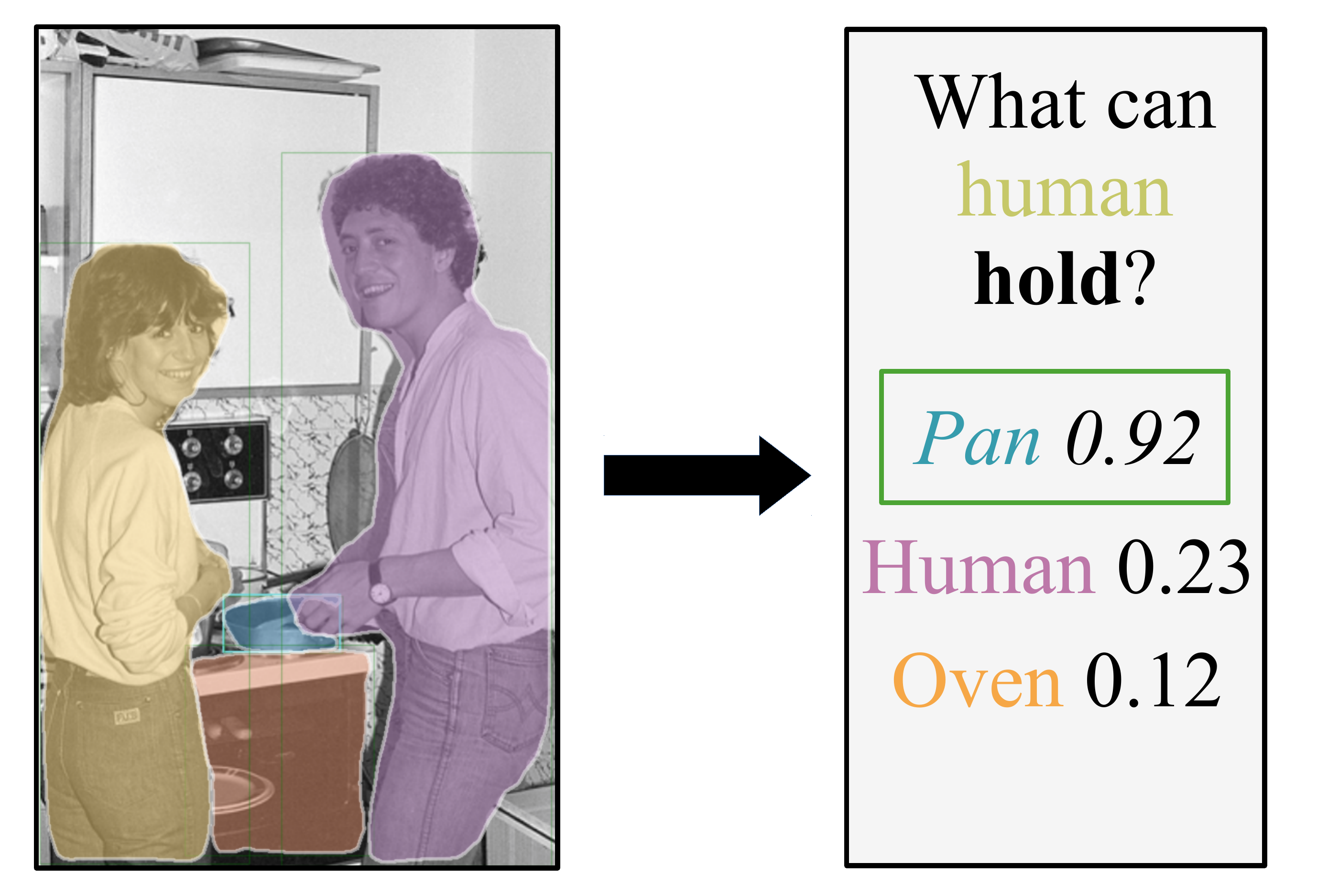}
    }
}
\caption{Some examples illustrating (a) the performance of COSMO on affordance prediction (Task 4) and (b) finding objects that affording specific action (Task 5). \label{fig:affordances}}
\end{figure}

Table \ref{tbl:affordable_object_estimation_f1_table} lists the performance of the methods for finding the objects affording a certain action. We see a significant difference between the performance of COSMO and those of GBM, RBM and RN. See also Figure \ref{fig:taskaffordance}, which shows some visual examples for predicting the object that affording specific action. \SEC{We estimated the chance levels as in Task 2.}

\begin{algorithm}
\caption{The algorithm for finding objects that afford a given action (Task 5). }\label{alg:task5}
\begin{algorithmic}[1]
\State \textbf{Input:} A scene, $\mathbf{s}$; an action, \textit{act}; the subject of action, \textit{subj}; the number of Gibbs steps, $k$.
\State \NEW{ \textbf{Output: } Index of the object of given action, ${i_o}$.}
\State $i_a$ = index of \textit{act} in affordance vocabulary.
\State $i_s$ = index of \textit{subj} in object vocabulary. 
\For{ $k$ sampling steps }
\State $\mathbf{o} \gets \mathbf{s_o}$ \Comment{Clamp objects to the visible nodes.}
\State $\mathbf{r} \gets \mathbf{s_r}$ \Comment{Clamp relations to the visible nodes.}
\State $\mathbf{a} \gets \mathbf{s_a}$ \Comment{Clamp affordances to the visible nodes.}
\State Sample hidden nodes \textbf{h}  using Eq. \ref{eqn:ph}.
\State Sample affordance node $\mathbf{a}_{{i_a}{i_s}{i_o}}$  using Eq. \ref{eqn:po}.
\EndFor
\end{algorithmic}
\end{algorithm}

\begin{table}[hbt]
\caption{Task 5 (finding objects affording a given action) performances. \label{tbl:affordable_object_estimation_f1_table}}
\centering
\footnotesize
\begin{tabular}{|l|ccc|}\hline                                                                 
&\textbf{Precision}  & \textbf{Recall}  & \textbf{F1-score}  \\ \hline \hline
COSMO & \textbf{0.3170}  & \textbf{0.4482} & \textbf{0.3714}    \\ 
GBM & 0.2537 & 0.0739 & 0.1144  \\ 
RBM & 0.0740 & 0.0869 & 0.0800   \\
RN & 0.0157 & 0.0689 & 0.0256   \\ 
\SEC{Chance level}  & \SEC{0.0056} & \SEC{0.0056}& \SEC{0.0056} \\ \hline
\end{tabular}
\end{table}

\subsection{Task 6: Who is the actor for this task?}

Robots should also be able to reason about the possible actors (subjects) of a given action or a task. Context plays a critical role here since it can modulate the candidate actors for a given action.

In this task, we evaluate the performances on finding proper subjects (actors) for a certain action with a specific object. For this end, some of the object nodes, which are the subject part of an affordance-triplet, and the affordance nodes are deactivated. Then, the model samples the hidden nodes from the partially observed scene. In the reconstruction phase, the model samples the deactivated objects and the affordance nodes that have the proper subject for the given action by using the context and the observed scene. This is formalized in Algorithm \ref{alg:task6}.

For this task, we use the same definitions of TP, FP, TN and FN in Equation \ref{eqn:task4hypothesiseqns}. However, in this task, $G^{+}_a$, $G^{-}_a$, $M^{+}_a$ and $M^{-}_a$ include affordance nodes that correspond to a specific \textit{action} and \textit{object} instead of all affordance nodes.

In Table \ref{tbl:actor_of_affordance_f1_table}, the performances of the methods are listed. We see that GBM performs better in terms of precision whereas COSMO yields a much better recall performance, leading to an overall better performance in terms of the F1-score. \SEC{We calculated chance levels with same method as in Task 2.}

\begin{algorithm}
\caption{The algorithm for finding the subject for a given action (Task 6).}\label{alg:task6}
\begin{algorithmic}[1]
\State \textbf{Input:} A scene, $\mathbf{s}$; an action, \textit{act}; the object of action, \textit{obj}; the number of Gibbs steps, $k$.
\State \NEW{ \textbf{Output: } Index of the subject of given action, ${i_s}$.}
\State $i_a$ = index of \textit{act} in affordance vocabulary.
\State $i_o$ = index of \textit{obj} in object vocabulary. 
\For{ $k$ sampling steps }
\State $\mathbf{o} \gets \mathbf{s_o}$ \Comment{Clamp objects to the visible nodes.}
\State $\mathbf{r} \gets \mathbf{s_r}$ \Comment{Clamp relations to the visible nodes.}
\State $\mathbf{a} \gets \mathbf{s_a}$ \Comment{Clamp affordances to the visible nodes.}
\State Sample hidden nodes \textbf{h}  using Eq. \ref{eqn:ph}.
\State Sample affordance node $\mathbf{a}_{{i_a}{i_s}{i_o}}$  using Eq. \ref{eqn:po}.
\EndFor
\end{algorithmic}
\end{algorithm}

\begin{table}[hbt]
\caption{Task 6 (What is the actor of the affordance?) performances. \label{tbl:actor_of_affordance_f1_table}}
\centering
\footnotesize
\begin{tabular}{|l|ccc|}\hline                                                                 
&\textbf{Precision}  & \textbf{Recall}  & \textbf{F1-score}  \\ \hline \hline
COSMO & 0.3055  & \textbf{0.4782} & \textbf{0.3728}    \\ 
GBM & \textbf{0.3333} & 0.0689 & 0.1142  \\ 
RBM & 0.0539 & 0.0586 & 0.0561   \\
RN  & 0.0312 & 0.1739 & 0.0529 \\ 
\SEC{Chance level}  & \SEC{0.0056} & \SEC{0.0056}& \SEC{0.0056} \\ \hline
\end{tabular}
\end{table}

\subsection{Task 7: Improving Object Detection}

In this task, we test whether we can use COSMO to rectify wrong detections and to find missing detections made by object detectors. For this purpose, we used three state-of-the-art object detection networks (namely, RetinaNet \cite{lin2017focal}, Faster R-CNN \cite{ren2015faster}, and Mask R-CNN \cite{he2017mask})  with the ResNet-101-FPN \cite{he2016deep} backbone model trained on the COCO dataset \cite{lin2014microsoft}. 

For this task, we first run the deep object detector on the input image. Then, we provide the detected objects to COSMO, and relax the network to see how COSMO updates the object nodes. We calculate an average precision over 100 randomly selected images and compare the performance of the deep detectors before and after applying COSMO. 

\begin{figure}[hbt!]
\centering{
    	\includegraphics[width=\textwidth]{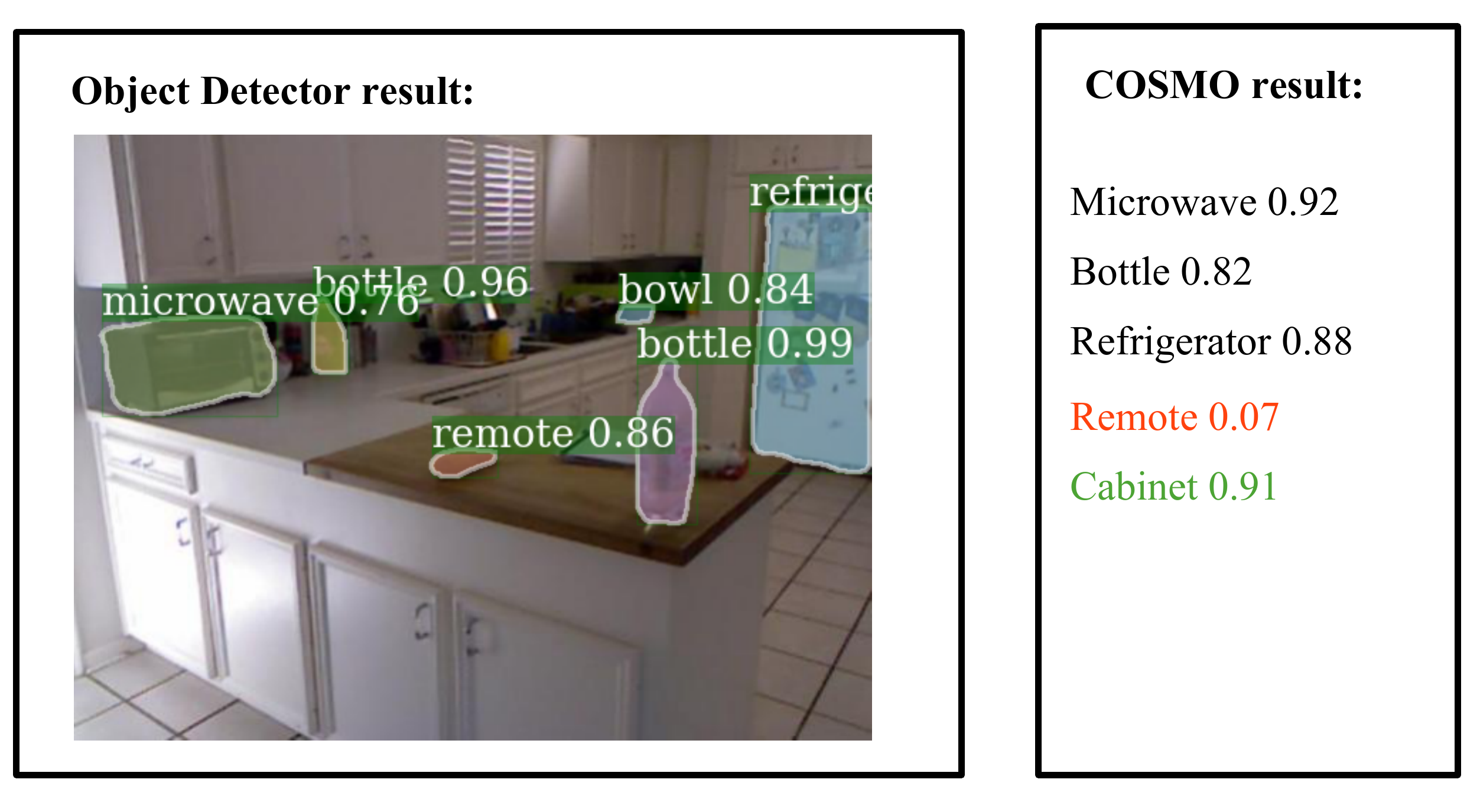}
}
\caption{An example showing that COSMO improves the output of an object detector. COSMO is provided by the objects that are labeled by the object detector and updates the object nodes. After this step, COSMO assigns low probability to ``remote'' object (i.e., it determines ``remote'' as out of context). Then, it assigns high probability to ``cabinet'' object that should exist in the scene according to the context (i.e., it determines ``cabinet'' is missing in the scene). Other objects are omitted for the sake of visibility. \label{fig:object_detector}}
\end{figure}

As shown in Table \ref{tbl:object_detectors}, COSMO significantly improves the detection performance of the deep networks. Looking at the visual example provided in Figure \ref{fig:object_detector}, we observe that COSMO can correct the mistakes made by the object detectors, and suggest objects that were missed by the detectors. \NEW{In the table, average precisions are calculated according to the definitions of TP, FP, TN and FN in Equation \ref{eqn:task2hypothesiseqns} as follows:
\begin{equation}
AP = \frac{1}{N} \sum_{i=1}^{N} \frac{TP_{s_i}}{TP_{s_i} + FP_{s_i}} ,
\end{equation}
\noindent where $N$ is the number of test scenes; $TP_{s_i}$ and $FP_{s_i}$ are the number of true positives and false positives in scene $s_i$ respectively.
}

\begin{table}[hbt]
\caption{Task 7: Improving object detections with COSMO. The average precision (AP) for different object detectors with and without COSMO are listed. \textbf{Detector} refers to the performance of the plain object detectors. \textbf{w COSMO}, \textbf{w GBM} and \textbf{w RBM} respectively indicate the results of COSMO, GBM and RBM applied on the outputs of the detectors. \textbf{w COSMO w/o aff.} shows the result of COSMO trained without affordance nodes. \label{tbl:object_detectors}}
\centering
\scriptsize
\begin{tabular}{|l|ccccc|}\hline  
&\textbf{Detector}  & \textbf{w COSMO} & \NEW{\textbf{w COSMO}} & \NEW{\textbf{w GBM}} & \NEW{\textbf{w RBM}}  \\ 
&  &  & \NEW{\textbf{w/o aff.}}  &  &  \\ \hline \hline
RetinaNet \cite{lin2017focal} & $0.4964$ & $0.6966$ & $\mathbf{0.6994}$ & {$0.6761$} & $0.2603$  \\ 
Faster R-CNN \cite{ren2015faster} & $0.4388$ & $0.6752$   & $\mathbf{0.6813}$ & $0.6294$ & $0.2258$ \\ 
Mask R-CNN \cite{he2017mask} & $0.4273$ & $0.6648$   & $\mathbf{0.6677}$ & $0.6114$ & $0.2227$ \\ \hline
\end{tabular}
\end{table}

\SEC{As shown in Table \ref{tbl:object_detectors}, COSMO that is trained without affordance nodes is slightly better than the regular COSMO since the total number of affordance nodes is quite few compared to the numbers of object and relation nodes. Therefore, the effect of object and relation nodes to hidden activations can be diminished by the affordance nodes.}


\subsection{Task 8: Random scene generation}

\begin{figure}
\centerline{
	\subfigure[]{
        	\includegraphics[width=0.6\textwidth]{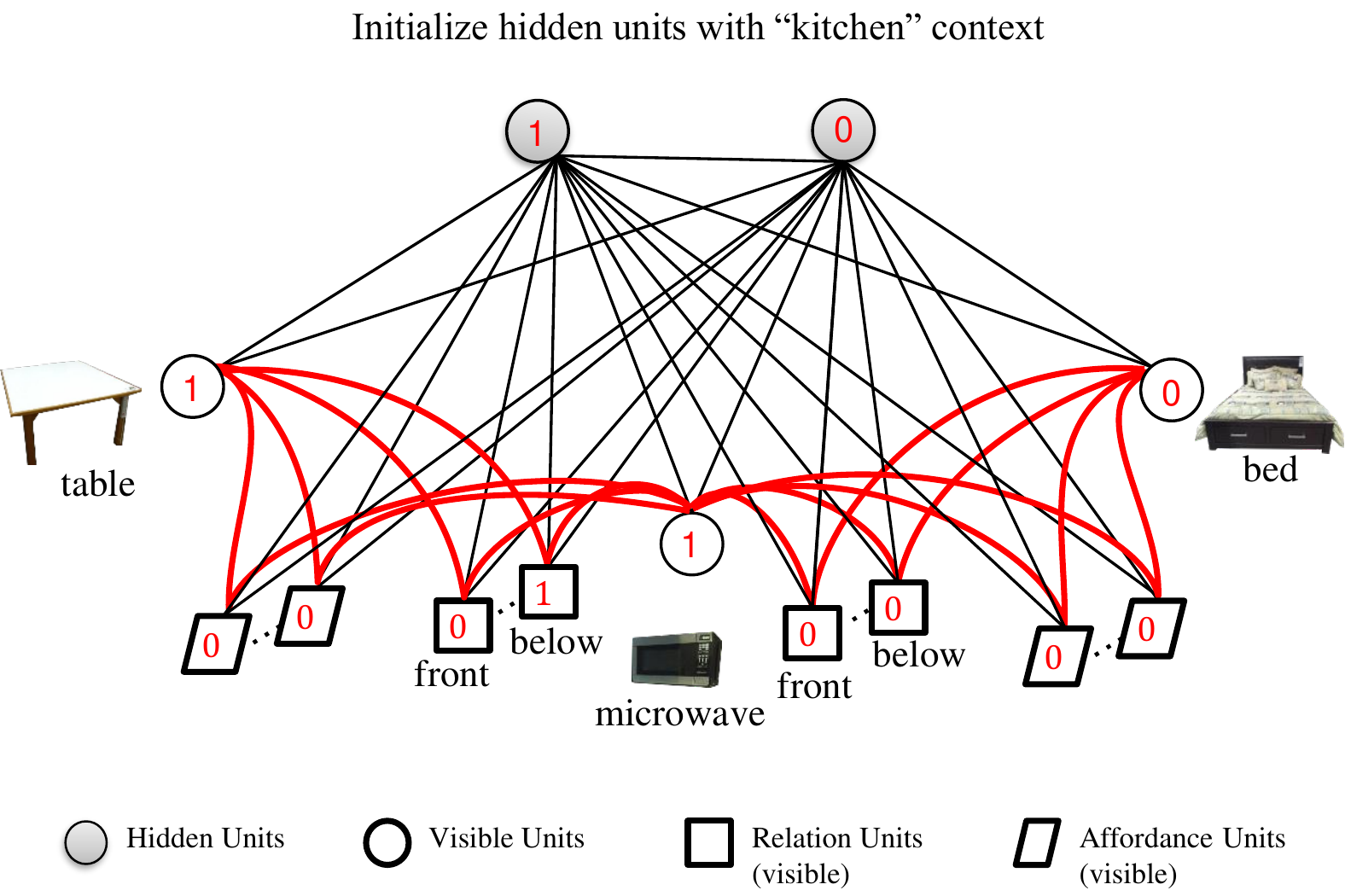}
    }
  	\subfigure[]{
    	\includegraphics[width=0.3\textwidth] {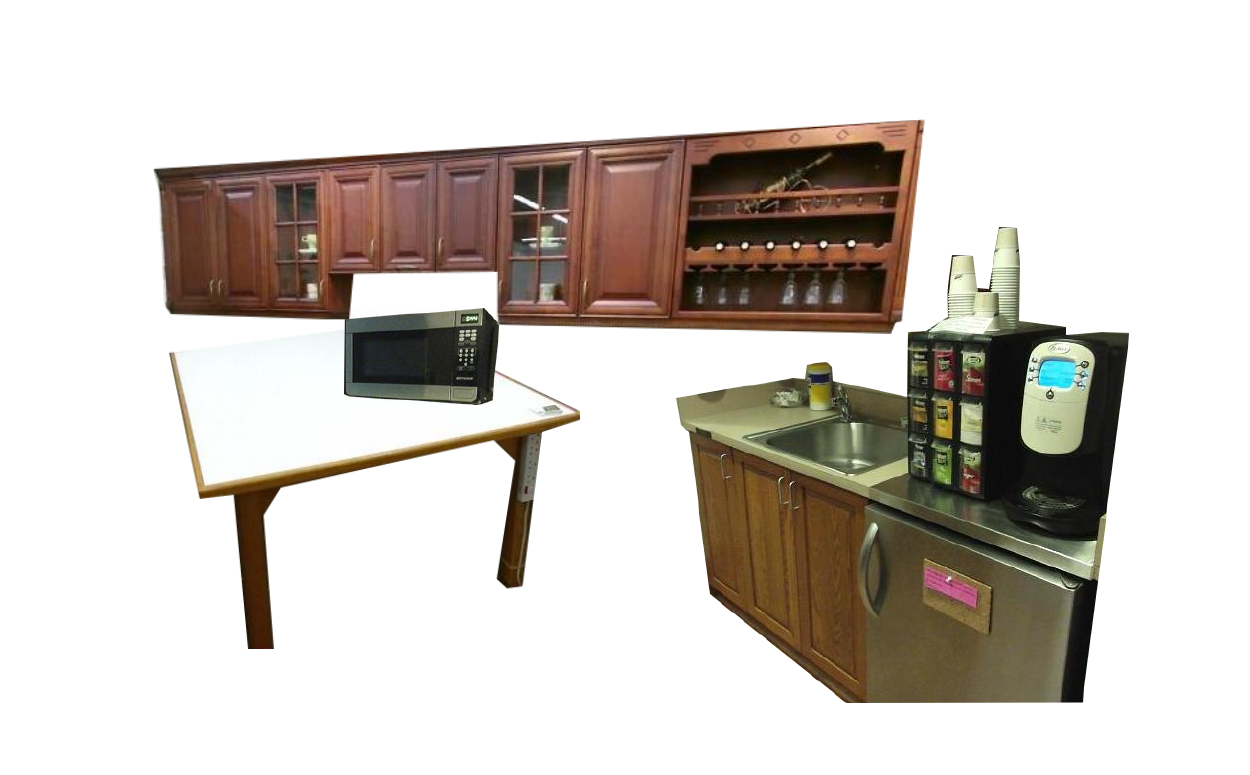}
    }
}
\caption{An example illustrating scene generation capability of COSMO (Task 8). (a) When a context (hidden node) is activated, (b) active nodes in the sampled visible nodes define a scene for the context. In (b), the ``selected' objects are placed in the scene based on the predicted spatial relations. \label{fig:randomscenegeneration}}
\end{figure}

In this task, we demonstrate how we can use another generative ability of COSMO: we can select a hidden node (or more of them, leaving the other hidden neurons randomly initialized or set to zero), and sample visible nodes (including relations and affordances) that describe a scene. Figure \ref{fig:randomscenegeneration} shows a visual example.  

\subsection{Experiments on a Real Robot}

\begin{figure}
\centerline{
	\includegraphics[width=0.7\textwidth]{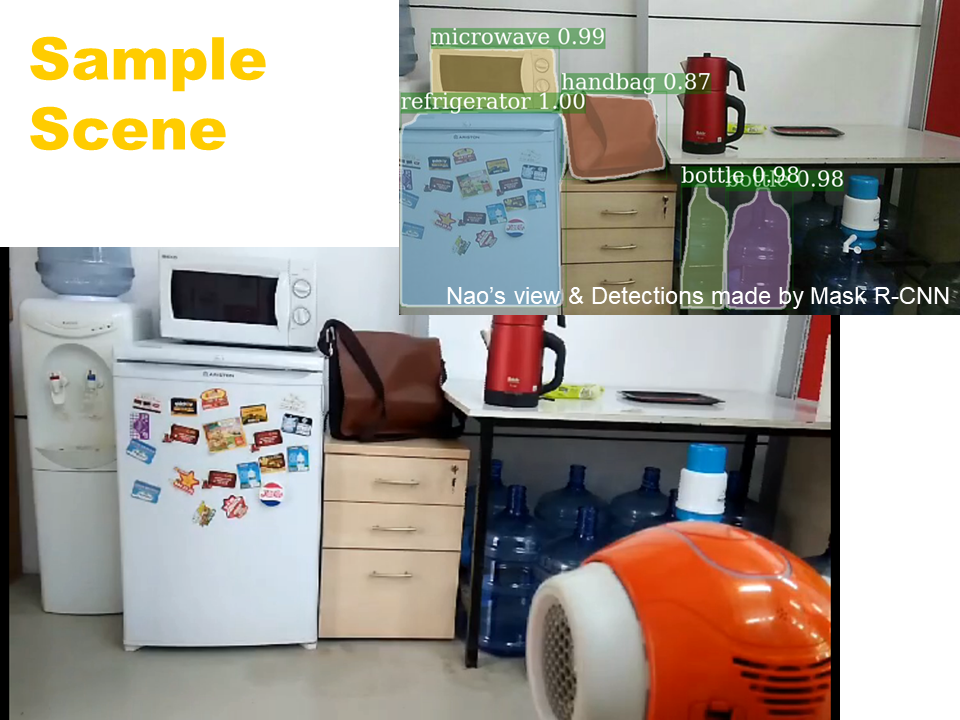}
}
\caption{A snapshot of an experiment performed with Nao. Nao uses Mask R-CNN to detect objects in the scene, and COSMO is initialized with these detections. Once this has been performed, Nao can reason about relations, affordances, missing objects or out-of-context objects in the scene. See the accompanying video (also provided at \url{https://bozcani.github.io/COSMO}) for the experiments. \label{fig:task_real_robot}}
\end{figure}

In this experiment, we evaluate COSMO on Nao and illustrate how Tasks 1-7 conducted in this section can be useful for a robot. For this purpose, Nao uses Mask R-CNN to detect objects in the scene, and COSMO is initialized with these detections (only object nodes are clamped with the detected objects, the other visible nodes (relations and affordances) are estimated after sampling the hidden nodes). See Figure \ref{fig:task_real_robot} for a snapshot.

Once COSMO is relaxed, Nao can reason about objects, relations, affordances, missing objects or out-of-context objects in the scene. An interactive experiment has been conducted with Nao where Nao answers questions about the scene using the active nodes in COSMO. See the accompanying video (provided at \url{https://bozcani.github.io/COSMO}) for the experiments. 

\subsection{Discussion of results}
\SEC{\textbf{Why is recall rather low in Task 2?}}

\SEC{In Task 2 (What is missing in the scene?), the recall rate is rather low due to a phenomenon that we name contextual bias. Contextual bias occurs when an object is removed from the scene, and our contextual model brings back an object that is tied with the context the most. In most cases, the recalled object is different from the removed object and therefore, this counts towards the recall rate. This suggests that, for this task, the recall (FP) definition may not be suitable -- a definition that counts recalled objects that are compatible with the context may have been better. However, we do not have access to an explicit object-context relationship information, which makes an alternative definition difficult.} 


\SEC{\textbf{Why is recall rather low in Task 3?}}

\SEC{In Task 3 (What is extra in the scene?), the extra object that should be removed from the model may have more contextual importance than the objects in the ground-truth test sample. In this case, the added object can dominate the scene context and the COSMO removes an object that belongs to the ground-truth sample, and it corrupts original input data. The same problem has been observed in our previous work \cite{bozcan2018missing}.}

\SEC{\textbf{Do these low values mean that the model is useless?}}

\SEC{No, it just suggests that the contextual models can predict objects in a scene according to the context. If the context clearly suggests an object, then the models are able to predict them; otherwise, the models cannot recall the removed objects that are not-strongly-contextualized or compatible with the current context. This does not mean that these models are useless, it just suggests that context can predict objects only suitable for the context, which may or may not be what one expects.}

\SEC{\textbf{Results of similar studies in the literature:}}

\SEC{Compared to the similar studies in the literature (see below for a small summary), the reported performances are reasonable since the problem of estimating an entity based on the context (without looking at the representation/features of the entity) is very challenging.}

\SEC{In the literature, results of finding missing objects by using only context models are rather low compared to the baselines. For example, Torralba and his colleagues \cite{choi2012tree} proposed a tree-based context model (TreeContext) and conducted several experiments including object presence prediction and finding out-of-context objects. For object presence prediction, TreeContext could not achieve significant performances compared to the baseline models.}

\SEC{Mottaghi and his colleagues \cite{mottaghi2014role}, proposed a contextual scene model based on Markov Random Fields. They encountered the recall issue for a variant of the task of finding a missing object (Tables 1 and 2 in \cite{mottaghi2014role}). They stated that the main reason for the low recall performance is the high variability of scenes in the dataset. This is an issue for us the current work as well since we merged two datasets including different types of scenes: SUNRGBD includes indoor scenes and Visual Genome includes mostly outdoor scenes.}

\SEC{\textbf{Quantitative comparison with  our previous work}}

\SEC{Numerical comparison of results with our previous work \cite{bozcan2018missing} is not directly feasible since (i) different measures are used in Task 1 and Task 2, (ii) the current dataset is more difficult and (iii) the affordance representation is added to the model. In this work, the training set is richer in scenes with more objects and relations compared to the dataset in our previous work.}

\SEC{\textbf{Combinatorial Explosion of Connections Problem}}

\SEC{To overcome combinatorial explosion owing to the connections, the raw input scene including objects, relations and affordances can be encoded by autoencoders in order to reduce the dimensionality and embed the inputs in an embedding space. The embedding space can be learned using autoencoders (e.g., using an RBM). Then, this embedding space can be used as the input of a Boltzmann Machine with real-valued visible nodes.}

\section{Conclusion}

In this paper, we proposed a novel method (COSMO) for contextualized scene modeling. For this purpose, we extended Boltzmann Machines (BMs) to include spatial relations and affordances via tri-way edges in the model. For integrating spatial relations and affordances into the model, we introduced shared nodes into BMs, allowing the concept of relations and affordances to be shared among different objects pairs. We evaluated and compared our model on several tasks on a real dataset and a real robot platform. 


On several challenging tasks, we demonstrated that our model is very suitable for scene modeling purposes with its generative and explicit nature. Being generative, we showed that a single COSMO model allows reasoning about many aspects of the scene given any partial information. On these tasks, COSMO performed consistently better in comparison to the baseline methods (general BMs and restricted BMs) and relational networks \cite{santoro2017simple}.

\subsection{Limitations and Future Work}

\NEW{COSMO has a limitation of assuming a fixed-length object vocabulary. Therefore, novel objects that have not  been encountered yet cannot be adapted to the model. However, in real settings, the set of objects can grow in time. To overcome this problem, the input layer should be designed in an incremental manner, as suggested in \cite{CelikkanatContext2014, dougan2017deep, dougan2017learning}.}

\NEW{The second limitation is related to scalability. The model considers all possible object pairs for given relations or affordances. Therefore, the number of relations and affordances nodes increases exponentially with an increase in the number of objects.} \SEC{In our work, we used COSMO for medium-scale scene modeling tasks to reduce to the effect of the scalability problem and to train the networks in reasonable durations.} \NEW{As an alternative, an embedding of relations and affordances can be obtained and integrated to COSMO in order to reduce the dimensionality and make COSMO more scalable.}

they try to address these issues by formulating and evaluating a novel BM model for a medium-scale scene modeling task.

\NEW{COSMO represents the environment in terms of the presence of objects. It does not handle multiple instances of an object in a scene (if there are). Even if we do not handle multiple objects explicitly, information about them is partially included into the system by the activations of relations. For example, if the model find relations of ``the lamp is left of the bed'' and ``the lamp is right of the bed'', it means that there are at least two lamps in the scene.}

\NEW{In our work, spatial relations are represented as qualitative abstractions (left, right, behind etc.) from metric data. This might be inadequate for tasks requiring reasoning about precise locations of objects.}

\NEW{Lastly, the number of object nodes is rather small compared to the number of affordance and relation nodes. Therefore, the effect of object nodes on the hidden activations can be dominated by the relation and affordance nodes. To overcome this problem, additional weights can be added to edges between object and hidden nodes in order to balance the contributions of the relations, affordances and objects to hidden activations.
}

\section*{Acknowledgments}

This work was supported by the Scientific and Technological Research Council of Turkey (T\"UB\.{I}TAK) through project called ``Context in Robots'' (project no 215E133). We gratefully acknowledge the support of NVIDIA Corporation with the donation of the Tesla K40 GPU used for this research. \NEW{We thank Yagmur Oymak and Idil Zeynep Alemdar for their contributions on preparing an  earlier version of the dataset.}

\section*{References}

\bibliographystyle{elsarticle-num}
\bibliography{references}
\appendix 
\section{\SEC{Analyzing the Hyper-parameters of GBM, RBM and RN}}

\SEC{In this section, we analyze the hyper-parameters of our baseline models (RBM, GBM and RN).}

\begin{figure}[hbt!]
\centerline{
	\subfigure{
    	\includegraphics[width=0.5\textwidth]{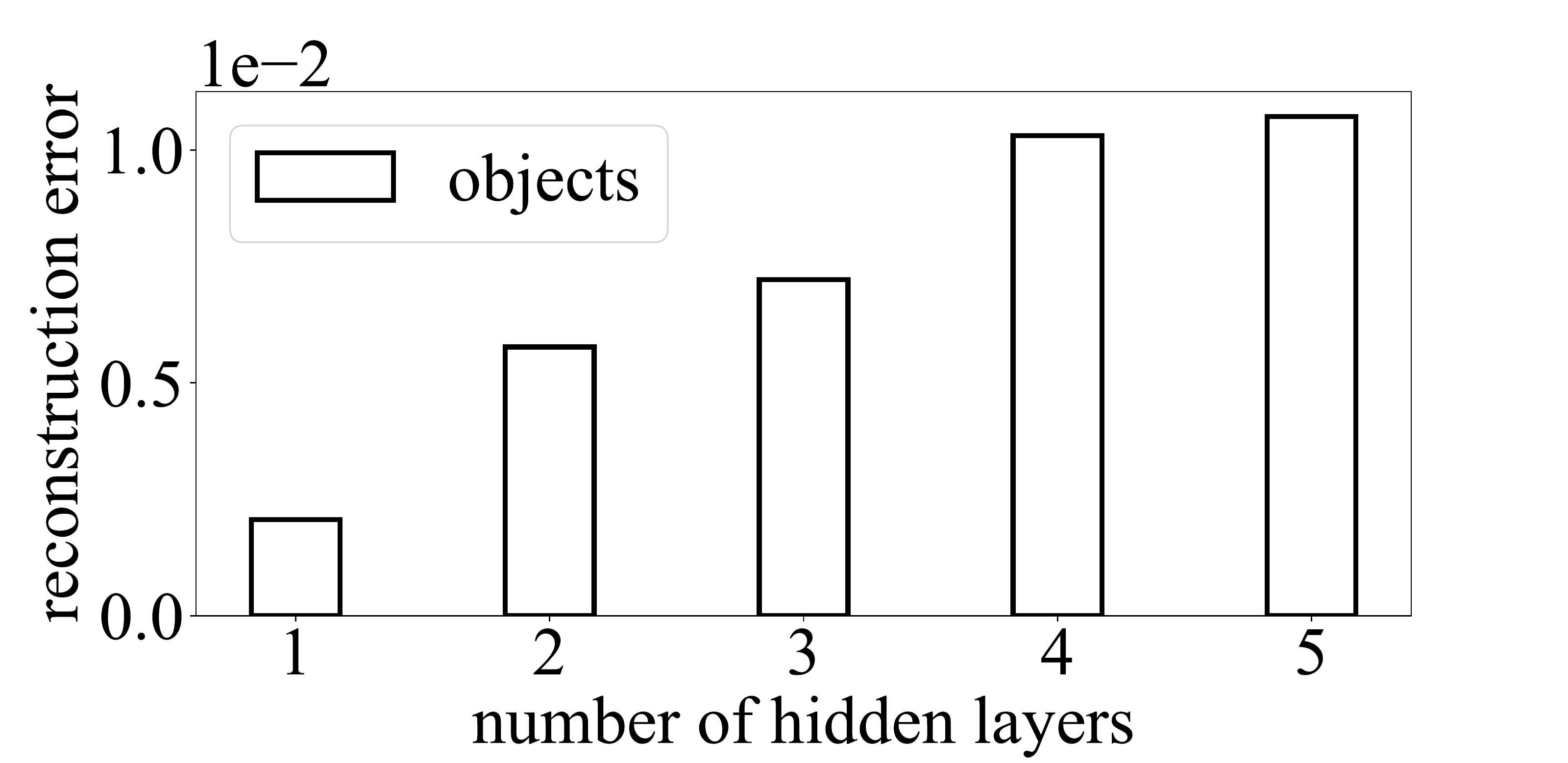}
    }
}
\centerline{
	\subfigure{
    	\includegraphics[width=0.5\textwidth]{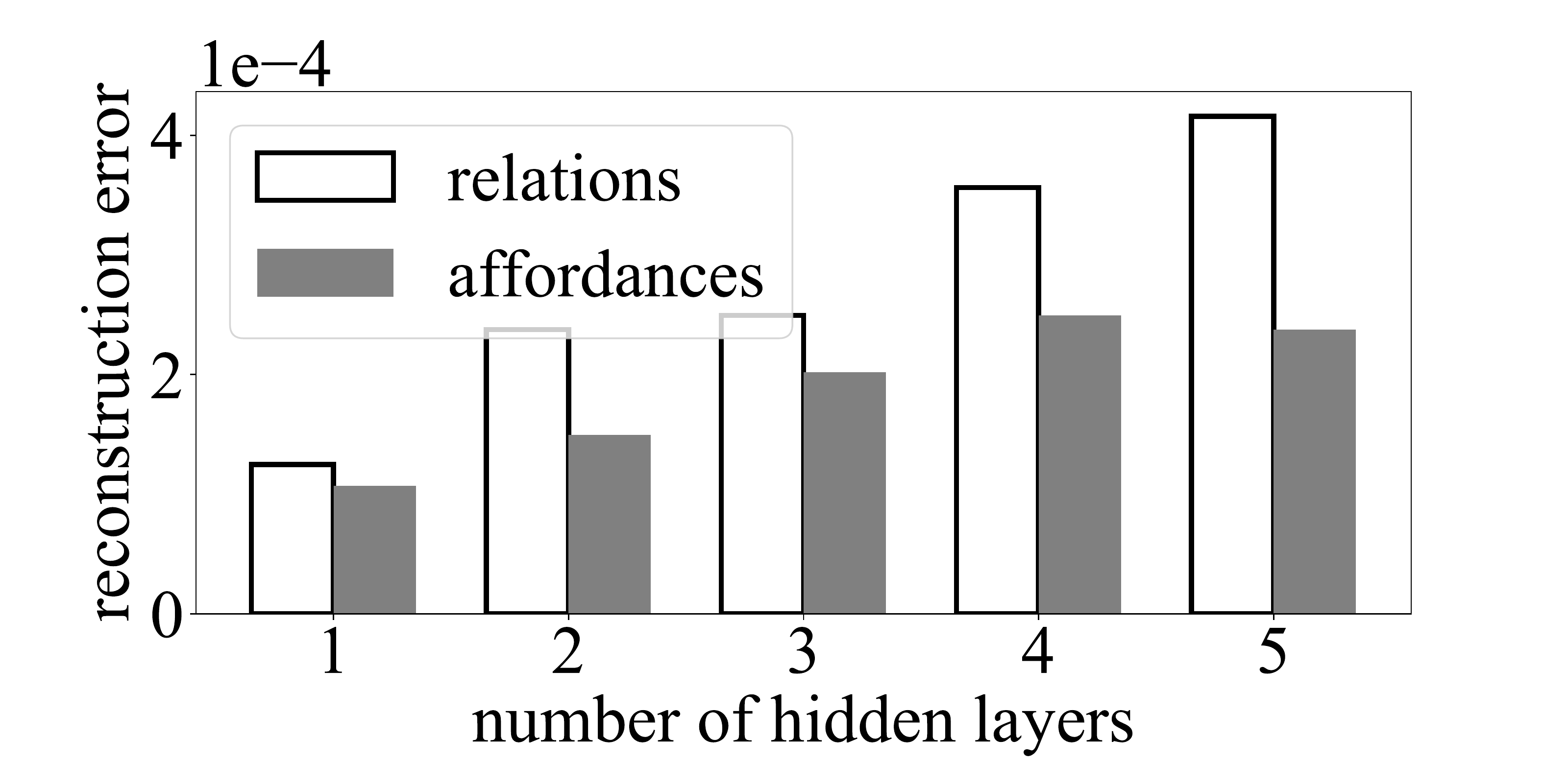}
    }
}
\caption{\SEC{Reconstruction errors of GBM for different number of hidden layers. (a) Reconstruction error for object nodes. (b) Reconstruction error for relation nodes and affordance nodes.} \label{fig:gbm_error_vs_layers}}
\end{figure}

\begin{figure}[hbt!]
\centerline{
	\subfigure{
    	\includegraphics[width=0.5\textwidth]{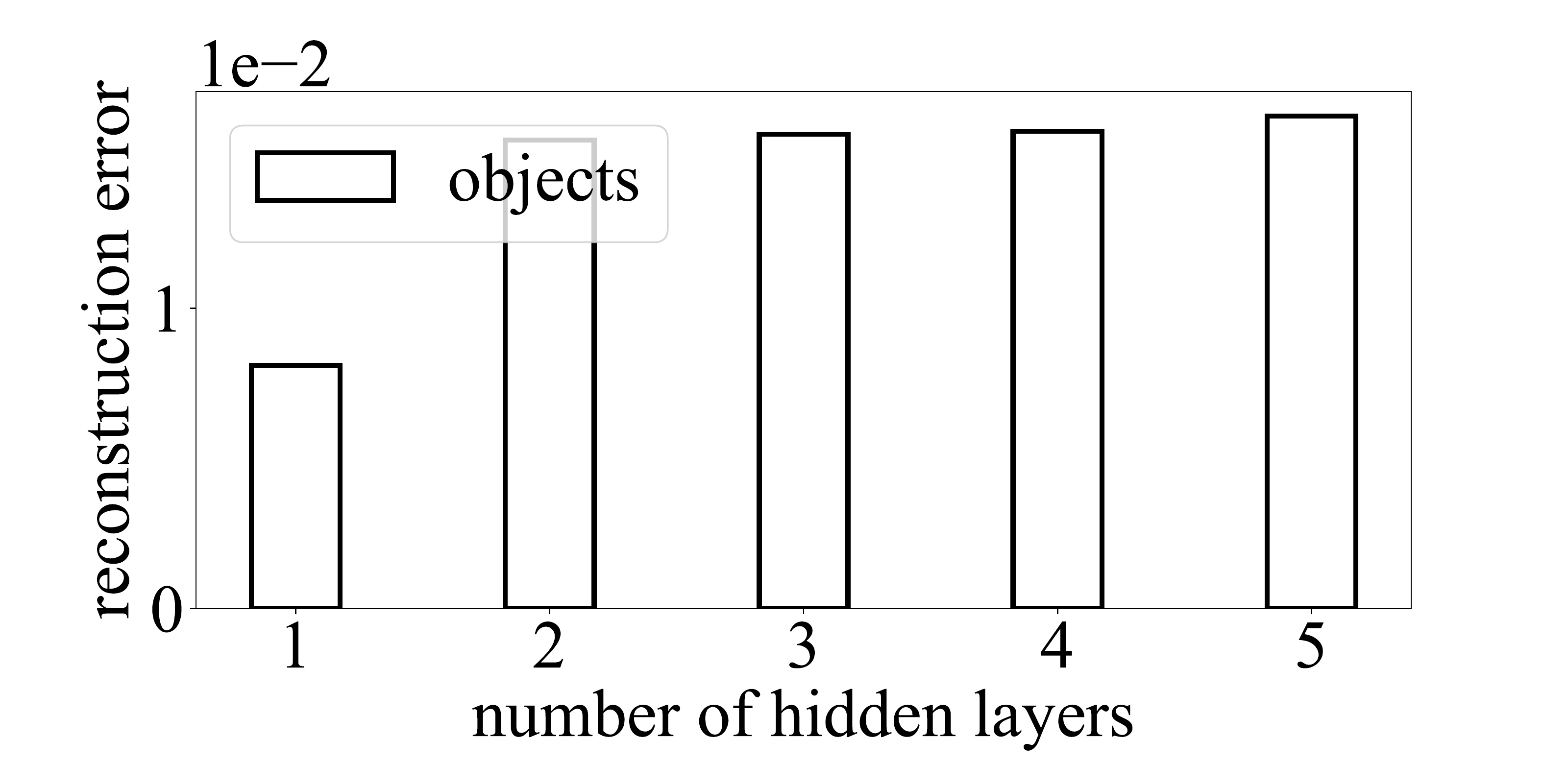}
    }
}
\centerline{
	\subfigure{
    	\includegraphics[width=0.5\textwidth]{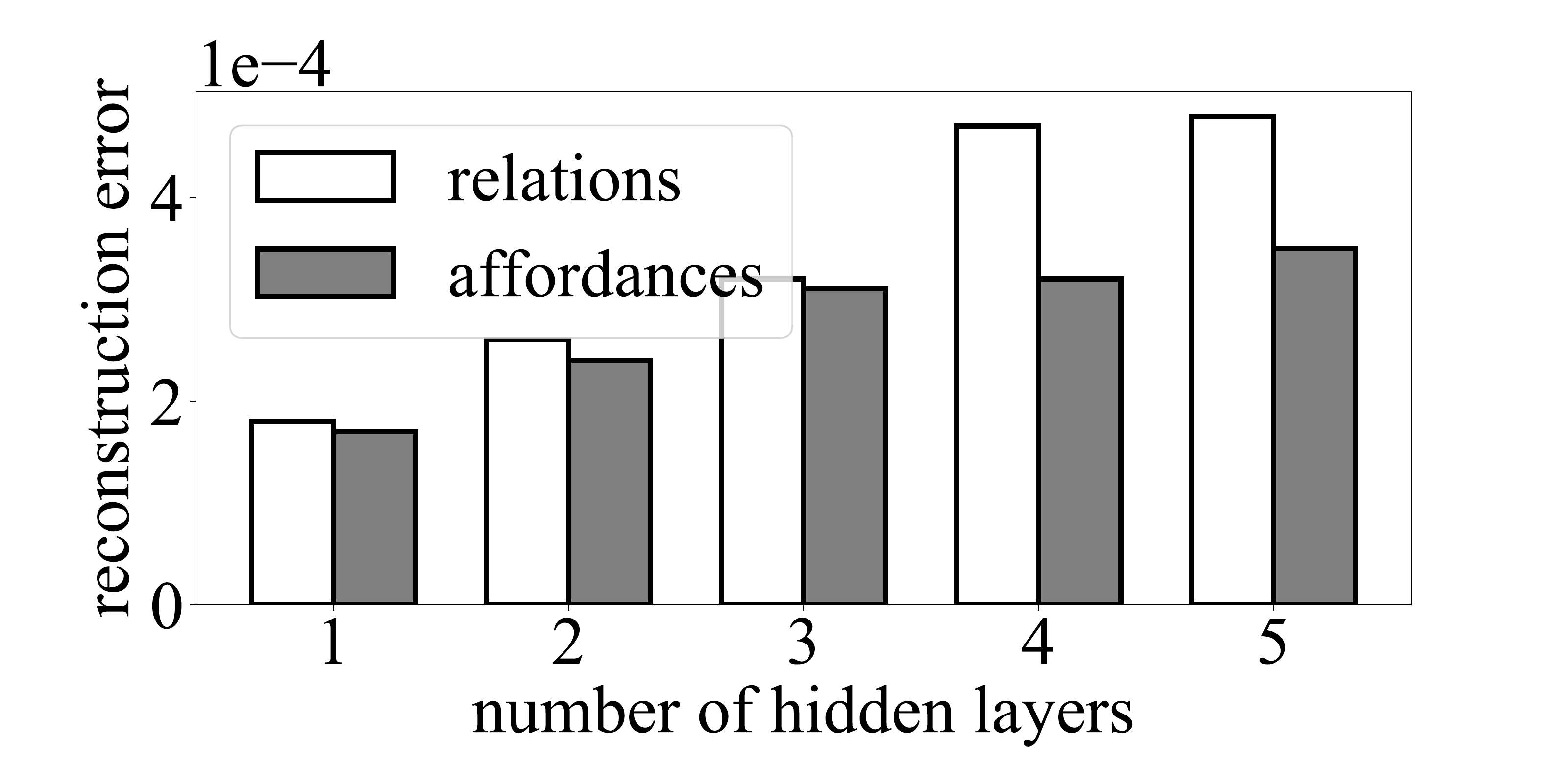}
    }
}
\caption{\SEC{Reconstruction errors of RBM for different number of hidden layers. (a) Reconstruction error for object nodes. (b) Reconstruction error for relation nodes and affordance nodes.} \label{fig:rbm_error_vs_layers}}
\end{figure}

\begin{figure}[hbt!]
\centerline{
	\subfigure{
    	\includegraphics[width=0.5\textwidth]{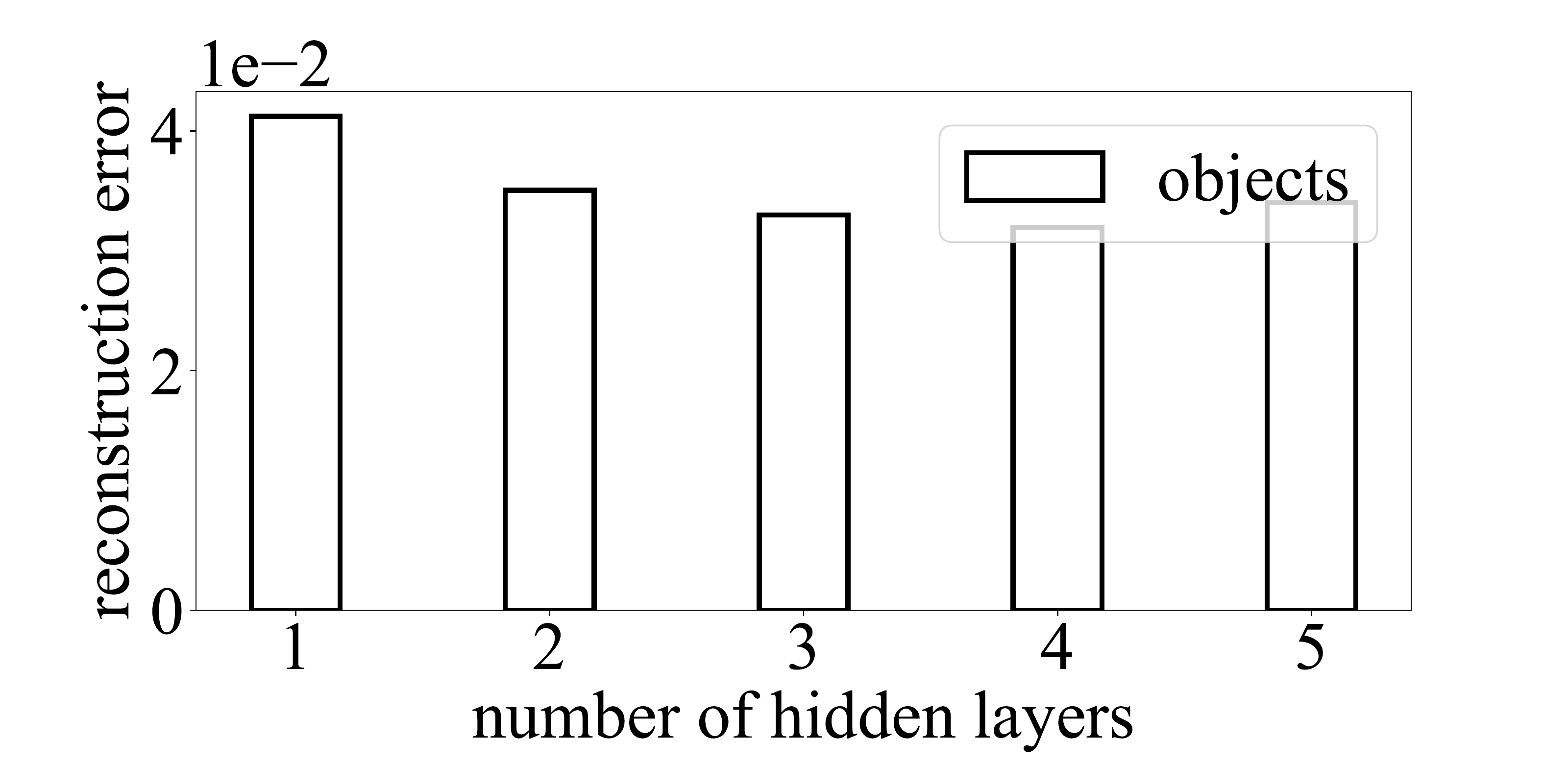}
    }
}
\centerline{
	\subfigure{
    	\includegraphics[width=0.5\textwidth]{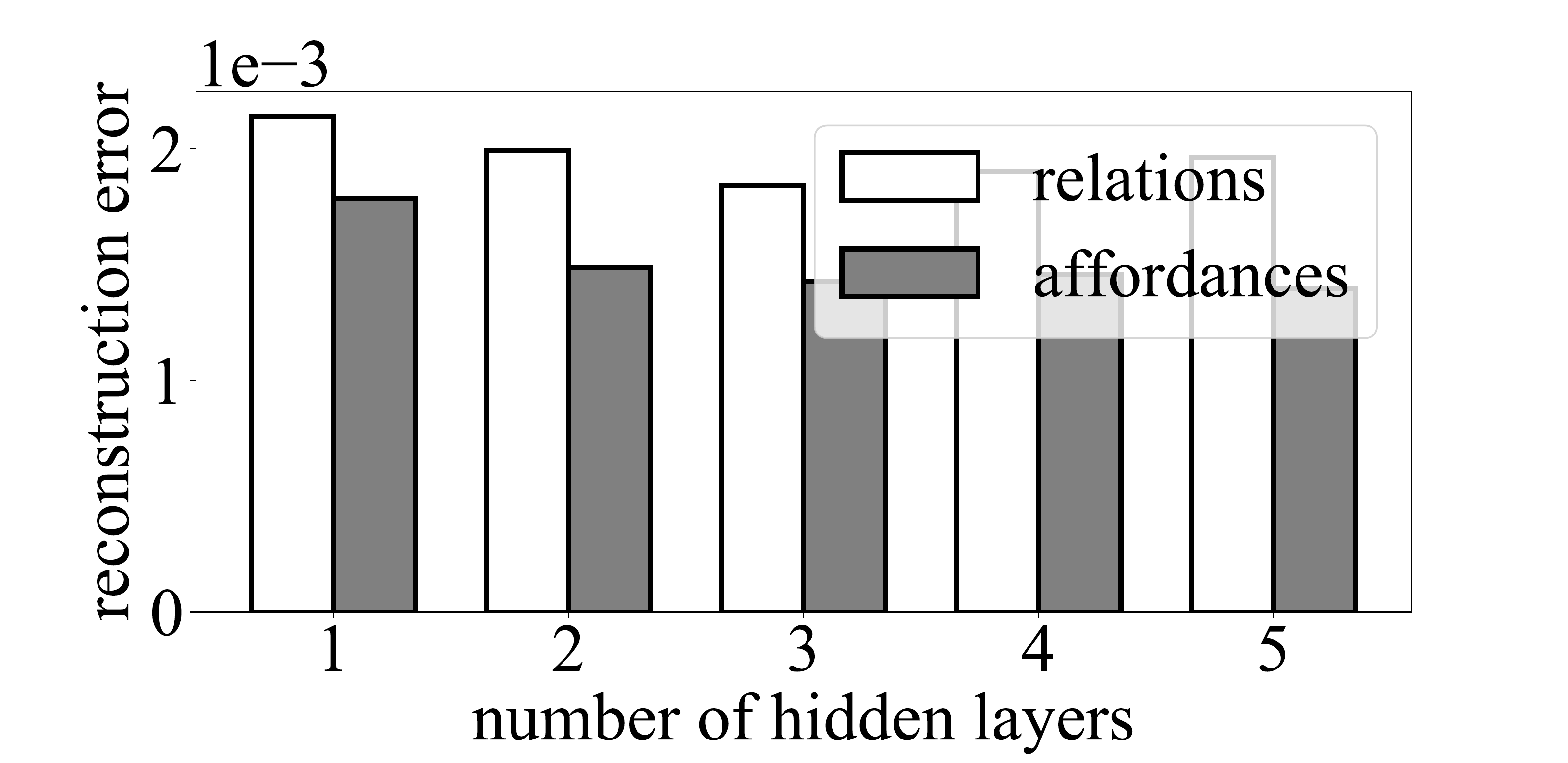}
    }
}
\caption{\SEC{Reconstruction errors of RN for different number of hidden layers. (a) Reconstruction error for object nodes. (b) Reconstruction error for relation nodes and affordance nodes.} \label{fig:rn_error_vs_layers}}
\end{figure}

\SEC{First, we investigate the effect of the number of hidden layers for GBM, RBM and RN. The results are shown in Figure \ref{fig:gbm_error_vs_layers}, \ref{fig:rbm_error_vs_layers}, \ref{fig:rn_error_vs_layers} respectively. For GBM (Fig. \ref{fig:gbm_error_vs_layers}) and RBM (Fig. \ref{fig:rbm_error_vs_layers}), the reconstruction error increases when the number of hidden layers increases as in COSMO (Fig. \ref{fig:error_vs_layers}). This issue occurs since variants of Boltzmann Machines requires more sampling steps of hidden layers as COSMO. This is not case for RN (Fig. \ref{fig:rn_error_vs_layers}) since, in artificial neural networks, hidden nodes are calculated by values of previous hidden nodes that have been already determined. In RN, although there is slight decrease in error with increasing number of hidden layers, it is not significant. That's why we choose the number of hidden layers as 1 for GBM and RBM and 2 for RN.}

\begin{figure}[hbt!]
\centerline{
	\subfigure{
    	\includegraphics[width=0.5\textwidth]{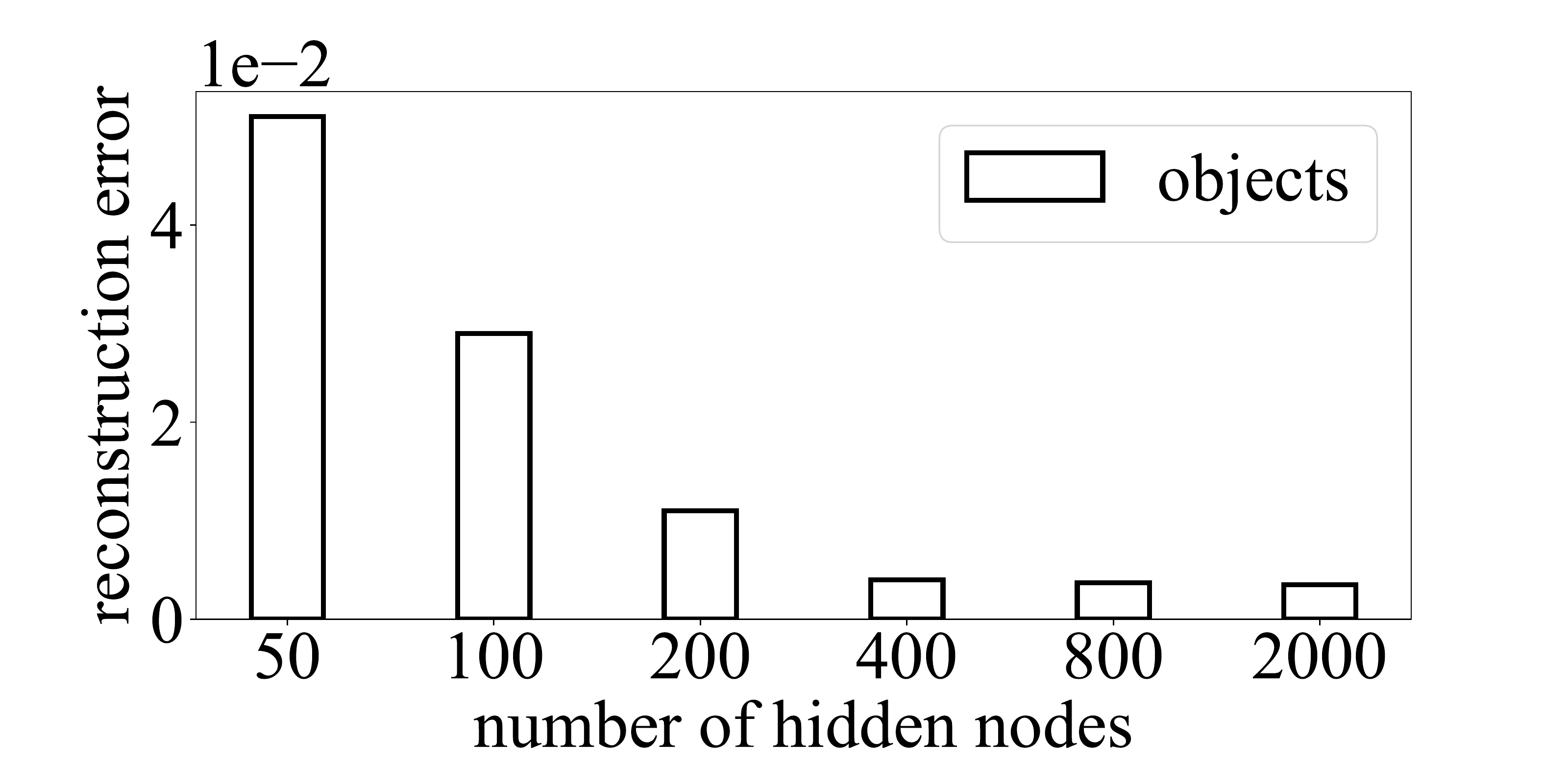}
    }
}
\centerline{
	\subfigure{
    	\includegraphics[width=0.5\textwidth]{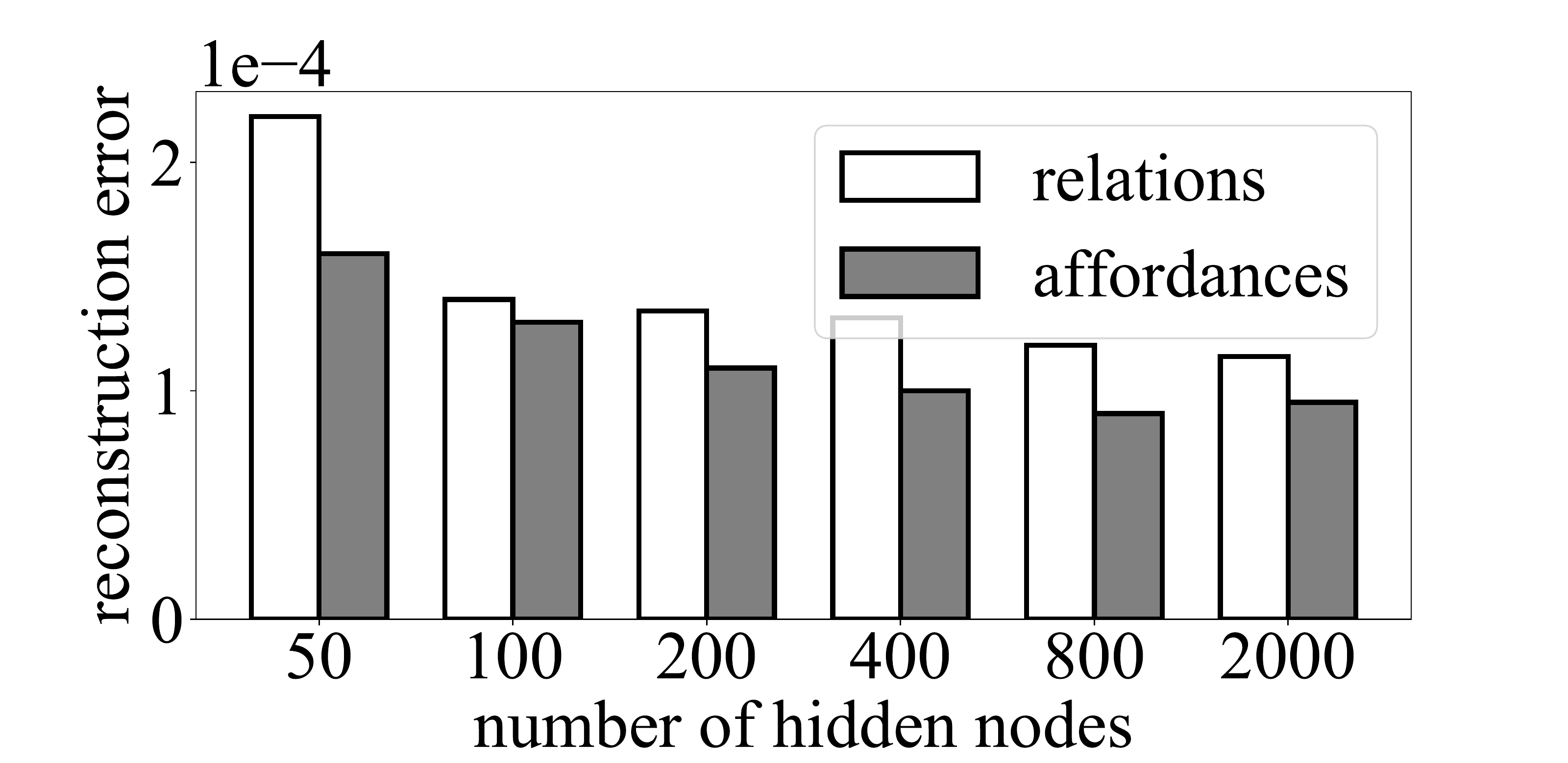}
    }
}
\caption{\SEC{Reconstruction errors of GBM for different number of hidden nodes. (a) Reconstruction error for object nodes. (b) Reconstruction error for relation nodes and affordance nodes.} \label{fig:gbm_error_vs_hiddens}}
\end{figure}

\begin{figure}[hbt!]
\centerline{
	\subfigure{
    	\includegraphics[width=0.5\textwidth]{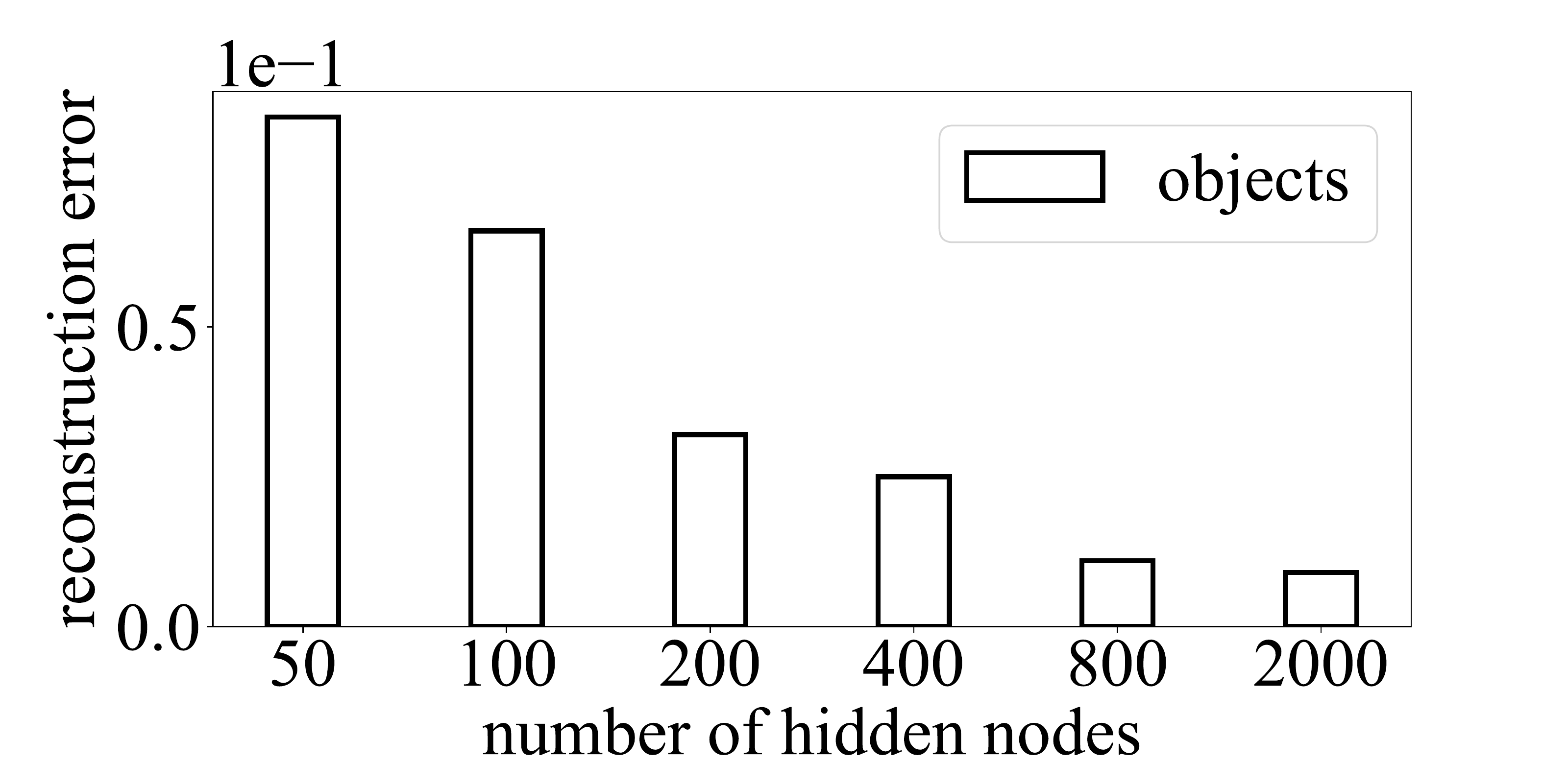}
    }
}
\centerline{
	\subfigure{
    	\includegraphics[width=0.5\textwidth]{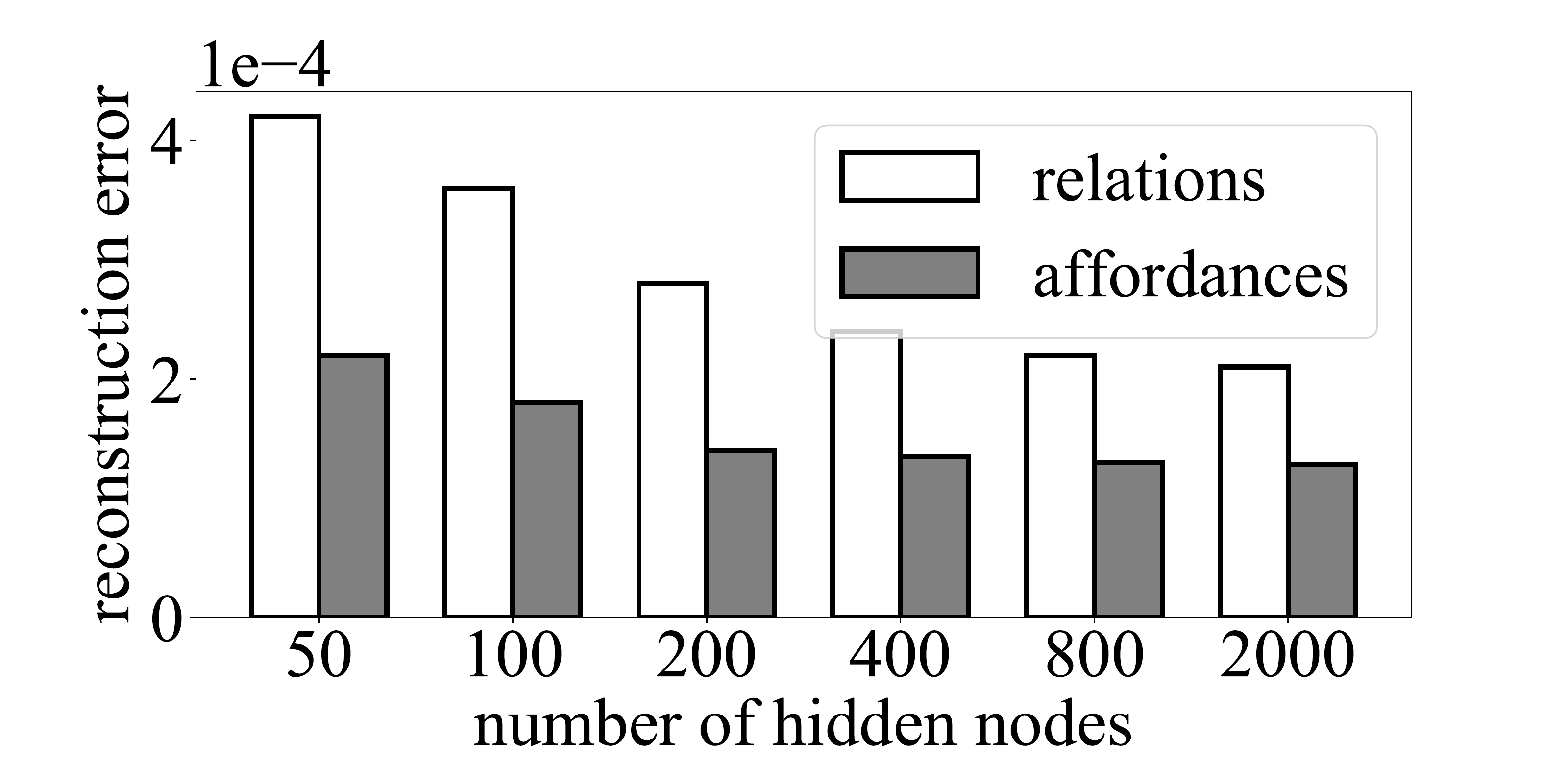}
    }
}
\caption{\SEC{Reconstruction errors of RBM for different number of hidden nodes. (a) Reconstruction error for object nodes. (b) Reconstruction error for relation nodes and affordance nodes.} \label{fig:rbm_error_vs_hiddens}}
\end{figure}

\begin{figure}[hbt!]
\centerline{
	\subfigure{
    	\includegraphics[width=0.5\textwidth]{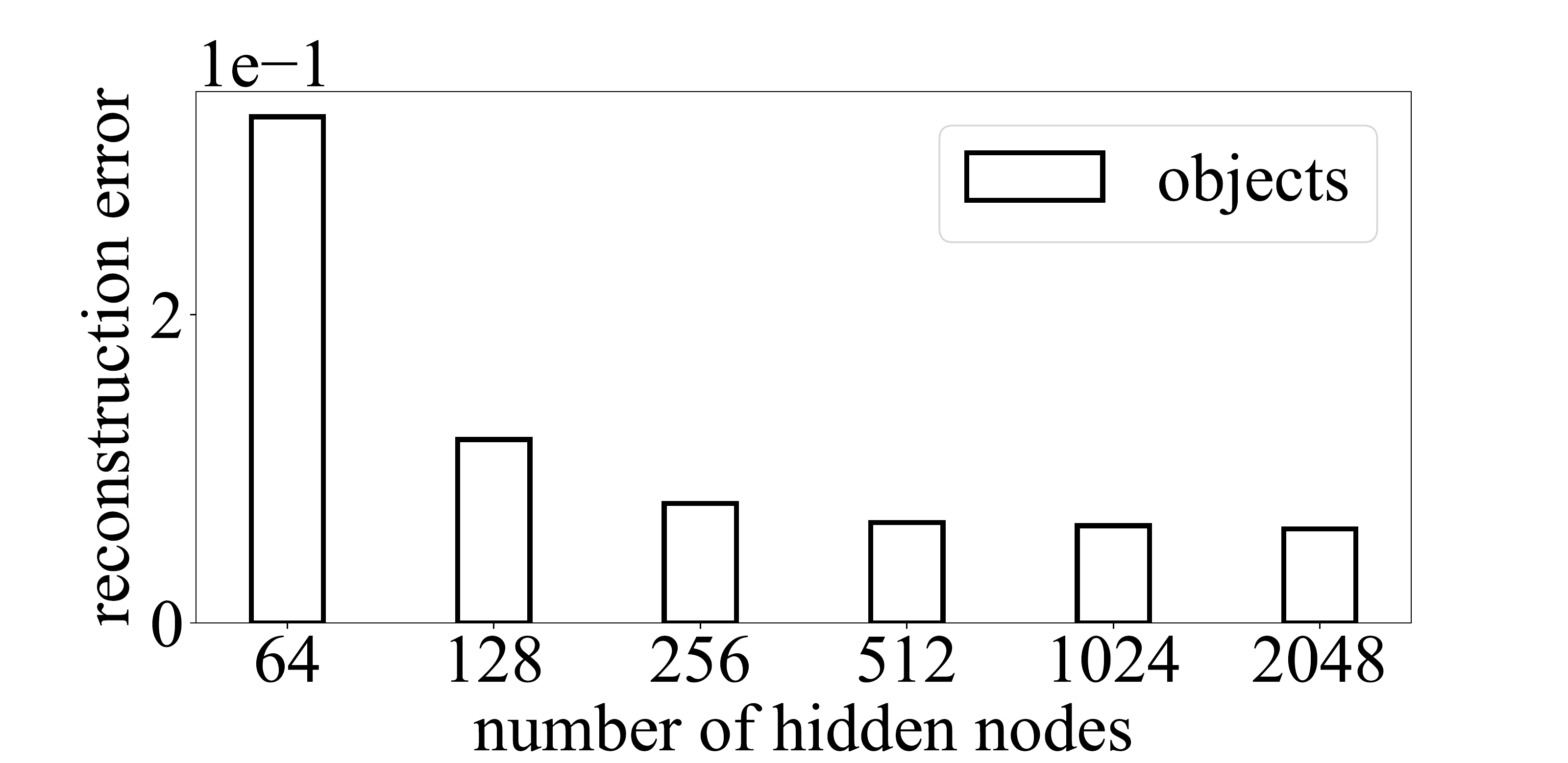}
    }
}
\centerline{
	\subfigure{
    	\includegraphics[width=0.5\textwidth]{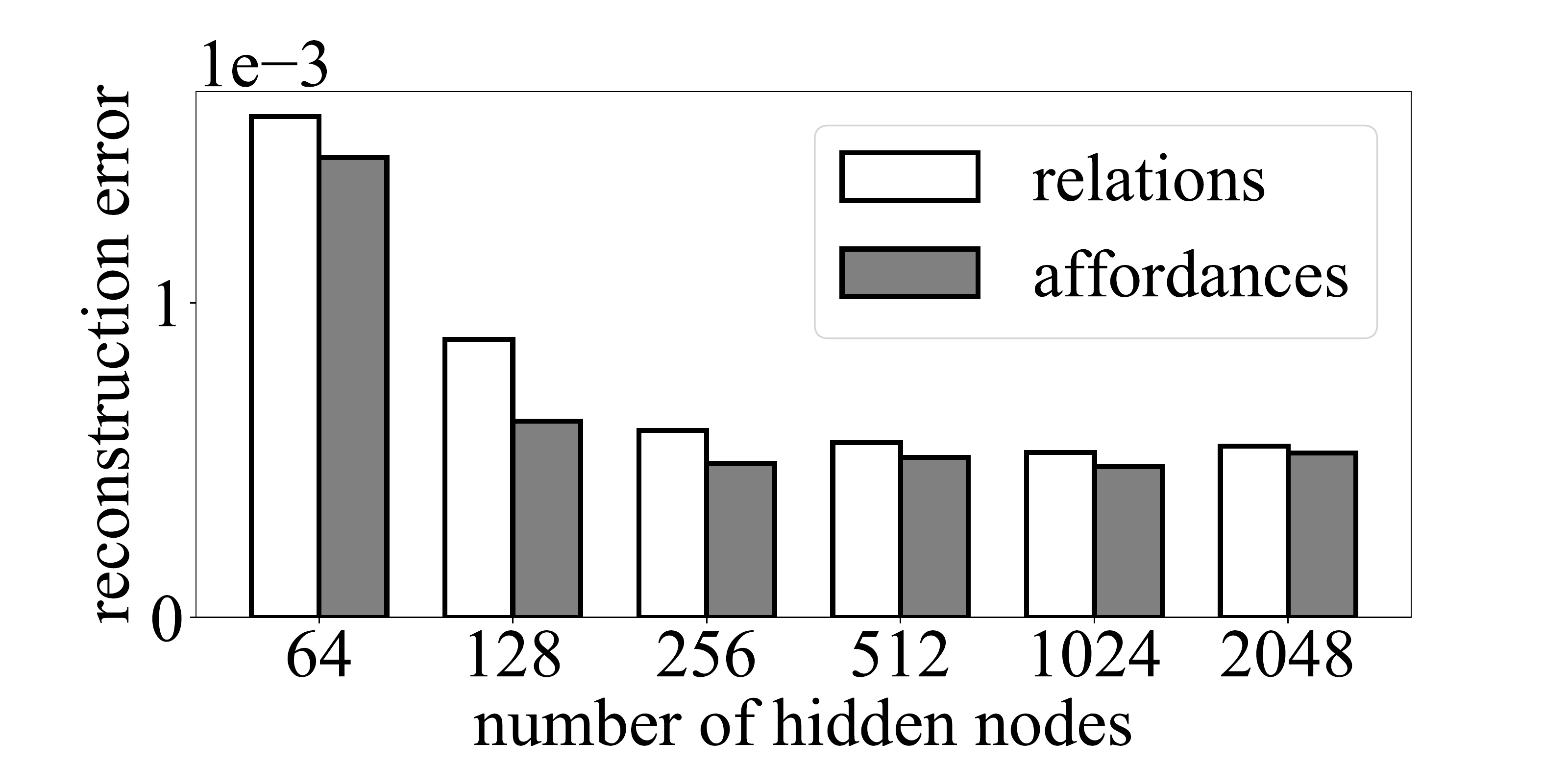}
    }
}
\caption{\SEC{Reconstruction errors of RN for different number of hidden nodes. (a) Reconstruction error for object nodes. (b) Reconstruction error for relation nodes and affordance nodes.} \label{fig:rn_error_vs_hiddens}}
\end{figure}

\SEC{Secondly, we analyzed effect of the number of hidden neurons in a hidden layer for GBM, RBM and RN. Reconstruction error decreases for increasing number of hidden nodes as expected for all models as shown in Figure \ref{fig:gbm_error_vs_hiddens}, \ref{fig:rbm_error_vs_hiddens}, \ref{fig:rn_error_vs_hiddens} respectively. We chose the number of hidden nodes as 400 for RBM and GBM and 1024 for RN since there is no significant decrease of reconstruction error more than these numbers of hidden nodes.}

\begin{figure}[hbt!]
\centerline{
	\subfigure{
    	\includegraphics[width=0.5\textwidth]{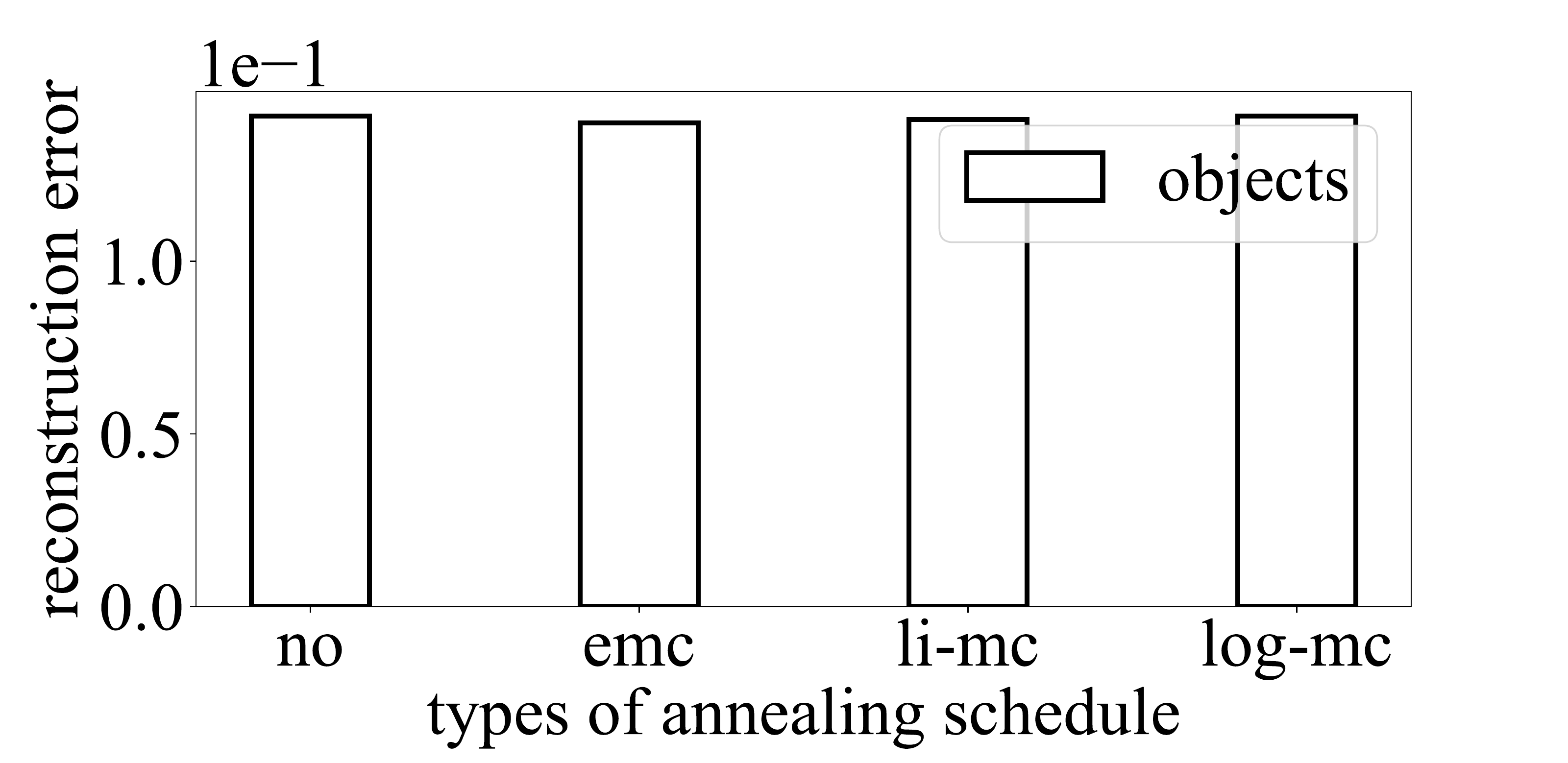}
    }
}
\centerline{
	\subfigure{
    	\includegraphics[width=0.5\textwidth]{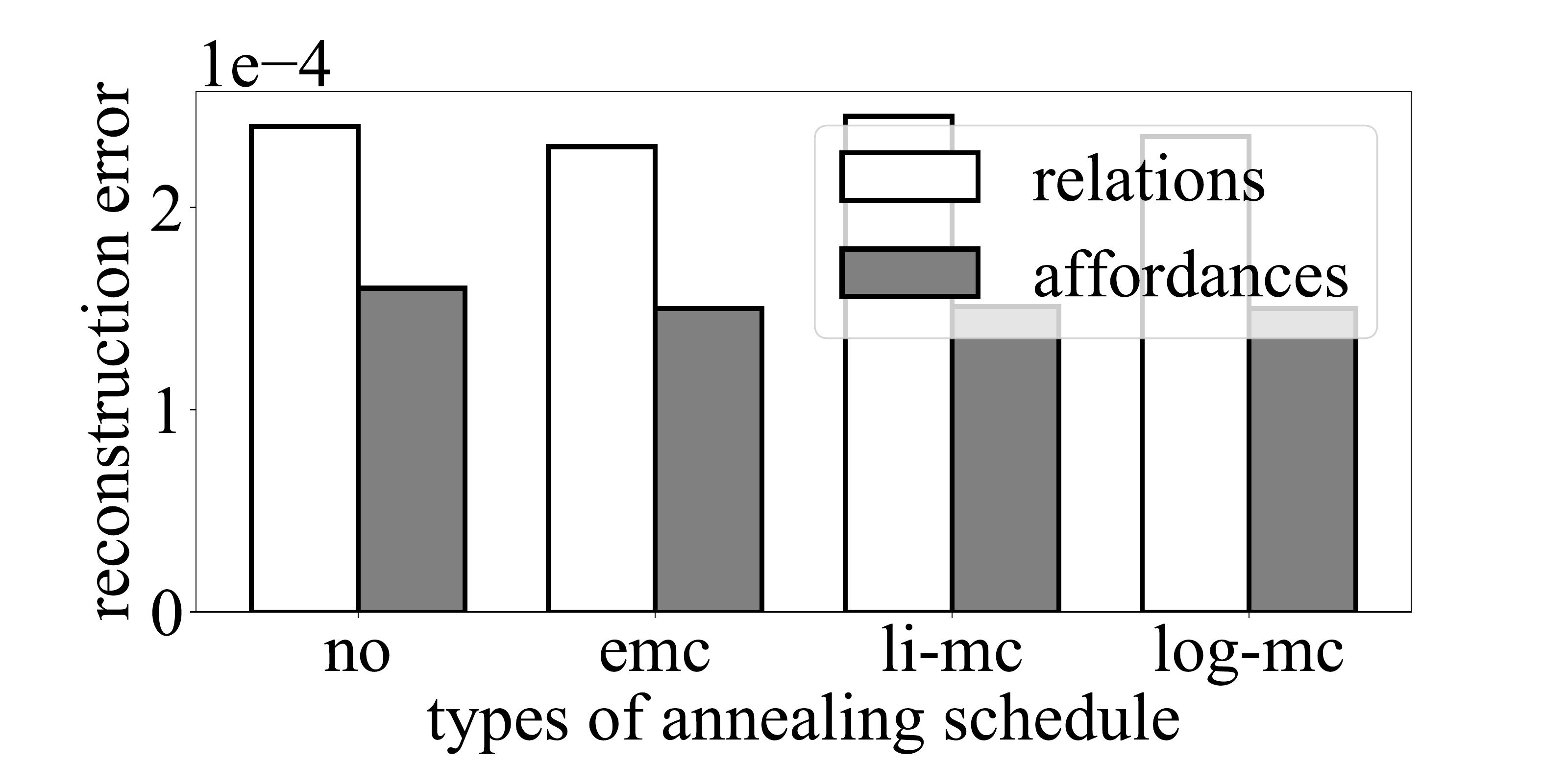}
    }
}
\caption{\SEC{Reconstruction errors of GBM for different annealing schedules with initial temperature 4.0. (a) Reconstruction error for object nodes. (b) Reconstruction error for relation and affordance nodes.}\label{fig:gbm_error_vs_temperature}}
\end{figure}

\begin{figure}[hbt!]
\centerline{
	\subfigure{
    	\includegraphics[width=0.5\textwidth]{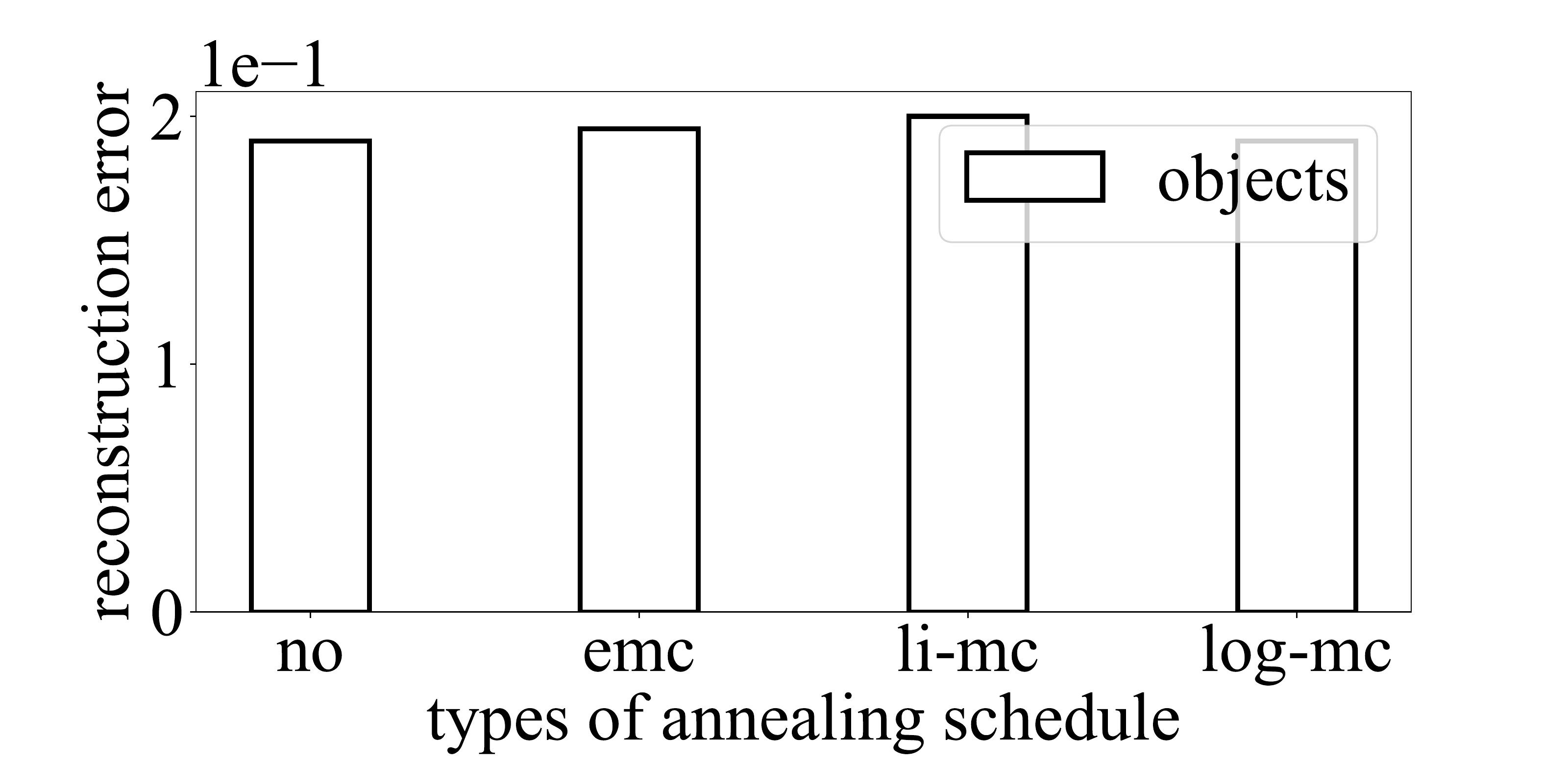}
    }
}
\centerline{
	\subfigure{
    	\includegraphics[width=0.5\textwidth]{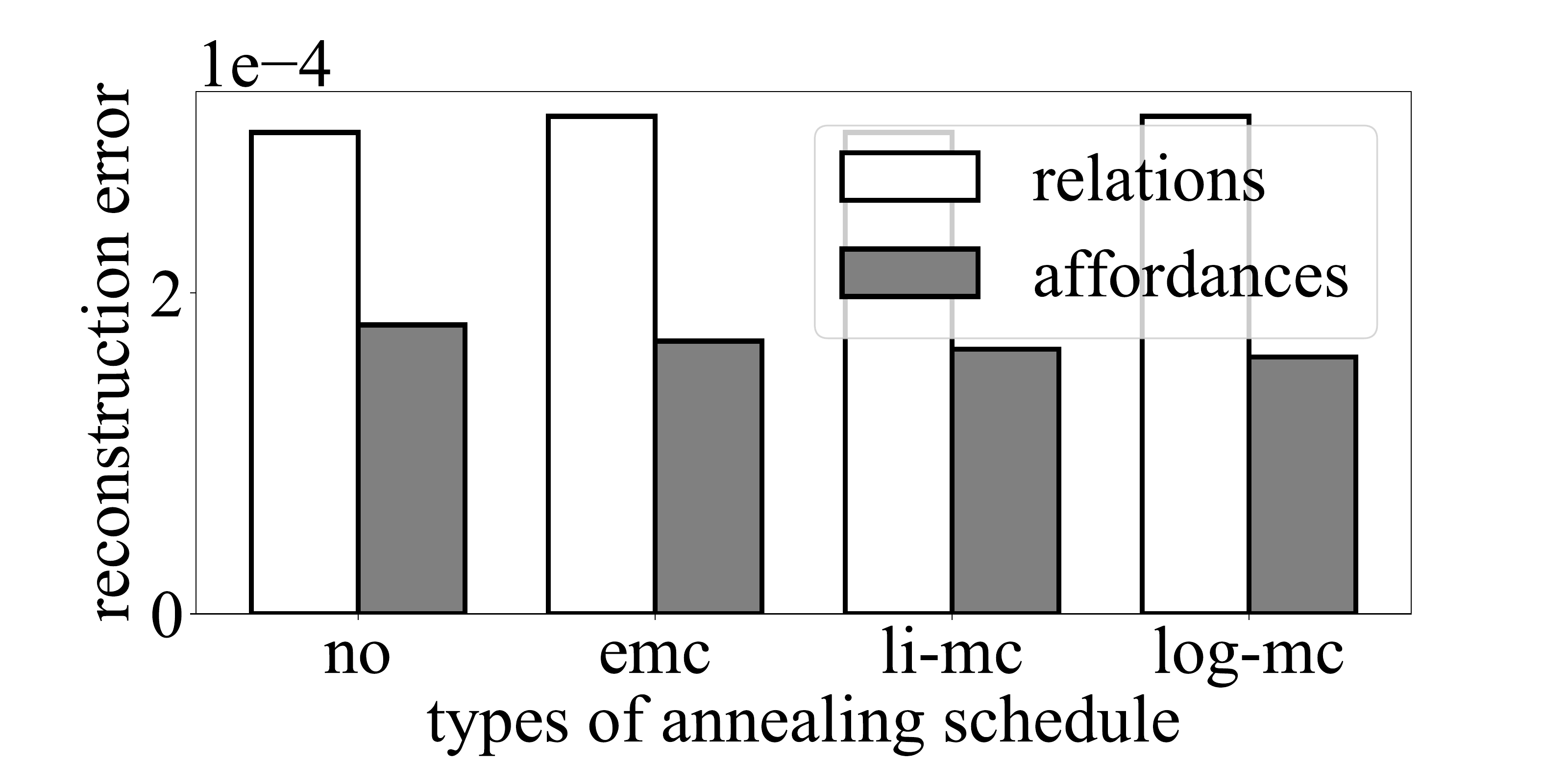}
    }
}
\caption{\SEC{Reconstruction errors of RBM for different annealing schedules with initial temperature 4.0. (a) Reconstruction error for object nodes. (b) Reconstruction error for relation and affordance nodes.}\label{fig:rbm_error_vs_temperature}}
\end{figure}

\SEC{Lastly, we analyzed effect of annealing method for GBM and RBM (not applicable for RN). As shown in Figure \ref{fig:gbm_error_vs_temperature},\ref{fig:rbm_error_vs_temperature}, different annealing schedules does not effect reconstruction error significantly as in COSMO (Fig. \ref{fig:error_vs_temperature}). We preferred emc annealing schedule.}

\end{document}